\documentclass{article}

\usepackage[eandd, final]{neurips_2026}

\usepackage[utf8]{inputenc} %
\usepackage[T1]{fontenc}    %
\usepackage{hyperref}       %
\usepackage{url}            %
\usepackage{booktabs}       %
\usepackage{amsfonts}       %
\usepackage{nicefrac}       %
\usepackage{microtype}      %
\usepackage[dvipsnames]{xcolor}         %
\usepackage{colortbl}   %
\usepackage{graphicx}   %
\usepackage{adjustbox}  %
\usepackage{float}     %
\usepackage{multirow}
\usepackage{longtable}
\usepackage{caption}
\usepackage{enumitem}

\usepackage{booktabs}
\usepackage{xspace}
\usepackage{amsmath}

\usepackage{cleveref}
\crefname{table}{Table}{Tables}
\crefname{figure}{Fig.}{Figs.}
\crefname{section}{Sec.}{Secs.}

\usepackage{pifont}

\definecolor{cvprblue}{rgb}{0.21,0.49,0.74}
\definecolor{coarse_col}{RGB}{103, 78, 167}
\definecolor{fine_col}{RGB}{106, 168, 79}
\definecolor{labels_col}{RGB}{234, 153, 153}

\definecolor{robin}{HTML}{009988}    %
\definecolor{dusky}{HTML}{0077BB}    %
\definecolor{naumann}{HTML}{EE7733}  %
\newcommand{\dataset}{PS-SPLIT\xspace}

\usepackage[normalem]{ulem}
\definecolor{addcol}{RGB}{0,140,0}
\definecolor{delcol}{RGB}{210,0,0}
\definecolor{whycol}{RGB}{0,70,220}
\newif\ifshowchanges
\showchangesfalse
\ifshowchanges
  
  \DeclareRobustCommand{\del}[1]{{\color{delcol}\sout{#1}}}
  \DeclareRobustCommand{\why}[1]{{\color{whycol}\footnotesize\textbf{[Change: #1]}}}
  \newenvironment{addedblock}{\color{addcol}}{}
\else
  
  \DeclareRobustCommand{\del}[1]{}
  \DeclareRobustCommand{\why}[1]{}
  
\fi

\newcommand{\ie}{i.e.,\@\xspace}
\newcommand{\eg}{e.g.,\@\xspace}
\newcommand{\vs}{vs.\@\xspace}

\title{Predicting Species Splits: A Challenging Fine-Grained Benchmark for Category Discovery}

\author{%
  Christian Lange\textsuperscript{1} \quad
  Joakim Bruslund Haurum\textsuperscript{2} \quad
  Bingchen Zhao\textsuperscript{1} \quad
  Oisin Mac Aodha\textsuperscript{1} \\
  \textsuperscript{1}University of Edinburgh \quad \quad
  \textsuperscript{2}University of Southern Denmark   
}

\begin{document}

\maketitle

\begin{abstract}
Widely used category discovery benchmarks have become increasingly saturated and often fail to reflect real-world discovery settings, making long-term evaluation of new methods difficult. We introduce PS-SPLIT, a large-scale fine-grained benchmark for category discovery, motivated by taxonomic splitting: the real-world process by which what was once considered a single species is found to comprise multiple distinct ones. PS-SPLIT contains 79K globally distributed, high-quality, community-collected bird images spanning 771 species, derived from 257 recent real expert-verified taxonomic splits, with associated geographic metadata and full taxonomic hierarchy. The visual differences between categories arising from real taxonomic splits can be very subtle, often requiring significant human expertise to tell apart, making PS-SPLIT a challenging benchmark which we hope will assist in advancing fine-grained category discovery research. Based on taxonomic splitting, we introduce the task of discovery by category splitting, where the goal is to determine which known categories should be subdivided into new finer-grained subcategories. We adapt four generalized category discovery methods to this setting, evaluate existing methods, and show that there is substantial room for future progress.
\end{abstract}

\section{Introduction}
\label{sec:intro}
It is now possible to train vision models to perform large-scale fine-grained visual categorization from labeled datasets spanning tens of thousands of possible categories.
However, conventional supervised training assumes that the training labels are mostly accurate and that the label distribution is static, \ie the categories do not change.
In practice, the real world is more complex as new concepts are constantly being discovered or invented and existing concepts are being split into new ones.
This is visible across many domains, \eg the rise of new fashion subcultures, the branching of musical genres, the evolution of linguistic and cultural categories, innovation in consumer products, and the discovery of new diseases in medicine.
The next frontier in computer vision is the development of models that can assist with this kind of open-ended concept discovery.

\begin{figure}[t]
\centering
\includegraphics[width=1.0\columnwidth]{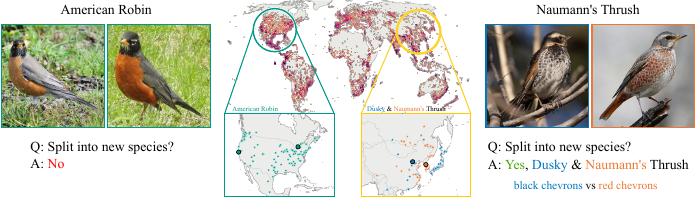}
\vspace{-12pt}
\caption{
\textbf{Real-world visual categories can split over time.} 
Here we illustrate two categories from our \dataset{} dataset. 
The American Robin, which can be found across North America, has not been subject to any taxonomic splits. 
In contrast, the \textcolor{dusky}{\textit{Dusky Thrush}} was previously considered to be a subspecies of \textcolor{naumann}{\textit{Naumann's Thrush}}, but has subsequently been determined to be a distinct species based on appearance/morphological, geographic, and genetic cues~\cite{duskythrush}.
The world map shows the distribution of images in \dataset, 
and locations where the individual birds depicted have been observed are denoted with colored circles in the regional maps.
}
\label{fig:teaser}
\vspace{-10pt}
\end{figure}

A particularly clear instance of this challenge arises in biology.
Every year, scientists and nature enthusiasts discover thousands of new species~\cite{mora2011many}.
Many of these are not the result of some new, previously unobserved, organism being spotted in the wild for the first time.
Instead, they are the product of experts deducing that what was previously thought to be one species is in fact two or more (see example species split in \cref{fig:teaser}).
A well-known recent example is the giraffe, previously thought to be one species, but now recognized as four distinct species illustrated by differences in their spot patterns~\cite{fennessy2016multi}.
The ultimate decision to split a species is typically based on genetic information, but is supported by more easily obtainable cues such as morphological/appearance differences or their geographic distribution~\cite{CHURCH20253556}.
Determining whether an existing species should be split is currently a very time-consuming task that requires significant domain expertise.
We envisage a complementary workflow whereby an automated system proposes potential species splits, that can then be later verified by human domain experts.

To enable research on this problem, we introduce \dataset, a new fine-grained benchmark for category discovery built from real taxonomic splits.
\dataset contains 79K community-collected images of 771 bird species sourced from the citizen-science platform iNaturalist~\cite{iNaturalist}.
It is organized around 257 taxonomic splits supported by scientific evidence that have been documented over the past 12 years. 
Each split is also matched with 257 taxonomically similar non-splitting `control' species. 
We release \dataset in two versions: \dataset-Full, with all 257 splits, and \dataset-Small, a 64-split subset of a similar size to existing category discovery benchmarks.
Each observation/image also includes geographic metadata and full taxonomic hierarchy information.
As the categories arise from real taxonomic splits, \ie from species that are so similar they were considered the same species until recently, visual differences can be very subtle and often require significant expertise to tell apart, making \dataset a very challenging benchmark for fine-grained category discovery.

Inspired by taxonomic splitting, \dataset enables a new task, \emph{discovery by category splitting} (DCS).
Our splitting task is related to existing category discovery work in computer vision~\cite{han2019learning,vaze22generalized,he2025category}. 
However, conventional category discovery methods typically explore the setting where a large number of labeled and unlabeled images are provided during training and the goal is to group the unlabeled images into new, or existing, categories.
In practice, we argue that a more realistic task involves taking images from existing categories and determining  if they should be split into multiple new fine-grained categories, or not.
Our framing of the problem is also connected to non-discovery work that trains on coarse supervision and then tests on fine categories~\cite{ristin2015categories,taherkhani2019weakly}.
However, unlike the majority of existing work that trains on coarse supervision, in our setting we cannot assume that \emph{all} the coarse entities should be split into finer ones, \ie only a small subset of existing categories are likely to be split into a new sub-categories. 
Furthermore, we would also like to generalize to categories that were not observed at training time, \ie the test categories may be disjoint from those in our training set. 

We make the following contributions:
(i) We introduce \dataset, a large-scale fine-grained benchmark for category discovery comprising 79K bird images from 771 species organized around 257  expert-verified taxonomic splits and matched controls, with associated geographic and taxonomic metadata.
(ii) We formalize the task of \emph{discovery by category splitting}, a new category-discovery task in which the model must decide, for each previously unseen coarse category, whether it should be split, into how many new categories, and which images belongs to which new category.
(iii) Detailed evaluation, where we adapt and benchmark existing  category discovery methods on \dataset, in addition to performing experiments on new category splitting versions of three existing fine-grained datasets (CUB-200-2011, Stanford Cars, and FGVC Aircraft). %

\section{Related Work}
\label{sec:rel_work}

\noindent\textbf{Category Discovery.}
Given a set of labeled and unlabeled images, category discovery methods aim to group images from the unlabeled set into novel categories~\cite{he2025category}. 
The task differs from unsupervised clustering in that the labeled images provide guidance as to what features are informative for forming categories. 
Different variants of the problem have been proposed, \eg where the unlabeled data only contains images from novel categories~\cite{han2019learning,fini2021unified}, where it contains a mix of novel and previously seen categories~\cite{vaze22generalized,cao2022open, wen2023simple}, or where discovery is performed in an online/incremental~\cite{zhao2023incremental,zhang2022grow} or non-transductive~\cite{du2023fly} setting. 
Most methods typically assume that the number of novel categories is known~\cite{han2021autonovel,fini2021unified,wen2023simple}, but there are also approaches for estimating the number using limited human supervision~\cite{perez2023human} or in a completely unsupervised manner~\cite{fini2021unified,vaze22generalized,zhao2023learningsemisupervisedgaussianmixture}. 
Graph-based discovery methods~\cite{zhang2022promptcal,otholt2024guided,cao2022pss,xing2021learning} have also been proposed that use graph clustering techniques~\cite{cao2022pss,xing2021learning} to automatically estimate the number of categories present. 
Concurrent to our work, \cite{liu2026letssplitupzeroshot} introduce a category-splitting task for fine-grained video understanding where coarse categories learned by pre-trained models are refined into finer ones.
Unlike our setting, they are given the coarse category to split and the names of its subcategories, and every evaluated category is split. We instead must decide, without names or labels, whether each unseen coarse category splits and into how many. %
In our experiments we adapt several existing generalized category discovery methods %
and demonstrate that  they struggle on \dataset.

\noindent\textbf{Learning from coarse supervision.}
Our problem setting also relates to learning from coarse supervision, a task  which has been explored under several different settings. 
The first is where only coarse labels are available at training time, but the test set contains finer instances of these coarse categories~\cite{huh2016makes,wu2018improving,touvron2021grafit}. 
\cite{wu2018improving,touvron2021grafit} both make use of a non-parametric memory bank which is learned from coarse supervision and fine-grained instances are compared to it using a nearest neighbor look-up at inference time. 
In the second setting, coarse labels are available for all images at training time, but a small subset of those images also have finer labels~\cite{ristin2015categories,taherkhani2019weakly}. 
\cite{su2021semi} explored a hierarchical semi-supervised setting where coarse labels were again provided for all images and finer labels only for some, but there were also additional `out-of-class' distractor categories in the training set without fine labels. 
Few-shot techniques have also been proposed that first train on coarse labels and  then evaluate using a small number of fine-grained labels~\cite{xu2023hyperbolic,bukchin2021fine}.  
Finally, there are approaches that combine coarse label supervision with finer-grained language information obtained from text knowledge bases~\cite{choudhury2024curious}. 
Unlike our setting, coarsely supervised methods typically assume that the label space in the train and test sets is the same.
Specifically, the test set contains a more granular (\ie finer-grained) splitting of the training labels.
In our setting, test categories may be novel and not observed during training.

\noindent\textbf{Geographic priors.} 
Knowing where an image was taken can  help determine which category is depicted in it. 
Combining location information with image classifiers can significantly improve classification performance~\cite{tang2015improving,macaodhaiccv19,chu2019geo}. 
One approach is to train a separate image classifier and spatial prior, \ie a model that predicts the likelihood of a category being present given a location, via non-parametric~\cite{berg2014birdsnap} or learning-based methods~\cite{macaodhaiccv19,coleICML2023,russwurm2024geographic}. 
Alternatively, methods that fuse location information into the image classifier during training have also been proposed~\cite{tang2015improving,chu2019geo,yang2022dynamic}.
These methods enable the model to learn how to combine the information from the two different sources.  
Our experiments show that intermediate feature representations encoded by existing spatial priors such as~\cite{coleICML2023} are helpful for splitting bird species that are present in \dataset.

\noindent\textbf{Fine-grained category discovery benchmarks.}
Category discovery methods are typically evaluated on a small set of standard fine-grained vision datasets, namely CUB-200-2011~\cite{wah2011caltech}, Stanford Cars~\cite{krause20133d}, FGVC Aircraft~\cite{maji2013fine}, and Herbarium19~\cite{tan2019herbarium}, along with less fine-grained datasets such as CIFAR-10/100~\cite{krizhevsky2009learning}.
None of these were designed for the discovery setting, often producing tasks that saturate quickly, \eg performance on CIFAR-10 is near-perfect, and headroom on the other benchmarks is shrinking.
\dataset is intended to complement these benchmarks with a fine-grained, naturally hierarchical, and substantially harder evaluation setting whose coarse--fine structure is grounded in real taxonomic decisions.
Ultra-fine-grained recognition benchmarks~\cite{yu2021benchmark} target categories that even experts find hard to separate (\eg plant cultivars). In contrast, \dataset has a defined label hierarchy determined by established species taxonomy, and the task involves discovering whether a category should be split.

\noindent\textbf{Biodiversity image benchmarks.}
A complementary line of work has built large-scale ecology and biodiversity image datasets for primarily evaluating classification.
The iNaturalist Challenge datasets~\cite{van2018inaturalist,van2021benchmarking} span thousands of species across plants, animals, and fungi, and were later adapted for incremental category discovery in~\cite{zhao2023incremental}.
BIOSCAN-5M~\cite{gharaee2024bioscan} pairs images with DNA barcodes for insect taxonomy.
Closer to discovery, Open-Insect~\cite{chen2025open} introduces an open-set recognition benchmark for novel insect species across geographic regions.  
These benchmarks emphasize scale and coverage and do contain visually very similar species, with many insect species in BIOSCAN-5M being visually indistinguishable.
CrypticBio~\cite{manolache2025crypticbio} collates visually confusing species based on past user misidentifications on iNaturalist and evaluates vision-language models on classifying them. 
\dataset also contains visually similar species, but selects them based on documented taxonomic changes, and focuses on category splitting rather than classification. 
\dataset complements these existing datasets by targeting the question of when a \emph{known} coarse category should itself be subdivided into finer ones, grounded in documented expert-defined taxonomic splits.

\section{The PS-SPLIT Benchmark}
\label{sec:dataset}

\begin{figure}[t]
\centering
\includegraphics[width=1.0\columnwidth]{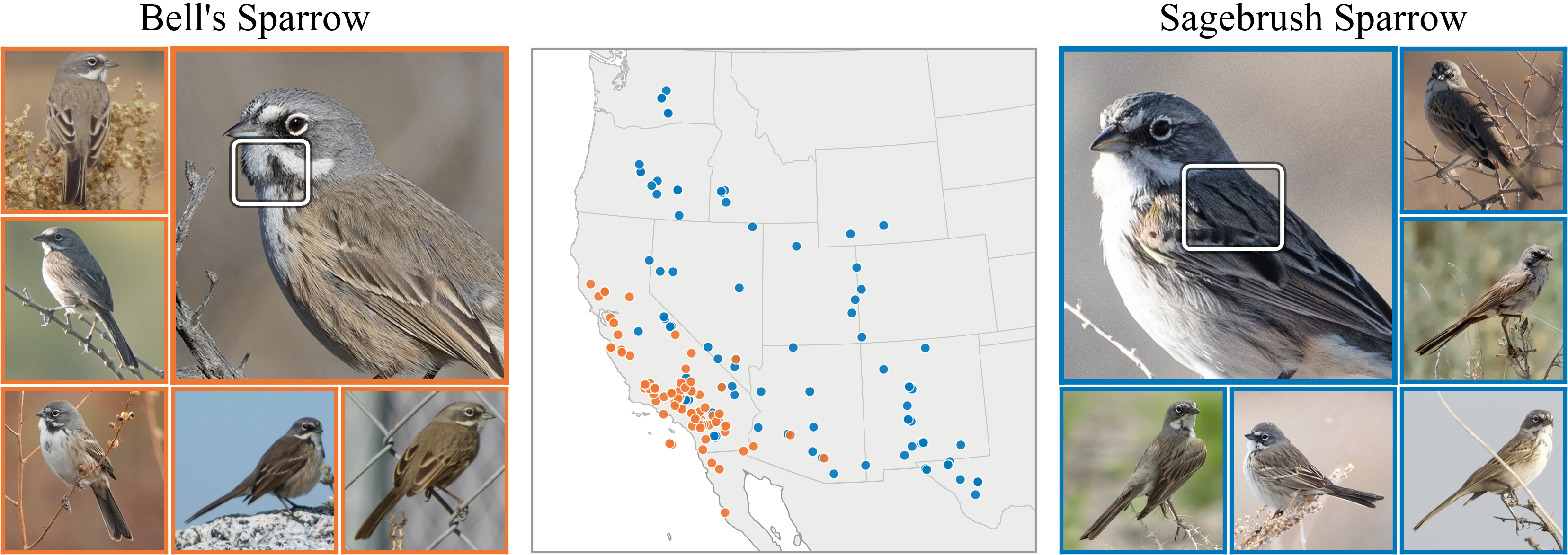}
\caption{
\textbf{Splitting example from \dataset.}
\textcolor{RoyalBlue}{Sagebrush Sparrow \textit{(Artemisiospiza nevadensis)}} tends to have more distinct black streaks on the back and is more widely geographically distributed, while \textcolor{orange}{Bell's Sparrow \textit{(Artemisiospiza belli)}} tends to have a more distinct malar (`mustache') and is found mostly in California. Distinctive features of each are shown in white boxes on the larger images.}
\label{fig:fig_2}
\vspace{-10pt}
\end{figure}

We introduce \dataset, a new fine-grained benchmark for category splitting built from real taxonomic splits documented on the citizen science platform iNaturalist~\cite{iNaturalist} over the past 12 years.
Each split contributes a coarse class (the previously accepted species), along with the two newly recognized fine species.
In addition, each split is paired with a taxonomically close \emph{control} coarse class that has not undergone a split.
Throughout, a \emph{coarse} class is the category before any split and a \emph{fine} class is a currently accepted species. Control coarse classes have not undergone a splitting event, so their coarse and fine labels coincide. 
The control design ensures that splitting decisions cannot be predicted from coarse-category size, broad taxonomic group, or geographic footprint alone.
\dataset contains 79{,}061 research-grade (\ie verified) bird images across 771 species drawn from 257 splits and 257 matched controls, with per-image latitude/longitude metadata and full taxonomic hierarchy (order, family, genus).
\cref{fig:fig_2} shows example images and locations for a taxonomic split included in \dataset, including small visual differences that may allow this split to be detected.

\begin{table}[t]
\centering
\caption{\textbf{\dataset statistics.} \dataset is a new multimodal benchmark for the task of category splitting. It pairs one splitting coarse class (containing two fine classes), with one taxonomically close control coarse class (a single species, so its coarse and fine labels coincide). Each aligned pair of coarse classes is assigned to either the known or novel partition. We report statistics for both \dataset-Full (top) and the smaller \dataset-Small (bottom) below. 
Mean $|\Delta|$ is the per-alignment absolute difference between the splitting-side and control-side image totals. 
It is near zero by design, because each control species is image-capped to match the total observation count of its splitting pair.}
\vspace{3pt}
\label{tab:aves_summary_full}
\resizebox{0.95\linewidth}{!}{%
\begin{tabular}{clrrrrrrrrr}
 & Partition & \# Images & \# Coarse & Splits & Controls & \# Fine & Imgs/coarse & Imgs/fine & Fine/coarse & Mean $|\Delta|$ \\
\midrule
\multirow{3}{*}{\rotatebox[origin=c]{90}{Full}}
 & Known & 39,499 & 256 & 128 & 128 & 384 & 154.3 & 102.9 & 1.50 & 2.4 \\
 & Novel & 39,562 & 258 & 129 & 129 & 387 & 153.3 & 102.2 & 1.50 & 2.0 \\
 & Total & 79,061 & 514 & 257 & 257 & 771 & 153.8 & 102.5 & 1.50 & 2.2 \\
\midrule
\multirow{3}{*}{\rotatebox[origin=c]{90}{Small}}
 & Known & 12,645 &  64 &  32 &  32 &  96 & 197.6 & 131.7 & 1.50 & 1.3 \\
 & Novel & 12,640 &  64 &  32 &  32 &  96 & 197.5 & 131.7 & 1.50 & 0.0 \\
 & Total & 25,285 & 128 &  64 &  64 & 192 & 197.5 & 131.7 & 1.50 & 0.7 \\
\bottomrule
\end{tabular}
}
\vspace{-10pt}
\end{table}

\noindent\textbf{Dataset construction.}
We source data from research-grade observations, \ie community verified images with associated metadata, on iNaturalist. 
Research-grade observations only require community agreement on taxonomic identity, but the images themselves could be of very low quality. 
As a result, additional filtering is needed to obtain a discriminative fine-grained dataset.
\dataset is built using five processing stages:
\textit{(i)} \textbf{split and control selection} from the iNaturalist taxonomic-changes feed~\cite{iNaturalistSplit} (two-way splits with at least 20 research-grade observations per generated species; controls drawn from the same genus or family with similar observation counts  and never having undergone a taxonomic change themselves.
We use binary splits as they are the most common, \ie at the time of writing there were 1,208 binary \vs 425 multi-way species-level splits on iNaturalist since 2012.
To evaluate category splitting approaches outside the binary setting, we also adapt existing fine-grained datasets to have multi-way splits (\cref{sec:exps}). 
\textit{(ii)} \textbf{image cap with user diversification}, retaining up to 100 images per species and preferring distinct uploaders/users so no single user's photography style dominates.
\textit{(iii)} an \textbf{object detector to remove small instances}, retaining detections from MegaDetector~\cite{beery2019efficient} with animal-class confidence $\geq$0.75 and shortest bounding-box edge $\geq$190\,px, then resizing each crop to 256$\times$256. 
\textit{(iv)} a \textbf{vision--language quality filter} (Qwen2.5-VL-7B-Instruct~\cite{bai2025qwen25vltechnicalreport}) that scores seven binary criteria (bird present, adult, not obscured, in focus, no text/watermark, no particularly unusual images, and no other prominent animals) and assigns a coarse bird pose label, with pose consistency enforced across species produced from a taxonomic split so that a split decision cannot be made simply because one species is photographed in a particular pose more often. \textit{(v)} A \textbf{visual-difference screen} in which a DINOv2~\cite{oquab2023dinov2} linear probe trained on 20\% of each species pair's images must exceed 60\% accuracy on classifying the remaining data for that pair of species, removing species that are visually indistinguishable and may have been split using non-visual information (\eg genetic information may be required to determine the split).
In practice this only removes 6 of the 263  (2.3\%) candidate taxonomic splits  as most bird (\ie aves) taxonomic splits do involve subtle visual differences.
As this filter relies on a pretrained representation, we analyze its potential for selection bias in Appendix~\ref{sec:filter_sensitivity}: method rankings remain stable under stricter thresholds (Spearman $\rho=0.96$ at 0.70, and $0.78$ at 0.90, which discards 62.6\% of coarse classes).
Additional details of our processing pipeline, including prompt templates, and worked examples are provided in Appendix~\ref{sec:construction}.

\noindent\textbf{Statistics.}
\Cref{tab:aves_summary_full} reports the top-level statistics of \dataset.
By construction, every splitting coarse class is paired with a control coarse class of nearly identical image count (mean $|\Delta|\!\approx\!2$ images per pair). 
Half of the pairs are then assigned as `known', where the fine classes are available during training, with the rest assigned as `novel' where the number of fine classes is unknown.
\dataset comes in two versions: \dataset-Full, which contains all 257 splits and controls, and a smaller \dataset-Small subset (25{,}285 images, 192 species, 64 splits + 64 controls) for budget-constrained evaluation, roughly matching the size of existing benchmarks for category discovery. 
The full \dataset dataset spans 22 orders, 94 families, and 262 genera.
Geographic coverage is global, weighted toward South and Southeast Asia (19.8\%), South America (15.8\%), and North America (12.3\%). 
Density heatmaps and per-country totals are in  Appendix~\ref{sec:geo_appendix}, with full taxonomic coverage  in Appendix~\ref{sec:taxonomy_appendix}.

Compared to datasets such as CUB-200-2011~\cite{wah2011caltech}, Stanford Cars~\cite{krause20133d}, FGVC Aircraft~\cite{maji2013fine}, \dataset is larger, and has a coarse--fine taxonomy that arises from real documented species-level taxonomic splits, ensuring that fine categories are very visually similar and challenging to separate, and provides geographic metadata.
These properties support the discovery by splitting task we introduce in \cref{sec:task} and provide a natural setting for evaluating geographic priors when visual classification is challenging.

\noindent\textbf{Licensing, intended use, and known biases.}
All included observations are research-grade iNaturalist records under licenses that permit redistribution for research.
\dataset is intended for benchmarking fine-grained category discovery and category-splitting algorithms. 
It should not be used as a source of ground-truth taxonomic decisions.
The dataset inherits biases from its citizen-science source, \eg the distribution of bird observations in \dataset correlates with the distribution of observers rather than the true distribution of bird species, and some regions of the globe contain few iNaturalist contributors and splits or control species in these locations may not have enough suitable images to be included in the dataset. 
This leads to some regions with very few observations, as shown in Appendix~\ref{sec:geo_appendix}.
The image cap and user-diversity preference flatten the distribution bias but do not remove broader geographic/taxonomic skews (\cref{sec:limitations}).
The full dataset is available at \href{https://github.com/PS-SPLIT/PSSPLIT}{https://github.com/PS-SPLIT/PSSPLIT}.

\section{The Discovery by Category Splitting Task}
\label{sec:task}
Here we introduce our category splitting task which is the basis of our evaluation on \dataset. 

\noindent\textbf{Inputs and outputs.}
At training time, the model sees a known set of coarse classes with two levels of label granularity per instance: a coarse label $y^c$ (the previously accepted category, before any split) and a fine label $y^f$ (the currently accepted category, after splitting).
Instances within a coarse class that splits into new species share $y^c$ but may have different $y^f$, while all instances within a non-splitting coarse class share the same $y^f$.
In the case of \dataset, each image instance additionally has geographic metadata $\mathbf{z}_i = (\text{lat}_i, \text{lon}_i)$. 
At test time, the model is given a disjoint set of previously unseen coarse classes, each represented by its instances and  their optional metadata.
For every test coarse class the model must decide \textit{(i)} whether the class should be split, \textit{(ii)} the number of fine classes $K$ if it does, and \textit{(iii)} the assignment of each image instance to a new fine class, if it splits.

\noindent\textbf{Notation.}
Training data is denoted as  $\mathcal{D}^s = \{(\mathbf{x}_i, \mathbf{z}_i, y^f_i, y^c_i)\}_{i=1}^{N^s}$ and test data as $\mathcal{D}^t = \{(\mathbf{x}_j, \mathbf{z}_j, y^c_j)\}_{j=1}^{N^t}$, with the test fine labels $y^f_j$ unobserved.
The coarse-label spaces are disjoint, $\mathcal{Y}^{c,s} \cap \mathcal{Y}^{c,t} = \emptyset$, so the model has no prior knowledge of which test coarse classes will split, and the total number of fine classes $|\mathcal{Y}^t|$ is unknown.
Importantly, if splitting occurs, the entity that is ultimately split into new categories has typically been previously designated as a single species and is already a \emph{fine-grained} concept, \ie the coarse class to be split is not a high-level visual concept (such as \textit{seabird}), but a more specific concept (such as \textit{Herring Gull}).  %

\noindent\textbf{Protocol.}
On \dataset we adopt a \textit{50/50 known/novel} split at the level of aligned coarse-class pairs such that each splitting class is paired with a control, and each pair is assigned in its entirety to either the known (training) or novel (test) partition (\cref{tab:aves_summary_full}). 
This ensures the test partition contains a balanced mix of classes that should split and classes that should not, and that a model cannot exploit splitting/control imbalance during training.
The maximum cardinality to predict per coarse class is $K_\text{max} = 2$ for \dataset, since all included splits are two-way (for cases where a split occurs).

\noindent\textbf{Evaluation metrics.}
We report performance using multiple metrics.
\emph{Clustering accuracy} (Cl.\,Acc) is the standard Hungarian-aligned accuracy of fine-label assignment~\cite{han2019learning}, computed per coarse class and macro-averaged. 
\emph{AMI} is the corresponding adjusted-mutual-information score.
We also report \emph{split-only} variants (Cl.\,Acc$_s$, AMI$_s$) restricted to coarse classes with true cardinality $K^*\!>\!1$, since these are the cases where the model has actually been asked to discover structure.
Multiple metrics are needed because simple baselines can perform well on one metric while being useless for the discovery by category splitting task, \eg never splitting achieves perfect accuracy on every control class (Appendix~\ref{sec:lower_bounds}). 
\emph{K accuracy} (K Acc) is the smooth score $1/(1+|\hat{K}-K^*|)$ averaged over all coarse classes, with split-only variant K Acc$_s$.
$\Delta\!\sum\!K \;=\; \sum \hat{K} - \sum K^*$ is the signed deviation of the total predicted fine class count from the truth, indicating whether a method tends to over- or under-split.
Finally, \emph{Split Precision} is the precision of the binary ``this coarse class should be split'' decision; of the coarse classes the model predicts as multi-fine, the fraction whose true cardinality is indeed greater than one.
Full metric definitions are provided in Appendix~\ref{sec:metrics_appendix}.

\section{Category Splitting Methods}
\label{sec:baselines}

We benchmark several methods on \dataset, organized into four families: visual backbones, geographic features, K estimation, and downstream clustering.
All baselines share the same overall pipeline: extract features for each instance from a visual backbone $f_\theta$  (optionally concatenated with a geographic feature from a location encoder $g_\phi$), estimate the number of fine classes $K$ for each test coarse class, then cluster the features into $K$ groups.
Additional architecture and training details can be found in Appendix~\ref{sec:transformer_appendix}.

\noindent\textbf{Visual backbones.}
We evaluate six backbones, all ViT-B-based so that they remain comparable.
A frozen DINOv1~\cite{caron2021emerging} and a frozen DINOv2~\cite{oquab2023dinov2}  with registers~\cite{darcet2024vision} serve as references that require no training.
We then fine-tune the DINOv2 with registers on the \emph{known} coarse classes of \dataset using the following category discovery methods: GCD~\cite{vaze22generalized}, SimGCD~\cite{wen2023simple}, UNO+~\cite{fini2021unified}, and SelEx~\cite{rastegar2024selexselfexpertisefinegrainedgeneralized}. 
This yields four trained backbones whose representations are tuned to fine-grained taxonomic distinctions seen during training.
The frozen and trained DINOv2 variants share the same pre-trained weights, so any improvement reflects fine-tuning rather than initialization.
All trained backbones are image-only, \ie they do not receive any location information during training.

\noindent\textbf{Geographic features.}
Where geographic metadata %
is available, we use a SINR~\cite{coleICML2023} location encoder trained with the AN\_FULL loss on the location training data from~\cite{coleICML2023} to produce a feature vector per observation based on the location the image was taken. 
This is then concatenated with the visual feature from $f_\theta$ before downstream $K$ estimation and clustering.
SINR is treated as a frozen feature extractor and is not fine-tuned on \dataset. 
Location information only enters the pipeline at the K-estimation and clustering stages, so the ``w/o Geo'' results are a location ablation of the pipeline. 

\noindent\textbf{K estimation.}
We also compare three families of methods for predicting the number of fine classes per test coarse class.
\textit{Tuning-based} methods select $K$ and perform clustering for each test class by tuning a single global hyperparameter on the known classes: \emph{Tuning-Agglo} sweeps the agglomerative-clustering distance threshold, \emph{Tuning-Spectral} sweeps the spectral-eigengap threshold~\cite{ng2001spectral, tutorial_spectral_eigengap}, and \emph{Tuning-XMeans} sweeps the PCA dimension before X-Means~\cite{pelleg2000x} selects $K$ via BIC. 
\textit{Brent-based} methods cluster a set of coarse classes with known fine classes and instances from an unknown coarse class jointly at varying $K$ and sweep through various values of $K$ to find the $K$ at which a held-out portion of the known data is best clustered, as in the K-estimator from GCD~\cite{vaze22generalized}. 
Pairing this estimator with KMeans, agglomerative, spectral, and GMM clustering yields four \emph{Brent-X} variants.
The Brent method with KMeans on a GCD backbone is typically used in generalized category discovery approaches that require the number of unknown classes to be known before training, as in SimGCD~\cite{wen2023simple} and SelEx~\cite{rastegar2024selexselfexpertisefinegrainedgeneralized}.
Finally, we also introduce a  \textit{Transformer K estimator}, which is a lightweight 2-layer 8-head transformer~\cite{vaswani2017attention} that takes all available backbone features from a coarse class as input and outputs a CLS token which is input to a cardinality predictor head that predicts $K$. 
It has roughly 500K parameters and trains in around ten minutes on a single A6000 GPU for the full \dataset.
Further details are provided in Appendix~\ref{sec:transformer_appendix}.
We additionally report an \emph{Oracle} as an upper bound that uses the true $K^*$.
We do \emph{not} include UNO+ or SimGCD parametric classification directly, as their training method assumes $K$ is known a priori, so we use these approaches as feature extractors for the K-estimation methods we do compare.

\noindent\textbf{Downstream clustering.}
By default, we use KMeans on the (visual, optionally visual+geo) features given the predicted $K$ from Brent or Transformer methods.
For the per-method clustering ablation in \cref{tab:aves_small_combined_methods}, we also report agglomerative, spectral, and GMM variants alongside KMeans and XMeans.

\section{Benchmark Results}
\label{sec:exps}

We benchmark the methods described in \cref{sec:baselines} on \dataset, training each method on the known coarse classes and reporting metrics on the held-out novel classes.
Implementation details for the inherited backbones follow their original publications. 
We report results on \dataset-Small in the main paper, %
and the corresponding results for \dataset-Full and our adapted fine-grained datasets can be found in Appendix~\ref{sec:additional_results}. 

\noindent\textbf{Headline results.}
\dataset is a challenging benchmark.
On \dataset-Small, the best non-oracle method reaches Cl.\,Acc.\ of $0.78$ against an oracle ceiling of $0.93$--$0.94$, a gap of roughly fifteen clustering-accuracy points (\cref{tab:aves_small_combined_methods}). 
On \dataset-Full (\cref{tab:aves_full_combined_methods}), the gap closes to roughly eleven points ($0.82$ \vs $0.93$).
Predicting that every coarse class does not split would achieve roughly $0.75$ clustering accuracy, as this would get perfect performance on the non-splitting control classes, and approximately 50\% on the splitting classes, due to the relatively balanced splitting fine class sizes.
However \emph{split-only} Cl.\,Acc$_s$ would only be $~$50\% and AMI$_s$ would be 0.
Lower-bound baselines (Never-Split, Always-Split, and Random+KMeans) are reported in \cref{tab:aves_small_methods_bounds,tab:aves_full_combined_methods} in the Appendix. 
None is competitive across all metrics, \eg Never-Split has AMI$_s=0.00$ and undefined Split Precision, while Always-Split over-splits every control ($\Delta\!\sum\!K=+32$, Split Precision $0.50$). 
Among the strongest configurations we observe are the transformer K-estimator paired with either DINOv2 or SelEx plus SINR features, but they sit well below the oracle.

\noindent\textbf{Backbone comparison.}
\Cref{tab:aves_small_combined_backbone} compares the eight backbone variants under both Brent + KMeans and the transformer K estimator on \dataset-Small.
Visual-only backbones cluster within a narrow band (Cl.\,Acc.\ $0.65$--$0.77$). 
There is no clear winner among DINOv1, DINOv2, GCD, SimGCD, UNO+, or SelEx, and several of them collapse toward ``always predict no split'' under Brent + KMeans (Cl.\,Acc.\ plateaus at $0.75$ with $\Delta\!\sum\!K = -32$ and undefined Split Precision, which is the value reached when a method predicts $K=1$ for every coarse class).
A SINR-only baseline that only uses location-based features  (row ``SINR'') is surprisingly competitive, indicating that geography carries substantial signal even without any visual features.
Concatenating SINR and DINOv2 features (row ``DINOv2 SINR'') is the strongest configuration under the transformer K estimator.

\noindent\textbf{K estimation $\times$ clustering.}
\Cref{tab:aves_small_combined_methods} sweeps clustering and K-estimation methods across four backbone setups (DINOv2 and SelEx, each with and without geographic features).
The transformer K-estimator and Brent-Spectral are competitive with each other with DINOv2 + SINR features, both reaching Cl.\,Acc.\ $= 0.78$ with $\Delta\!\sum\!K = +0$. 
With SelEx + SINR features the transformer (Cl.\,Acc.\ $0.78$, Cl.\,Acc.$_s$ $0.72$) is the strongest method.
Tuning-based methods are weak on this benchmark: Tuning-XMeans degenerates to ``never split'' across all setups, Tuning-Spectral does the same with DINOv2 + SINR features, and Tuning-Agglo over-splits dramatically whenever SINR features are added (Small: $\Delta\!\sum\!K = +212$ on DINOv2, $+186$ on SelEx; Full: $+534$ and $+501$, several times the next-largest absolute deviation in either table).

The transformer K-estimator consistently obtains higher \emph{split-only} metrics (Cl.\,Acc.$_s$, K Acc.$_s$) than Brent + KMeans across all four backbone setups, which we read as it being more reliable at recognizing the genuine splits, but in visual-only experiments it pays for this by over-predicting splits overall ($\Delta\!\sum\!K \!>\! 0$) so its Split Precision sits in the middle of the pack.
Adding location information to the same backbone systematically tames this tendency. 
On \dataset-Small with SelEx, the transformer's $\Delta\!\sum\!K$ falls from $+24$ to $+4$, and on the full dataset it falls from $+118$ (SelEx, no geo) to $-12$ (SelEx + geo), the only row in either methods table where the transformer under-predicts $K$.
On the full dataset, Brent-Agglo on SelEx + geo achieves the highest non-oracle Cl.\,Acc.\ ($0.82$) along with one of the highest Split Precision values ($0.69$). 
Other methods that nominally exceed this Split Precision (\eg Brent-KMeans + DINOv2 without geo on \dataset-Small at $0.80$) reach that value by predicting splits very rarely (\cref{tab:aves_small_combined_methods}, $\Delta\!\sum\!K\!=\!-17$).

\noindent\textbf{Effect of geographic information.}
Adding geographic features raises the oracle Cl.\,Acc.\ from $0.83$--$0.85$ to $0.93$--$0.94$ regardless of whether the visual backbone is default DINOv2 or the SelEx fine-tuned version. 
This indicates that geographic information carries roughly ten clustering-accuracy points of signal that visual features alone do not recover.
The non-oracle methods recover only part of this headroom, but consistently improve in the same direction. 
For example, on \dataset-Small the transformer's $\Delta\!\sum\!K$ moves toward zero on both DINOv2 ($+7\!\to\!+0$) and SelEx ($+24\!\to\!+4$); on the full dataset the analogous shifts are larger ($+57\!\to\!+30$ on DINOv2 and $+118\!\to\!-12$ on SelEx).
While geographic information is clearly useful for splitting, vision alone also carries substantial signal. A supervised DINOv2 linear probe trained with the fine labels reaches Cl.\,Acc.$_s$ $0.83$ without location (\cref{tab:aves_small_methods_bounds}), well above the vision-only discovery methods (at most $0.68$). 
This demonstrates that there is indeed sufficient visual signal in the images to split the categories. 
Appendix~\ref{sec:geo_breakdown} further describes the impact of geographic information and breaks down performance by continent. 

\noindent\textbf{Adapted fine-grained datasets.}
We additionally adapt CUB-200-2011~\cite{wah2011caltech}, Stanford Cars~\cite{krause20133d}, and FGVC Aircraft~\cite{maji2013fine} to the discovery by splitting setting by treating each dataset's natural higher-level grouping (genus / make / family) as the coarse class and the leaf class as the fine class; geographic features are unavailable for these datasets and are therefore omitted. 
Unlike \dataset, these datasets contain multi-way splits (up to $K=30$ for CUB-200-2011, $K=22$ for Stanford Cars, and $K=8$ for FGVC Aircraft) and unequal numbers of splitting and non-splitting coarse classes (\cref{tab:fg_stats}), and the methods we evaluate on \dataset can be directly applied to this $K>2$ setting. 
Per-dataset results are provided in Appendix~\ref{sec:additional_results}.
We additionally present results for the standard generalized category discovery task for these datasets and \dataset in Appendix~\ref{sec:gcd}.

\noindent\textbf{Failure modes.}
The transformer K-estimator's positive $\Delta\!\sum\!K$ when using only visual features reflects a tendency to predict spurious splits in coarse classes whose true cardinality is one, which depresses Split Precision below what its high split-only recall would otherwise allow.
Adding geographic features reduces or reverses this tendency, \ie in three of four backbone setups it pulls $\Delta\!\sum\!K$ closer to zero, and on the full dataset with SelEx + geo it pushes the transformer into mild \emph{under}-prediction ($\Delta\!\sum\!K = -12$).
Tuning-Agglo exhibits the opposite extreme failure mode, over-splitting almost every coarse class whenever geographic features are concatenated.

\begin{table}[t]
\centering
\caption{\textbf{Category splitting results on \dataset{}-Small.} Top: Brent + KMeans K estimation. Bottom: transformer K estimation. Best value per column is bolded within each section. Rows marked with $^*$ use additional geographic features. } 
\vspace{3pt}
\resizebox{\linewidth}{!}{
\begin{tabular}{c|l|cccc|cc|cc}
\toprule
 & Backbone & Cl.\,Acc. & Cl.\,Acc.$_\text{s}$ & AMI & AMI$_\text{s}$ & K Acc. & K Acc.$_\text{s}$ & $\Delta\!\sum\!K$ & Split Prec. \\
\midrule
\multirow{8}{*}{\rotatebox{90}{Brent-KMeans}} & DINOv1 & 0.70 & 0.52 & 0.31 & 0.00 & 0.73 & 0.66 & -10 & 0.45 \\
 & DINOv2 & 0.77 & 0.57 & 0.49 & 0.07 & \textbf{0.82} & 0.69 & -17 & \textbf{0.80} \\
 & UNO+ & 0.75 & 0.51 & 0.50 & 0.00 & 0.75 & 0.50 & -32 & -- \\
 & GCD & 0.73 & 0.54 & 0.42 & 0.06 & 0.74 & 0.59 & -19 & 0.46 \\
 & SimGCD & 0.75 & 0.51 & 0.50 & 0.00 & 0.75 & 0.50 & -32 & -- \\
 & SelEx & 0.75 & 0.54 & 0.42 & 0.03 & 0.75 & 0.59 & -20 & 0.50 \\
 & SINR$^*$ & \textbf{0.79} & \textbf{0.67} & \textbf{0.52} & \textbf{0.28} & 0.79 & \textbf{0.70} & -11 & 0.62 \\
 & DINOv2 SINR$^*$ & 0.75 & 0.64 & 0.42 & 0.24 & 0.73 & 0.66 & \textbf{-9} & 0.43 \\
\midrule
\multirow{8}{*}{\rotatebox{90}{Transformer}} & DINOv1 & 0.65 & 0.55 & 0.15 & 0.02 & 0.76 & 0.88 & +15 & 0.51 \\
 & DINOv2 & 0.70 & 0.62 & 0.28 & 0.16 & 0.76 & 0.81 & +7 & 0.51 \\
 & UNO+ & 0.72 & 0.58 & 0.30 & 0.07 & 0.77 & 0.78 & +1 & 0.55 \\
 & GCD & 0.68 & 0.68 & 0.18 & 0.21 & 0.76 & 0.94 & +23 & 0.51 \\
 & SimGCD & 0.66 & 0.66 & 0.12 & 0.17 & 0.77 & \textbf{1.00} & +30 & 0.52 \\
 & SelEx & 0.68 & 0.66 & 0.15 & 0.17 & 0.75 & 0.94 & +24 & 0.50 \\
 & SINR$^*$ & 0.77 & \textbf{0.75} & 0.39 & \textbf{0.41} & 0.77 & 0.86 & +11 & 0.53 \\
 & DINOv2 SINR$^*$ & \textbf{0.78} & 0.69 & \textbf{0.45} & 0.33 & \textbf{0.78} & 0.78 & \textbf{+0} & \textbf{0.56} \\
\bottomrule
\end{tabular}
}
\label{tab:aves_small_combined_backbone}
\vspace{-10pt}
\end{table}

\begin{table}[t]
\centering
\caption{\textbf{Category splitting results on \dataset{}-Small}.  Here, we vary K-estimation / clustering methods, across DINOv2 and the SelEx-finetuned backbone, each with and without geographic features. Best value per column is bolded within each section, oracle excluded.}
\vspace{3pt}
\resizebox{\linewidth}{!}{
\begin{tabular}{c|l|cccc|cc|cc}
\toprule
 & Method & Cl.\,Acc. & Cl.\,Acc.$_\text{s}$ & AMI & AMI$_\text{s}$ & K Acc. & K Acc.$_\text{s}$ & $\Delta\!\sum\!K$ & Split Prec. \\
\midrule
\multirow{9}{*}{\rotatebox{90}{DINOv2, w/o Geo}} & Oracle & 0.83 & 0.66 & 0.60 & 0.20 & 1.00 & 1.00 & +0 & 1.00 \\
 & Tuning-Agglo. & 0.59 & 0.58 & 0.12 & 0.15 & 0.56 & 0.67 & +74 & 0.51 \\
 & Tuning-Spectral & 0.70 & 0.55 & 0.32 & 0.05 & 0.71 & 0.62 & -11 & 0.38 \\
 & Tuning-XMeans & 0.75 & 0.51 & \textbf{0.50} & 0.00 & 0.75 & 0.50 & -32 & -- \\
 & Brent-KMeans & \textbf{0.77} & 0.57 & 0.49 & 0.07 & \textbf{0.82} & 0.69 & -17 & \textbf{0.80} \\
 & Brent-Agglo. & 0.74 & 0.53 & 0.43 & 0.02 & 0.76 & 0.59 & -21 & 0.55 \\
 & Brent-Spectral & 0.69 & 0.55 & 0.31 & 0.06 & 0.70 & 0.61 & -11 & 0.33 \\
 & Brent-GMM & 0.73 & 0.53 & 0.40 & 0.02 & 0.76 & 0.62 & -17 & 0.53 \\
 & Transformer & 0.70 & \textbf{0.62} & 0.28 & \textbf{0.16} & 0.76 & \textbf{0.81} & \textbf{+7} & 0.51 \\
\midrule
\multirow{9}{*}{\rotatebox{90}{DINOv2, w/ Geo}} & Oracle & 0.93 & 0.86 & 0.80 & 0.60 & 1.00 & 1.00 & +0 & 1.00 \\
 & Tuning-Agglo. & 0.49 & 0.59 & 0.28 & \textbf{0.53} & 0.30 & 0.34 & +212 & 0.51 \\
 & Tuning-Spectral & 0.75 & 0.51 & \textbf{0.50} & 0.00 & 0.75 & 0.50 & -32 & -- \\
 & Tuning-XMeans & 0.75 & 0.51 & \textbf{0.50} & 0.00 & 0.75 & 0.50 & -32 & -- \\
 & Brent-KMeans & 0.75 & 0.64 & 0.42 & 0.24 & 0.73 & 0.66 & -9 & 0.43 \\
 & Brent-Agglo. & 0.76 & 0.61 & 0.45 & 0.17 & 0.75 & 0.64 & -14 & 0.50 \\
 & Brent-Spectral & 0.78 & 0.69 & 0.44 & 0.33 & \textbf{0.78} & \textbf{0.78} & \textbf{+0} & 0.56 \\
 & Brent-GMM & 0.76 & 0.59 & 0.46 & 0.14 & 0.77 & 0.66 & -15 & \textbf{0.59} \\
 & Transformer & \textbf{0.78} & \textbf{0.69} & 0.45 & 0.33 & \textbf{0.78} & \textbf{0.78} & \textbf{+0} & 0.56 \\
\midrule
\multirow{9}{*}{\rotatebox{90}{SelEx, w/o Geo}} & Oracle & 0.83 & 0.67 & 0.59 & 0.18 & 1.00 & 1.00 & +0 & 1.00 \\
 & Tuning-Agglo. & 0.73 & 0.54 & 0.41 & 0.03 & \textbf{0.76} & 0.62 & -17 & \textbf{0.53} \\
 & Tuning-Spectral & 0.75 & 0.51 & 0.47 & -0.00 & 0.74 & 0.52 & -29 & 0.33 \\
 & Tuning-XMeans & \textbf{0.75} & 0.51 & \textbf{0.50} & 0.00 & 0.75 & 0.50 & -32 & -- \\
 & Brent-KMeans & 0.75 & 0.54 & 0.42 & 0.03 & 0.75 & 0.59 & -20 & 0.50 \\
 & Brent-Agglo. & 0.72 & 0.51 & 0.41 & 0.00 & 0.73 & 0.55 & -23 & 0.33 \\
 & Brent-Spectral & 0.72 & 0.62 & 0.32 & 0.12 & \textbf{0.76} & 0.75 & \textbf{-1} & 0.52 \\
 & Brent-GMM & 0.70 & 0.53 & 0.31 & 0.02 & 0.73 & 0.67 & -8 & 0.46 \\
 & Transformer & 0.68 & \textbf{0.66} & 0.15 & \textbf{0.17} & 0.75 & \textbf{0.94} & +24 & 0.50 \\
\midrule
\multirow{9}{*}{\rotatebox{90}{SelEx, w/ Geo}} & Oracle & 0.94 & 0.88 & 0.82 & 0.63 & 1.00 & 1.00 & +0 & 1.00 \\
 & Tuning-Agglo. & 0.51 & 0.62 & 0.30 & \textbf{0.54} & 0.34 & 0.37 & +186 & 0.51 \\
 & Tuning-Spectral & 0.77 & 0.60 & \textbf{0.50} & 0.16 & \textbf{0.79} & 0.66 & -17 & \textbf{0.67} \\
 & Tuning-XMeans & 0.75 & 0.51 & 0.50 & 0.00 & 0.75 & 0.50 & -32 & -- \\
 & Brent-KMeans & 0.77 & 0.62 & 0.49 & 0.21 & 0.77 & 0.66 & -15 & 0.59 \\
 & Brent-Agglo. & 0.75 & 0.61 & 0.42 & 0.18 & 0.73 & 0.64 & -12 & 0.45 \\
 & Brent-Spectral & 0.74 & 0.67 & 0.35 & 0.27 & 0.73 & 0.75 & \textbf{+2} & 0.47 \\
 & Brent-GMM & 0.74 & 0.58 & 0.44 & 0.12 & 0.73 & 0.59 & -18 & 0.43 \\
 & Transformer & \textbf{0.78} & \textbf{0.72} & 0.43 & 0.36 & 0.78 & \textbf{0.81} & +4 & 0.56 \\
\bottomrule
\end{tabular}
}
\label{tab:aves_small_combined_methods}
\vspace{-10pt}
\end{table}

\section{Discussion and Limitations}
\label{sec:limitations}

\noindent\textbf{Real splitting decisions are not strictly visual.}
The underlying assumption of \dataset is that real taxonomic splits provide evidence in image and location data that a model can recover.
In practice, while a candidate split can be hypothesized from visual and geographic cues, genetic evidence is the ultimate arbiter of species delimitation~\cite{rundle2005ecological}.
Methods evaluated on \dataset should therefore only be used to propose candidate splits for downstream expert review and genetic confirmation, not as a substitute for taxonomic decisions.
We do not use genetic data as it is less widely available than images and locations, and is not available for all species. 
We leave this extension for future work. 

\noindent\textbf{Birds only.}
\dataset is restricted to birds to focus an already challenging task on a taxonomic group where sufficiently high quality data is available.
Birds account for 38\% of the iNaturalist taxa with reported splits; are heavily observed, so split information is more reliable; they have a well-defined and stable taxonomy (unlike \eg plants); and using a single taxonomic group enables the learning of fine-grained visual features that generalize across species.
The limitation is that generalization to other taxa is not directly evaluated in \dataset, although our adapted versions of Stanford Cars and FGVC Aircraft provide non-bird evaluations of the same task (Appendix~\ref{sec:additional_results}).
Multimodal biodiversity datasets such as BIOSCAN-5M~\cite{gharaee2024bioscan} could complement \dataset with genetic evidence and taxonomic coverage of very different organisms.

\noindent\textbf{Two-way splits only.}
We restrict to two-way splits and set $K_{\max}=2$ accordingly.
Most recent taxonomic decisions reported on iNaturalist are two-way, but some are not (\eg the giraffe example in \cref{sec:intro} resolved into four species).
The methods we evaluated  could predict $K>2$ but we do not provide examples for $K>2$ for \dataset due to the limited number of species that would be available for training and evaluation.
However, our adapted fine-grained datasets do contain multi-way splits (up to $K=30$; Appendix~\ref{sec:additional_results}). 
Hierarchical splitting across taxonomic levels~\cite{liu2025hyperbolic} is another extension, though most taxonomic splits occur at the species level.

\noindent\textbf{Citizen-science biases.}
\dataset inherits well-documented biases of citizen science data.
Bias towards certain taxonomic groups is one, \ie 
birds are highly observed which simplifies dataset construction, but limits generalization outside of this group.
Observation density correlates with contributor density rather than habitat density --- dense coverage in North America and Western Europe, but sparse in large parts of central Africa and Asia (Appendix~\ref{sec:geo_appendix}).
Splits are also more likely to have sufficient observations when the new species are clearly geographically separable as the new species labels can be automatically assigned to each observation based on location, while less geographically clear splits require per image manual review. 
This may bias \dataset toward geography-dependent cases.
We analyze the regional effect of geographic features in Appendix~\ref{sec:geo_breakdown}.
Our VLM quality filter may also reduce intra-species variation, but we found it necessary given the uncurated iNaturalist images, and did not observe significant biases beyond it being slightly conservative (\cref{fig:nutcracker_vlm}).

\noindent\textbf{Control matching is taxonomic, not ecological or visual.}
Controls are matched at the genus or family level with similar observation counts, but we do not match for ecological niche, habitat type, or photographic difficulty.
Controls are not required to be visually indistinguishable from their splitting species, but being taxonomically close they are typically visually similar, which prevents learning shortcuts such as ``all ducks should be split''.

\noindent\textbf{Splitting only.}
\dataset focuses exclusively on taxonomic splitting.
In practice, in some instances, taxonomic decisions also \emph{merge} previously distinct species, and some of our control species may undergo a future split or merge.
We leave the merging counterpart of this task, and the joint split--merge setting, to future work.

\section{Conclusion}
\label{sec:conclusion}
We introduced \dataset, a fine-grained benchmark for category discovery built from 257 real taxonomic splits documented on iNaturalist, paired with 257 matched non-splitting controls and per-image geographic metadata.
Together with the dataset we formalized \emph{discovery by category splitting}, a task in which a model must decide whether each previously unseen coarse class should be subdivided, into one or more fine classes, and how the data instances should be assigned.
We benchmarked several distinct discovery methods with two no-training references, three K-estimation families, and a lightweight transformer K-estimator we contribute alongside the dataset.
No method we evaluated approaches the oracle upper bound, demonstrating that \dataset is  harder than the fine-grained discovery datasets it was designed to complement.
We hope \dataset and the discovery by splitting task will support sustained progress on more open-ended category discovery methods that are better aligned with challenging real-world problems.

\clearpage 
\newpage 
\noindent{\bf Acknowledgments.} 
We thank the iNaturalist users for sharing observations and images.   
OMA was in part supported by a Royal Society Research Grant.  
JBH was supported by the Pioneer Centre for AI (DNRF grant number P1).

\bibliographystyle{plain}
\bibliography{main}

\begin{thebibliography}{10}

\bibitem{duskythrush}
Boyang An, Danyang Yin, Songlin Huang, Nassoro Mohamed, and Bo~Li.
\newblock Taxonomy of naumann’s and dusky, red-throated and black-throated
  thrush complexes (aves: Turdidae) using morphological and molecular
  characteristics.
\newblock {\em Animal Biology}, 74:213--236, 04 2024.

\bibitem{bai2025qwen25vltechnicalreport}
Shuai Bai, Keqin Chen, Xuejing Liu, Jialin Wang, Wenbin Ge, Sibo Song, Kai
  Dang, Peng Wang, Shijie Wang, Jun Tang, Humen Zhong, Yuanzhi Zhu, Mingkun
  Yang, Zhaohai Li, Jianqiang Wan, Pengfei Wang, Wei Ding, Zheren Fu, Yiheng
  Xu, Jiabo Ye, Xi~Zhang, Tianbao Xie, Zesen Cheng, Hang Zhang, Zhibo Yang,
  Haiyang Xu, and Junyang Lin.
\newblock {Qwen2.5-VL} technical report.
\newblock {\em arXiv:2502.13923}, 2025.

\bibitem{beery2019efficient}
Sara Beery, Dan Morris, and Siyu Yang.
\newblock Efficient pipeline for camera trap image review.
\newblock {\em arXiv:1907.06772}, 2019.

\bibitem{berg2014birdsnap}
Thomas Berg, Jiongxin Liu, Seung~Woo Lee, Michelle~L Alexander, David~W Jacobs,
  and Peter~N Belhumeur.
\newblock Birdsnap: Large-scale fine-grained visual categorization of birds.
\newblock In {\em CVPR}, 2014.

\bibitem{bukchin2021fine}
Guy Bukchin, Eli Schwartz, Kate Saenko, Ori Shahar, Rogerio Feris, Raja Giryes,
  and Leonid Karlinsky.
\newblock Fine-grained angular contrastive learning with coarse labels.
\newblock In {\em CVPR}, 2021.

\bibitem{cao2022open}
Kaidi Cao, Maria Brbic, and Jure Leskovec.
\newblock Open-world semi-supervised learning.
\newblock In {\em ICLR}, 2022.

\bibitem{cao2022pss}
Tianyue Cao, Yongxin Wang, Yifan Xing, Tianjun Xiao, Tong He, Zheng Zhang, Hao
  Zhou, and Joseph Tighe.
\newblock {PSS}: Progressive sample selection for open-world visual
  representation learning.
\newblock In {\em ECCV}, 2022.

\bibitem{caron2021emerging}
Mathilde Caron, Hugo Touvron, Ishan Misra, Herv{\'e} J{\'e}gou, Julien Mairal,
  Piotr Bojanowski, and Armand Joulin.
\newblock Emerging properties in self-supervised vision transformers.
\newblock In {\em ICCV}, 2021.

\bibitem{chen2025open}
Yuyan Chen, Nico Lang, B~Christian Schmidt, Aditya Jain, Yves Basset, Sara
  Beery, Maxim Larriv{\'e}e, and David Rolnick.
\newblock {Open-Insect}: Benchmarking open-set recognition of novel species in
  biodiversity monitoring.
\newblock {\em NeurIPS Datasets and Benchmarks Track}, 2025.

\bibitem{choudhury2024curious}
Subhabrata Choudhury, Iro Laina, Christian Rupprecht, and Andrea Vedaldi.
\newblock The curious layperson: Fine-grained image recognition without expert
  labels.
\newblock {\em IJCV}, 2024.

\bibitem{chu2019geo}
Grace Chu, Brian Potetz, Weijun Wang, Andrew Howard, Yang Song, Fernando
  Brucher, Thomas Leung, and Hartwig Adam.
\newblock Geo-aware networks for fine-grained recognition.
\newblock In {\em ICCV Workshops}, 2019.

\bibitem{CHURCH20253556}
Samuel~H. Church, River~B. Abedon, Namrata Ahuja, Colin~J. Anthony, Dalila
  Destanović, Diego~A. Ramirez, Lourdes~M. Rojas, Maria~E. Albinsson, Itziar
  {Álvarez Trasobares}, Reza~E. Bergemann, Ozren Bogdanovic, et~al.
\newblock Population genomics of a sailing siphonophore reveals genetic
  structure in the open ocean.
\newblock {\em Current Biology}, 2025.

\bibitem{Clements2024eBirdClements}
J.~F. Clements, P.~C. Rasmussen, T.~S. Schulenberg, M.~J. Iliff, T.~A.
  Fredericks, J.~A. Gerbracht, D.~Lepage, A.~Spencer, S.~M. Billerman, B.~L.
  Sullivan, M.~Smith, and C.~L. Wood.
\newblock The {eBird/Clements} checklist of {Birds of the World}: v2024.
\newblock \url{https://www.birds.cornell.edu/clementschecklist/download/},
  2024.

\bibitem{coleICML2023}
Elijah Cole, Grant Van~Horn, Christian Lange, Alexander Shepard, Patrick Leary,
  Pietro Perona, Scott Loarie, and Oisin Mac~Aodha.
\newblock Spatial implicit neural representations for global-scale species
  mapping.
\newblock In {\em ICML}, 2023.

\bibitem{darcet2024vision}
Timoth{\'e}e Darcet, Maxime Oquab, Julien Mairal, and Piotr Bojanowski.
\newblock Vision transformers need registers.
\newblock In {\em ICLR}, 2024.

\bibitem{du2023fly}
Ruoyi Du, Dongliang Chang, Kongming Liang, Timothy Hospedales, Yi-Zhe Song, and
  Zhanyu Ma.
\newblock On-the-fly category discovery.
\newblock In {\em CVPR}, 2023.

\bibitem{fennessy2016multi}
Julian Fennessy, Tobias Bidon, Friederike Reuss, Vikas Kumar, Paul Elkan,
  Maria~A Nilsson, Melita Vamberger, Uwe Fritz, and Axel Janke.
\newblock Multi-locus analyses reveal four giraffe species instead of one.
\newblock {\em Current Biology}, 2016.

\bibitem{fini2021unified}
Enrico Fini, Enver Sangineto, St{\'e}phane Lathuili{\`e}re, Zhun Zhong, Moin
  Nabi, and Elisa Ricci.
\newblock A unified objective for novel class discovery.
\newblock In {\em ICCV}, 2021.

\bibitem{gharaee2024bioscan}
Zahra Gharaee, Scott~C. Lowe, ZeMing Gong, Pablo Millan~Arias, Nicholas
  Pellegrino, Austin~T. Wang, Joakim~Bruslund Haurum, Iuliia Zarubiieva, Lila
  Kari, Dirk Steinke, Graham~W. Taylor, Paul Fieguth, and Angel~X. Chang.
\newblock {BIOSCAN-5M}: A multimodal dataset for insect biodiversity.
\newblock In {\em NeurIPS Datasets and Benchmarks Track}, 2024.

\bibitem{han2021autonovel}
Kai Han, Sylvestre-Alvise Rebuffi, Sebastien Ehrhardt, Andrea Vedaldi, and
  Andrew Zisserman.
\newblock {AutoNovel}: Automatically discovering and learning novel visual
  categories.
\newblock {\em TPAMI}, 2022.

\bibitem{han2019learning}
Kai Han, Andrea Vedaldi, and Andrew Zisserman.
\newblock Learning to discover novel visual categories via deep transfer
  clustering.
\newblock In {\em ICCV}, 2019.

\bibitem{he2025category}
Zhenqi He, Yuanpei Liu, and Kai Han.
\newblock Category discovery: An open-world perspective.
\newblock {\em arXiv:2509.22542}, 2025.

\bibitem{huh2016makes}
Minyoung Huh, Pulkit Agrawal, and Alexei~A Efros.
\newblock What makes {ImageNet} good for transfer learning?
\newblock {\em arXiv:1608.08614}, 2016.

\bibitem{iNaturalist}
{iNaturalist}.
\newblock inaturalist, 2026.
\newblock Accessed on 2026-05-05.

\bibitem{iNaturalistSplit}
{iNaturalist}.
\newblock inaturalist taxon changes, 2026.
\newblock Accessed on 2026-05-05.

\bibitem{krause20133d}
Jonathan Krause, Michael Stark, Jia Deng, and Li~Fei-Fei.
\newblock {3D} object representations for fine-grained categorization.
\newblock In {\em ICCV Workshops}, 2013.

\bibitem{krizhevsky2009learning}
Alex Krizhevsky.
\newblock Learning multiple layers of features from tiny images.
\newblock Technical report, University of Toronto, 2009.

\bibitem{pmlr-v97-lee19d}
Juho Lee, Yoonho Lee, Jungtaek Kim, Adam Kosiorek, Seungjin Choi, and Yee~Whye
  Teh.
\newblock {Set Transformer}: A framework for attention-based
  permutation-invariant neural networks.
\newblock In {\em ICML}, 2019.

\bibitem{liu2026letssplitupzeroshot}
Kaiting Liu and Hazel Doughty.
\newblock Let's split up: Zero-shot classifier edits for fine-grained video
  understanding.
\newblock In {\em ICLR}, 2026.

\bibitem{liu2025hyperbolic}
Yuanpei Liu, Zhenqi He, and Kai Han.
\newblock Hyperbolic category discovery.
\newblock In {\em CVPR}, 2025.

\bibitem{macaodhaiccv19}
Oisin Mac~Aodha, Elijah Cole, and Pietro Perona.
\newblock Presence-only geographical priors for fine-grained image
  classification.
\newblock In {\em ICCV}, 2019.

\bibitem{maji2013fine}
Subhransu Maji, Esa Rahtu, Juho Kannala, Matthew Blaschko, and Andrea Vedaldi.
\newblock Fine-grained visual classification of aircraft.
\newblock {\em arXiv:1306.5151}, 2013.

\bibitem{manolache2025crypticbio}
Georgiana Manolache, Gerard Schouten, and Joaquin Vanschoren.
\newblock {CrypticBio}: A large multimodal dataset for visually confusing
  species.
\newblock In {\em NeurIPS}, 2025.

\bibitem{mora2011many}
Camilo Mora, Derek~P Tittensor, Sina Adl, Alastair~GB Simpson, and Boris Worm.
\newblock How many species are there on earth and in the ocean?
\newblock {\em PLoS Biology}, 2011.

\bibitem{ng2001spectral}
Andrew Ng, Michael Jordan, and Yair Weiss.
\newblock On spectral clustering: Analysis and an algorithm.
\newblock {\em NeurIPS}, 2001.

\bibitem{oquab2023dinov2}
Maxime Oquab, Timoth{\'e}e Darcet, Th{\'e}o Moutakanni, Huy Vo, Marc
  Szafraniec, Vasil Khalidov, Pierre Fernandez, Daniel Haziza, Francisco Massa,
  Alaaeldin El-Nouby, et~al.
\newblock {DINOv2}: Learning robust visual features without supervision.
\newblock {\em TMLR}, 2024.

\bibitem{otholt2024guided}
Jona Otholt, Christoph Meinel, and Haojin Yang.
\newblock Guided cluster aggregation: A hierarchical approach to generalized
  category discovery.
\newblock In {\em WACV}, 2024.

\bibitem{pelleg2000x}
Dan Pelleg and Andrew~W Moore.
\newblock {X-means}: Extending {k-means} with efficient estimation of the
  number of clusters.
\newblock In {\em ICML}, 2000.

\bibitem{perez2023human}
Gustavo Perez, Daniel Sheldon, Grant Van~Horn, and Subhransu Maji.
\newblock Human-in-the-loop visual re-id for population size estimation.
\newblock In {\em ECCV}, 2024.

\bibitem{rastegar2024selexselfexpertisefinegrainedgeneralized}
Sarah Rastegar, Mohammadreza Salehi, Yuki~M Asano, Hazel Doughty, and Cees~GM
  Snoek.
\newblock {SelEx}: Self-expertise in fine-grained generalized category
  discovery.
\newblock In {\em ECCV}, 2024.

\bibitem{ristin2015categories}
Marko Ristin, Juergen Gall, Matthieu Guillaumin, and Luc Van~Gool.
\newblock From categories to subcategories: large-scale image classification
  with partial class label refinement.
\newblock In {\em CVPR}, 2015.

\bibitem{rundle2005ecological}
Howard~D Rundle and Patrik Nosil.
\newblock Ecological speciation.
\newblock {\em Ecology Letters}, 2005.

\bibitem{russwurm2024geographic}
Marc Ru{\ss}wurm, Konstantin Klemmer, Esther Rolf, Robin Zbinden, and Devis
  Tuia.
\newblock Geographic location encoding with spherical harmonics and sinusoidal
  representation networks.
\newblock In {\em ICLR}, 2024.

\bibitem{su2021semi}
Jong-Chyi Su and Subhransu Maji.
\newblock Semi-supervised learning with taxonomic labels.
\newblock In {\em BMVC}, 2021.

\bibitem{taherkhani2019weakly}
Fariborz Taherkhani, Hadi Kazemi, Ali Dabouei, Jeremy Dawson, and Nasser~M
  Nasrabadi.
\newblock A weakly supervised fine label classifier enhanced by coarse
  supervision.
\newblock In {\em ICCV}, 2019.

\bibitem{tan2019herbarium}
Kiat~Chuan Tan, Yulong Liu, Barbara Ambrose, Melissa Tulig, and Serge Belongie.
\newblock The {Herbarium Challenge} 2019 dataset.
\newblock {\em arXiv:1906.05372}, 2019.

\bibitem{tang2015improving}
Kevin Tang, Manohar Paluri, Li~Fei-Fei, Rob Fergus, and Lubomir Bourdev.
\newblock Improving image classification with location context.
\newblock In {\em ICCV}, 2015.

\bibitem{touvron2021grafit}
Hugo Touvron, Alexandre Sablayrolles, Matthijs Douze, Matthieu Cord, and
  Herv{\'e} J{\'e}gou.
\newblock Grafit: Learning fine-grained image representations with coarse
  labels.
\newblock In {\em ICCV}, 2021.

\bibitem{van2021benchmarking}
Grant Van~Horn, Elijah Cole, Sara Beery, Kimberly Wilber, Serge Belongie, and
  Oisin Mac~Aodha.
\newblock Benchmarking representation learning for natural world image
  collections.
\newblock In {\em CVPR}, 2021.

\bibitem{van2018inaturalist}
Grant Van~Horn, Oisin Mac~Aodha, Yang Song, Yin Cui, Chen Sun, Alex Shepard,
  Hartwig Adam, Pietro Perona, and Serge Belongie.
\newblock The {iNaturalist} species classification and detection dataset.
\newblock In {\em CVPR}, 2018.

\bibitem{vaswani2017attention}
Ashish Vaswani, Noam Shazeer, Niki Parmar, Jakob Uszkoreit, Llion Jones,
  Aidan~N Gomez, {\L}ukasz Kaiser, and Illia Polosukhin.
\newblock Attention is all you need.
\newblock {\em NeurIPS}, 2017.

\bibitem{vaze22generalized}
Sagar Vaze, Kai Han, Andrea Vedaldi, and Andrew Zisserman.
\newblock Generalized category discovery.
\newblock In {\em CVPR}, 2022.

\bibitem{tutorial_spectral_eigengap}
Ulrike von Luxburg.
\newblock A tutorial on spectral clustering.
\newblock {\em Statistics and Computing}, 2007.

\bibitem{wah2011caltech}
Catherine Wah, Steve Branson, Peter Welinder, Pietro Perona, and Serge
  Belongie.
\newblock {The {Caltech-UCSD Birds-200-2011} Dataset}.
\newblock Technical Report CNS-TR-2011-001, California Institute of Technology,
  2011.

\bibitem{wen2023simple}
Xin Wen, Bingchen Zhao, and Xiaojuan Qi.
\newblock Parametric classification for generalized category discovery: A
  baseline study.
\newblock In {\em ICCV}, 2023.

\bibitem{wu2018improving}
Zhirong Wu, Alexei~A Efros, and Stella~X Yu.
\newblock Improving generalization via scalable neighborhood component
  analysis.
\newblock In {\em ECCV}, 2018.

\bibitem{xing2021learning}
Yifan Xing, Tong He, Tianjun Xiao, Yongxin Wang, Yuanjun Xiong, Wei Xia, David
  Wipf, Zheng Zhang, and Stefano Soatto.
\newblock Learning hierarchical graph neural networks for image clustering.
\newblock In {\em ICCV}, 2021.

\bibitem{xu2023hyperbolic}
Shu-Lin Xu, Yifan Sun, Faen Zhang, Anqi Xu, Xiu-Shen Wei, and Yi~Yang.
\newblock Hyperbolic space with hierarchical margin boosts fine-grained
  learning from coarse labels.
\newblock In {\em NeurIPS}, 2023.

\bibitem{yang2022dynamic}
Lingfeng Yang, Xiang Li, Renjie Song, Borui Zhao, Juntian Tao, Shihao Zhou,
  Jiajun Liang, and Jian Yang.
\newblock Dynamic {MLP} for fine-grained image classification by leveraging
  geographical and temporal information.
\newblock In {\em CVPR}, 2022.

\bibitem{yu2021benchmark}
Xiaohan Yu, Yang Zhao, Yongsheng Gao, Xiaohui Yuan, and Shengwu Xiong.
\newblock Benchmark platform for ultra-fine-grained visual categorization
  beyond human performance.
\newblock In {\em ICCV}, 2021.

\bibitem{zhang2022promptcal}
Sheng Zhang, Salman Khan, Zhiqiang Shen, Muzammal Naseer, Guangyi Chen, and
  Fahad~Shahbaz Khan.
\newblock Promptcal: Contrastive affinity learning via auxiliary prompts for
  generalized novel category discovery.
\newblock In {\em CVPR}, 2023.

\bibitem{zhang2022grow}
Xinwei Zhang, Jianwen Jiang, Yutong Feng, Zhi-Fan Wu, Xibin Zhao, Hai Wan,
  Mingqian Tang, Rong Jin, and Yue Gao.
\newblock Grow and merge: A unified framework for continuous categories
  discovery.
\newblock In {\em NeurIPS}, 2022.

\bibitem{zhao2023incremental}
Bingchen Zhao and Oisin Mac~Aodha.
\newblock Incremental generalized category discovery.
\newblock In {\em ICCV}, 2023.

\bibitem{zhao2023learningsemisupervisedgaussianmixture}
Bingchen Zhao, Xin Wen, and Kai Han.
\newblock Learning semi-supervised {Gaussian} mixture models for generalized
  category discovery.
\newblock In {\em ICCV}, 2023.

\end{thebibliography}

\newpage
\appendix
\setcounter{table}{0}   
\renewcommand{\thetable}{A\arabic{table}}
\setcounter{figure}{0}
\renewcommand{\thefigure}{A\arabic{figure}}
{\LARGE Appendix}

\section{Additional Discovery by Category Splitting Results}
\label{sec:additional_results}

This appendix reports the backbone, K-estimation, and clustering comparisons used in \cref{sec:exps} on (i) the full \dataset, and (ii) three adapted fine-grained datasets that have no geographic metadata.
The full-\dataset results in \cref{tab:aves_full_combined_backbone,tab:aves_full_combined_methods} extend the small-subset comparisons of \cref{tab:aves_small_combined_backbone,tab:aves_small_combined_methods}; the FG-dataset results in \cref{tab:fg_combined_backbone,tab:fg_combined_backbone_split} characterise how \dataset compares to existing fine-grained category-discovery benchmarks under the same evaluation protocol.

\noindent\textbf{Backbone comparison on the full dataset.}
The eight-backbone comparison on the full \dataset (\cref{tab:aves_full_combined_backbone}) mirrors the small-subset patterns of \cref{tab:aves_small_combined_backbone}: visual-only backbones cluster within a narrow band (Cl.\,Acc.\ $0.65$--$0.80$ across both K-estimators), and concatenating SINR geographic features improves all metrics under both K-estimators.
Two findings deserve highlighting on the full dataset.
First, the SINR-only baseline is among the strongest configurations regardless of K-estimator (Cl.\,Acc.\ $0.81$ under Brent-KMeans, $0.79$ under the transformer; Cl.\,Acc.$_s$ $0.71$ and $0.79$ respectively), reinforcing the conclusion that geography alone carries a portion of the distinguishing signal in \dataset, possibly as many images on iNaturalist are relabeled after a taxonomic split by the geographic region that the image is from, ensuring geographic information is useful for this task.
Second, several visual-only backbones under Brent + KMeans collapse close to a ``never split'' regime.
UNO+ is the most extreme: it reaches Cl.\,Acc.\ $0.80$ purely by under-predicting splits ($\Delta\!\sum\!K = -124$, with predicted total $K$ only marginally above the no-split lower bound), and the handful of splits it does predict are all incorrect (Split Prec.\ $0.00$).
SimGCD shows the same pattern in milder form ($\Delta\!\sum\!K = -118$, Split Prec.\ $0.45$).
This is a reminder that a high all-class Cl.\,Acc.\ on \dataset must be read alongside Split Precision and $\Delta\!\sum\!K$: a method that always declares ``no split'' already scores Cl.\,Acc.\ $0.80$ on the full dataset (Never-Split in \cref{tab:aves_full_combined_methods}) at no cost, and any improvement beyond that can come either from genuine discrimination or from confident no-split calls on easy controls.
Visual information does still provide a signal for category splitting, and improved backbones and approaches to this task may be able to extract more signal from vision alone, allowing \dataset to be a useful benchmark for future discovery by category splitting approaches.

\begin{table}[H]
\centering
\caption{\textbf{Discovery by Category splitting on PS-SPLIT.} Top: Brent + KMeans K estimation. Bottom: transformer K estimation.}
\vspace{3pt}
\resizebox{\linewidth}{!}{
\begin{tabular}{c|l|cccc|cc|cc}
\toprule
 & Backbone & Cl.\,Acc. & Cl.\,Acc.$_\text{s}$ & AMI & AMI$_\text{s}$ & K Acc. & K Acc.$_\text{s}$ & $\Delta\!\sum\!K$ & Split Prec. \\
\midrule
\multirow{8}{*}{\rotatebox{90}{Brent-KMeans}} & DINOv1 & 0.75 & 0.60 & 0.35 & 0.00 & 0.74 & 0.64 & -52 & 0.48 \\
 & DINOv2 & 0.77 & 0.64 & 0.40 & 0.06 & \textbf{0.78} & 0.69 & -47 & \textbf{0.59} \\
 & UNO+ & 0.80 & 0.61 & 0.48 & 0.00 & 0.74 & 0.50 & -124 & 0.00 \\
 & GCD & 0.77 & 0.67 & 0.39 & 0.14 & 0.77 & \textbf{0.72} & \textbf{-25} & 0.56 \\
 & SimGCD & 0.80 & 0.62 & \textbf{0.48} & 0.01 & 0.75 & 0.52 & -118 & 0.45 \\
 & SelEx & 0.77 & 0.64 & 0.39 & 0.08 & 0.77 & 0.69 & -42 & 0.55 \\
 & SINR & \textbf{0.81} & \textbf{0.71} & 0.47 & \textbf{0.20} & 0.77 & 0.67 & -52 & 0.56 \\
 & DINOv2 SINR & 0.81 & 0.67 & 0.47 & 0.14 & 0.76 & 0.62 & -72 & 0.56 \\
\midrule
\multirow{8}{*}{\rotatebox{90}{Transformer}} & DINOv1 & 0.65 & 0.59 & 0.13 & 0.04 & 0.77 & 0.92 & +79 & 0.52 \\
 & DINOv2 & 0.71 & 0.65 & 0.22 & 0.11 & 0.77 & 0.88 & +57 & 0.53 \\
 & UNO+ & 0.80 & 0.61 & \textbf{0.48} & 0.00 & 0.75 & 0.52 & -118 & 0.45 \\
 & GCD & 0.72 & 0.70 & 0.25 & 0.20 & 0.78 & 0.91 & +70 & 0.54 \\
 & SimGCD & 0.67 & 0.66 & 0.14 & 0.14 & 0.75 & 0.93 & +93 & 0.50 \\
 & SelEx & 0.69 & 0.69 & 0.12 & 0.20 & 0.75 & \textbf{0.98} & +118 & 0.50 \\
 & SINR & 0.79 & \textbf{0.79} & 0.43 & \textbf{0.46} & 0.80 & 0.90 & +52 & 0.57 \\
 & DINOv2 SINR & \textbf{0.80} & 0.77 & 0.45 & 0.41 & \textbf{0.81} & 0.86 & \textbf{+30} & \textbf{0.59} \\
\bottomrule
\end{tabular}
}
\label{tab:aves_full_combined_backbone}
\end{table}

\noindent\textbf{Method ablation on the full dataset.}
The four-block per-method ablation (\cref{tab:aves_full_combined_methods}) extends \cref{tab:aves_small_combined_methods} to the full dataset, and also includes the upper and lower bounds described below.
Three patterns are stable across both dataset sizes:
\textit{(i)} Tuning-XMeans degenerates to ``never split'' on every backbone setup;
\textit{(ii)} Tuning-Agglo over-splits dramatically whenever geographic features are added (\eg $\Delta\!\sum\!K = +534$ on DINOv2 and $+501$ on SelEx on the full dataset, the largest absolute deviations in either methods table);
\textit{(iii)} the transformer K-estimator obtains the highest split-only metrics (Cl.\,Acc.$_s$, K Acc.$_s$) across most rows, at the cost of over-predicting splits when only visual features are available.
Two findings are clearer on the full dataset than on the small subset.
First, on SelEx with geographic features the transformer is the only row in either methods table where $\Delta\!\sum\!K$ becomes \emph{negative} ($-12$); this is a large swing from its same-backbone visual-only value of $+118$ and shows that geography can reverse its over-splitting tendency.
Second, Brent-Agglo on SelEx with geographic features achieves the highest non-oracle Cl.\,Acc.\ ($0.82$) anywhere in either methods table and one of the highest Split Precision values ($0.69$); under our protocol it is the strongest single-method result we observe on the full dataset.
The remaining gap to the SelEx + geo oracle ($0.93$) is roughly eleven points, slightly smaller than on the small subset.

\noindent\textbf{Upper and lower bounds.}
\label{sec:lower_bounds}
\label{sec:supervised}
\Cref{tab:aves_small_methods_bounds} expands \cref{tab:aves_small_combined_methods} with upper and lower bounds for each backbone setting on \dataset-Small, and \cref{tab:aves_full_combined_methods} does the same for the full dataset.

\noindent\textbf{Lower bounds.}
Never-Split predicts $\hat K=1$ for every coarse class, and Always-Split predicts $\hat K=2$ and runs KMeans, and Random+KMeans flips a fair coin per coarse class to decide $\hat K \in \{1,2\}$ and runs KMeans if split.
Never-Split achieves a high Cl.\,Acc.\ ($0.75$) and AMI ($0.50$) because it is perfect on every non-splitting control, which is why multiple metrics are needed: simple baselines can perform well on one metric while being useless for discovery by category splitting. The metrics that expose this are the split-only variants and Split Precision, \ie Never-Split has AMI$_s=0.00$ and undefined Split Precision as it never predicts a split.
Always-Split obtains the same split-only Cl.\,Acc.\ and AMI as the oracle, as both perform KMeans with the correct $K=2$ on every splitting class, but it predicts a split for every control ($\Delta\!\sum\!K=+32$ on \dataset-Small, Split Precision $0.50$).

\noindent\textbf{Supervised upper bound.}
To investigate the potential of vision-only features, the supervised probe rows report a separate logistic-regression classifier for each novel coarse class, trained with the fine labels on a stratified 70\% of the training-pool images of each novel coarse class and evaluated on its test-pool images (mean over 5 seeds). These probes use privileged supervised information that is not available in the discovery by category splitting setting, and non-splitting species achieve perfect accuracy as there is only a single fine class to predict. 
As the supervised probe is given the true $K^*$, like the Oracle, its K-estimation metrics are perfect by construction.
Even without geographic information or backbone fine-tuning there is enough visual signal to distinguish between the species in most splits (Cl.\,Acc.$_s$ $0.83$ with DINOv2 on \dataset-Small and $0.80$ on the full dataset), and adding location features makes the supervised problem almost trivial on \dataset-Small (Cl.\,Acc.$_s$ $0.98$; $0.92$--$0.93$ on the full dataset). In contrast, in the discovery setting, where the number of fine classes is not known and no fine labels are provided, the best methods remain far from these values, suggesting significant room for improvement as new methods are developed.

\begin{table}[H]
\centering
\caption{\textbf{Category splitting results on \dataset{}-Small, with upper and lower bounds.} Expanded version of \cref{tab:aves_small_combined_methods}. Each section lists two upper bounds, three lower bounds, and then the K-estimation / clustering methods; see Appendix~\ref{sec:lower_bounds}. Both upper bounds use privileged information that is not available to discovery methods. The Oracle is given the true number of fine classes $K^*$ of every coarse class and clusters its images with KMeans. The supervised probe is given $K^*$ and the fine labels of 70\% of each novel coarse class's training-pool images, and classifies the test-pool images with a logistic-regression classifier. Because both are given $K^*$, their K Acc., $\Delta\!\sum\!K$ and Split Prec.\ are perfect by construction. Random+KMeans is averaged over 200 draws. Best value per column among the methods is bolded within each section, bounds excluded. True $\sum K^*=96$.}
\label{tab:aves_small_methods_bounds}
\vspace{3pt}
\begin{adjustbox}{max width=\linewidth, max totalheight=0.80\textheight}
\begin{tabular}{c|l|cccc|cc|cc}
\toprule
 & Method & Cl.\,Acc. & Cl.\,Acc.$_\text{s}$ & AMI & AMI$_\text{s}$ & K Acc. & K Acc.$_\text{s}$ & $\Delta\!\sum\!K$ & Split Prec. \\
\midrule
\multirow{13}{*}{\rotatebox{90}{DINOv2, w/o Geo}} & Oracle & 0.83 & 0.66 & 0.60 & 0.20 & 1.00 & 1.00 & +0 & 1.00 \\
 & Supervised probe & 0.91 & 0.83 & 0.71 & 0.43 & 1.00 & 1.00 & +0 & 1.00 \\
\cmidrule(l){2-10}
 & Never-Split & 0.75 & 0.51 & 0.50 & 0.00 & 0.75 & 0.50 & -32 & -- \\
 & Always-Split & 0.65 & 0.66 & 0.10 & 0.20 & 0.75 & 1.00 & +32 & 0.50 \\
 & Random+KMeans & 0.70 & 0.58 & 0.30 & 0.10 & 0.75 & 0.75 & -0.1 & 0.50 \\
\cmidrule(l){2-10}
 & Tuning-Agglo. & 0.59 & 0.58 & 0.12 & 0.15 & 0.56 & 0.67 & +74 & 0.51 \\
 & Tuning-Spectral & 0.70 & 0.55 & 0.32 & 0.05 & 0.71 & 0.62 & -11 & 0.38 \\
 & Tuning-XMeans & 0.75 & 0.51 & \textbf{0.50} & 0.00 & 0.75 & 0.50 & -32 & -- \\
 & Brent-KMeans & \textbf{0.77} & 0.57 & 0.49 & 0.07 & \textbf{0.82} & 0.69 & -17 & \textbf{0.80} \\
 & Brent-Agglo. & 0.74 & 0.53 & 0.43 & 0.02 & 0.76 & 0.59 & -21 & 0.55 \\
 & Brent-Spectral & 0.69 & 0.55 & 0.31 & 0.06 & 0.70 & 0.61 & -11 & 0.33 \\
 & Brent-GMM & 0.73 & 0.53 & 0.40 & 0.02 & 0.76 & 0.62 & -17 & 0.53 \\
 & Transformer & 0.70 & \textbf{0.62} & 0.28 & \textbf{0.16} & 0.76 & \textbf{0.81} & \textbf{+7} & 0.51 \\
\midrule
\multirow{13}{*}{\rotatebox{90}{DINOv2, w/ Geo}} & Oracle & 0.93 & 0.86 & 0.80 & 0.60 & 1.00 & 1.00 & +0 & 1.00 \\
 & Supervised probe & 0.99 & 0.98 & 0.95 & 0.90 & 1.00 & 1.00 & +0 & 1.00 \\
\cmidrule(l){2-10}
 & Never-Split & 0.75 & 0.51 & 0.50 & 0.00 & 0.75 & 0.50 & -32 & -- \\
 & Always-Split & 0.76 & 0.86 & 0.30 & 0.60 & 0.75 & 1.00 & +32 & 0.50 \\
 & Random+KMeans & 0.75 & 0.68 & 0.40 & 0.30 & 0.75 & 0.75 & -0.1 & 0.50 \\
\cmidrule(l){2-10}
 & Tuning-Agglo. & 0.49 & 0.59 & 0.28 & \textbf{0.53} & 0.30 & 0.34 & +212 & 0.51 \\
 & Tuning-Spectral & 0.75 & 0.51 & \textbf{0.50} & 0.00 & 0.75 & 0.50 & -32 & -- \\
 & Tuning-XMeans & 0.75 & 0.51 & \textbf{0.50} & 0.00 & 0.75 & 0.50 & -32 & -- \\
 & Brent-KMeans & 0.75 & 0.64 & 0.42 & 0.24 & 0.73 & 0.66 & -9 & 0.43 \\
 & Brent-Agglo. & 0.76 & 0.61 & 0.45 & 0.17 & 0.75 & 0.64 & -14 & 0.50 \\
 & Brent-Spectral & 0.78 & 0.69 & 0.44 & 0.33 & \textbf{0.78} & \textbf{0.78} & \textbf{+0} & 0.56 \\
 & Brent-GMM & 0.76 & 0.59 & 0.46 & 0.14 & 0.77 & 0.66 & -15 & \textbf{0.59} \\
 & Transformer & \textbf{0.78} & \textbf{0.69} & 0.45 & 0.33 & \textbf{0.78} & \textbf{0.78} & \textbf{+0} & 0.56 \\
\midrule
\multirow{13}{*}{\rotatebox{90}{SelEx, w/o Geo}} & Oracle & 0.83 & 0.67 & 0.59 & 0.18 & 1.00 & 1.00 & +0 & 1.00 \\
 & Supervised probe & 0.91 & 0.81 & 0.69 & 0.39 & 1.00 & 1.00 & +0 & 1.00 \\
\cmidrule(l){2-10}
 & Never-Split & 0.75 & 0.51 & 0.50 & 0.00 & 0.75 & 0.50 & -32 & -- \\
 & Always-Split & 0.67 & 0.67 & 0.09 & 0.18 & 0.75 & 1.00 & +32 & 0.50 \\
 & Random+KMeans & 0.71 & 0.59 & 0.29 & 0.09 & 0.75 & 0.75 & -0.1 & 0.50 \\
\cmidrule(l){2-10}
 & Tuning-Agglo. & 0.73 & 0.54 & 0.41 & 0.03 & \textbf{0.76} & 0.62 & -17 & \textbf{0.53} \\
 & Tuning-Spectral & 0.75 & 0.51 & 0.47 & -0.00 & 0.74 & 0.52 & -29 & 0.33 \\
 & Tuning-XMeans & \textbf{0.75} & 0.51 & \textbf{0.50} & 0.00 & 0.75 & 0.50 & -32 & -- \\
 & Brent-KMeans & 0.75 & 0.54 & 0.42 & 0.03 & 0.75 & 0.59 & -20 & 0.50 \\
 & Brent-Agglo. & 0.72 & 0.51 & 0.41 & 0.00 & 0.73 & 0.55 & -23 & 0.33 \\
 & Brent-Spectral & 0.72 & 0.62 & 0.32 & 0.12 & \textbf{0.76} & 0.75 & \textbf{-1} & 0.52 \\
 & Brent-GMM & 0.70 & 0.53 & 0.31 & 0.02 & 0.73 & 0.67 & -8 & 0.46 \\
 & Transformer & 0.68 & \textbf{0.66} & 0.15 & \textbf{0.17} & 0.75 & \textbf{0.94} & +24 & 0.50 \\
\midrule
\multirow{13}{*}{\rotatebox{90}{SelEx, w/ Geo}} & Oracle & 0.94 & 0.88 & 0.82 & 0.63 & 1.00 & 1.00 & +0 & 1.00 \\
 & Supervised probe & 0.99 & 0.98 & 0.95 & 0.90 & 1.00 & 1.00 & +0 & 1.00 \\
\cmidrule(l){2-10}
 & Never-Split & 0.75 & 0.51 & 0.50 & 0.00 & 0.75 & 0.50 & -32 & -- \\
 & Always-Split & 0.76 & 0.88 & 0.32 & 0.63 & 0.75 & 1.00 & +32 & 0.50 \\
 & Random+KMeans & 0.76 & 0.69 & 0.41 & 0.31 & 0.75 & 0.75 & -0.1 & 0.50 \\
\cmidrule(l){2-10}
 & Tuning-Agglo. & 0.51 & 0.62 & 0.30 & \textbf{0.54} & 0.34 & 0.37 & +186 & 0.51 \\
 & Tuning-Spectral & 0.77 & 0.60 & \textbf{0.50} & 0.16 & \textbf{0.79} & 0.66 & -17 & \textbf{0.67} \\
 & Tuning-XMeans & 0.75 & 0.51 & 0.50 & 0.00 & 0.75 & 0.50 & -32 & -- \\
 & Brent-KMeans & 0.77 & 0.62 & 0.49 & 0.21 & 0.77 & 0.66 & -15 & 0.59 \\
 & Brent-Agglo. & 0.75 & 0.61 & 0.42 & 0.18 & 0.73 & 0.64 & -12 & 0.45 \\
 & Brent-Spectral & 0.74 & 0.67 & 0.35 & 0.27 & 0.73 & 0.75 & \textbf{+2} & 0.47 \\
 & Brent-GMM & 0.74 & 0.58 & 0.44 & 0.12 & 0.73 & 0.59 & -18 & 0.43 \\
 & Transformer & \textbf{0.78} & \textbf{0.72} & 0.43 & 0.36 & 0.78 & \textbf{0.81} & +4 & 0.56 \\
\bottomrule
\end{tabular}
\end{adjustbox}
\end{table}

\begin{table}[H]
\centering
\caption{\textbf{Category splitting on \dataset{}-Full, with upper and lower bounds.} Same layout, bounds and bolding as \cref{tab:aves_small_methods_bounds}. The Oracle and the supervised probe are upper bounds that use privileged information (the true $K^*$, and for the probe also fine labels), so their K-estimation metrics are perfect by construction. True $\sum K^* = 387$.}
\vspace{3pt}
\begin{adjustbox}{max width=\linewidth, max totalheight=0.88\textheight}
\begin{tabular}{c|l|cccc|cc|cc}
\toprule
 & Method & Cl.\,Acc. & Cl.\,Acc.$_\text{s}$ & AMI & AMI$_\text{s}$ & K Acc. & K Acc.$_\text{s}$ & $\Delta\!\sum\!K$ & Split Prec. \\
\midrule
\multirow{13}{*}{\rotatebox{90}{DINOv2, w/o Geo}} & Oracle & 0.83 & 0.66 & 0.57 & 0.14 & 1.00 & 1.00 & +0 & 1.00 \\
 & Supervised probe & 0.90 & 0.80 & 0.63 & 0.27 & 1.00 & 1.00 & +0 & 1.00 \\
\cmidrule(l){2-10}
 & Never-Split & 0.80 & 0.61 & 0.50 & 0.00 & 0.75 & 0.50 & -129 & -- \\
 & Always-Split & 0.66 & 0.66 & 0.07 & 0.14 & 0.75 & 1.00 & +129 & 0.50 \\
 & Random+KMeans & 0.73 & 0.63 & 0.28 & 0.07 & 0.75 & 0.75 & -0.1 & 0.50 \\
\cmidrule(l){2-10}
 & Tuning-Agglo. & 0.67 & 0.62 & 0.20 & \textbf{0.12} & 0.62 & 0.68 & +180 & 0.51 \\
 & Tuning-Spectral & 0.78 & 0.63 & 0.41 & 0.04 & 0.75 & 0.61 & -74 & 0.51 \\
 & Tuning-XMeans & \textbf{0.80} & 0.61 & \textbf{0.50} & 0.00 & 0.75 & 0.50 & -129 & -- \\
 & Brent-KMeans & 0.77 & 0.64 & 0.40 & 0.06 & \textbf{0.78} & 0.69 & -47 & \textbf{0.59} \\
 & Brent-Agglo. & 0.76 & 0.61 & 0.38 & 0.04 & 0.75 & 0.64 & -57 & 0.50 \\
 & Brent-Spectral & 0.75 & 0.63 & 0.37 & 0.06 & 0.76 & 0.68 & -40 & 0.53 \\
 & Brent-GMM & 0.76 & 0.62 & 0.33 & 0.03 & 0.75 & 0.69 & \textbf{-34} & 0.51 \\
 & Transformer & 0.71 & \textbf{0.65} & 0.22 & 0.11 & 0.77 & \textbf{0.88} & +57 & 0.53 \\
\midrule
\multirow{13}{*}{\rotatebox{90}{DINOv2, w/ Geo}} & Oracle & 0.93 & 0.85 & 0.80 & 0.59 & 1.00 & 1.00 & +0 & 1.00 \\
 & Supervised probe & 0.96 & 0.92 & 0.83 & 0.66 & 1.00 & 1.00 & +0 & 1.00 \\
\cmidrule(l){2-10}
 & Never-Split & 0.80 & 0.61 & 0.50 & 0.00 & 0.75 & 0.50 & -129 & -- \\
 & Always-Split & 0.76 & 0.85 & 0.30 & 0.59 & 0.75 & 1.00 & +129 & 0.50 \\
 & Random+KMeans & 0.78 & 0.73 & 0.40 & 0.29 & 0.75 & 0.75 & -0.1 & 0.50 \\
\cmidrule(l){2-10}
 & Tuning-Agglo. & 0.60 & 0.70 & 0.30 & \textbf{0.51} & 0.44 & 0.52 & +534 & 0.49 \\
 & Tuning-Spectral & 0.80 & 0.70 & 0.47 & 0.21 & 0.76 & 0.65 & -55 & 0.53 \\
 & Tuning-XMeans & 0.80 & 0.61 & \textbf{0.50} & 0.00 & 0.75 & 0.50 & -129 & -- \\
 & Brent-KMeans & \textbf{0.81} & 0.67 & 0.47 & 0.14 & 0.76 & 0.62 & -72 & 0.56 \\
 & Brent-Agglo. & 0.80 & 0.66 & 0.46 & 0.11 & 0.75 & 0.59 & -82 & 0.49 \\
 & Brent-Spectral & 0.78 & 0.71 & 0.41 & 0.25 & 0.76 & 0.73 & \textbf{-14} & 0.51 \\
 & Brent-GMM & 0.80 & 0.69 & 0.46 & 0.19 & 0.78 & 0.69 & -43 & 0.58 \\
 & Transformer & 0.80 & \textbf{0.77} & 0.45 & 0.41 & \textbf{0.81} & \textbf{0.86} & +30 & \textbf{0.59} \\
\midrule
\multirow{13}{*}{\rotatebox{90}{SelEx, w/o Geo}} & Oracle & 0.85 & 0.69 & 0.60 & 0.21 & 1.00 & 1.00 & +0 & 1.00 \\
 & Supervised probe & 0.90 & 0.80 & 0.64 & 0.27 & 1.00 & 1.00 & +0 & 1.00 \\
\cmidrule(l){2-10}
 & Never-Split & 0.80 & 0.61 & 0.50 & 0.00 & 0.75 & 0.50 & -129 & -- \\
 & Always-Split & 0.68 & 0.69 & 0.10 & 0.21 & 0.75 & 1.00 & +129 & 0.50 \\
 & Random+KMeans & 0.74 & 0.65 & 0.30 & 0.10 & 0.75 & 0.75 & -0.1 & 0.50 \\
\cmidrule(l){2-10}
 & Tuning-Agglo. & 0.75 & 0.66 & 0.30 & 0.14 & 0.72 & 0.72 & +47 & 0.51 \\
 & Tuning-Spectral & 0.79 & 0.65 & 0.45 & 0.09 & 0.76 & 0.62 & -73 & 0.55 \\
 & Tuning-XMeans & \textbf{0.80} & 0.61 & \textbf{0.50} & 0.00 & 0.75 & 0.50 & -129 & -- \\
 & Brent-KMeans & 0.77 & 0.64 & 0.39 & 0.08 & \textbf{0.77} & 0.69 & -42 & 0.55 \\
 & Brent-Agglo. & 0.79 & 0.63 & 0.42 & 0.06 & 0.77 & 0.64 & -67 & \textbf{0.56} \\
 & Brent-Spectral & 0.76 & 0.67 & 0.37 & 0.14 & 0.77 & 0.73 & \textbf{-19} & 0.54 \\
 & Brent-GMM & 0.77 & 0.62 & 0.34 & 0.04 & 0.76 & 0.69 & -34 & 0.53 \\
 & Transformer & 0.69 & \textbf{0.69} & 0.12 & \textbf{0.20} & 0.75 & \textbf{0.98} & +118 & 0.50 \\
\midrule
\multirow{13}{*}{\rotatebox{90}{SelEx, w/ Geo}} & Oracle & 0.93 & 0.85 & 0.79 & 0.59 & 1.00 & 1.00 & +0 & 1.00 \\
 & Supervised probe & 0.96 & 0.93 & 0.83 & 0.66 & 1.00 & 1.00 & +0 & 1.00 \\
\cmidrule(l){2-10}
 & Never-Split & 0.80 & 0.61 & 0.50 & 0.00 & 0.75 & 0.50 & -129 & -- \\
 & Always-Split & 0.76 & 0.85 & 0.29 & 0.59 & 0.75 & 1.00 & +129 & 0.50 \\
 & Random+KMeans & 0.78 & 0.73 & 0.39 & 0.29 & 0.75 & 0.75 & -0.1 & 0.50 \\
\cmidrule(l){2-10}
 & Tuning-Agglo. & 0.60 & 0.71 & 0.31 & \textbf{0.53} & 0.46 & 0.54 & +501 & 0.49 \\
 & Tuning-Spectral & 0.80 & 0.69 & 0.47 & 0.20 & 0.75 & 0.64 & -60 & 0.51 \\
 & Tuning-XMeans & 0.80 & 0.61 & 0.50 & 0.00 & 0.75 & 0.50 & -129 & -- \\
 & Brent-KMeans & 0.81 & 0.69 & 0.49 & 0.18 & 0.77 & 0.64 & -67 & 0.60 \\
 & Brent-Agglo. & \textbf{0.82} & 0.68 & \textbf{0.52} & 0.16 & 0.78 & 0.63 & -81 & \textbf{0.69} \\
 & Brent-Spectral & 0.78 & 0.72 & 0.42 & 0.27 & 0.75 & 0.72 & -15 & 0.51 \\
 & Brent-GMM & 0.82 & 0.70 & 0.49 & 0.21 & 0.79 & 0.69 & -50 & 0.63 \\
 & Transformer & 0.81 & \textbf{0.73} & 0.48 & 0.32 & \textbf{0.80} & \textbf{0.78} & \textbf{-12} & 0.62 \\
\bottomrule
\end{tabular}
\end{adjustbox}
\label{tab:aves_full_combined_methods}
\end{table}

\noindent\textbf{Adapted fine-grained datasets.}
We adapt CUB-200-2011~\cite{wah2011caltech}, Stanford Cars~\cite{krause20133d}, and FGVC Aircraft~\cite{maji2013fine} to the discovery-by-splitting setting by treating each dataset's natural higher-level grouping (genus / make / family) as the coarse class and the leaf class as the fine class.
None of these datasets ships geographic metadata, so only visual backbones are evaluated.
Importantly, and unlike \dataset, these adapted datasets are not restricted to binary splits: a coarse class can contain up to 30 (CUB-200-2011), 22 (Stanford Cars), or 8 (FGVC Aircraft) fine classes, and the numbers of splitting ($K>1$) and non-splitting ($K=1$) coarse classes are not balanced.
Summary statistics are given in \cref{tab:fg_stats}.
All of the K-estimation and clustering methods we evaluate on \dataset apply directly to this multi-way setting, and the transformer K-estimator simply sets $K_{\max}$ to the largest number of fine classes in any training coarse class.
We restricted \dataset itself to binary splits as these are the most common form of species-level taxonomic split, with 1,208 binary splits compared to 425 multi-way splits (the majority being three-way) reported on iNaturalist since 2012.

\begin{table}[H]
\centering
\caption{\textbf{Statistics of the adapted fine-grained datasets.} Coarse classes are the natural higher-level grouping of each dataset (genus / make / family) and fine classes are the leaf classes. Max K is the largest number of fine classes in a coarse class, and K=1 / K>1 are the numbers of non-splitting and splitting coarse classes. In contrast to \dataset, these datasets contain multi-way splits and unequal numbers of splitting and non-splitting classes.}
\label{tab:fg_stats}
\vspace{3pt}
\resizebox{\linewidth}{!}{
\begin{tabular}{llrrrrrrrrr}
\toprule
Dataset & Partition & \# Images & \# Coarse & \# Fine & Imgs/coarse & Imgs/fine & Fine/coarse & Max K & K=1 & K>1 \\
\midrule
\multirow{3}{*}{CUB-200-2011} & Known & 5,930 & 19 & 100 & 312.1 & 59.3 & 5.26 & 30 & 6 & 13 \\
 & Novel & 5,858 & 19 & 100 & 308.3 & 58.6 & 5.26 & 22 & 6 & 13 \\
 & Total & 11,788 & 38 & 200 & 310.2 & 58.9 & 5.26 & 30 & 12 & 26 \\
\midrule
\multirow{3}{*}{Stanford Cars} & Known & 8,026 & 24 & 98 & 334.4 & 81.9 & 4.08 & 22 & 10 & 14 \\
 & Novel & 8,159 & 25 & 98 & 326.4 & 83.3 & 3.92 & 15 & 9 & 16 \\
 & Total & 16,185 & 49 & 196 & 330.3 & 82.6 & 4.00 & 22 & 19 & 30 \\
\midrule
\multirow{3}{*}{FGVC Aircraft} & Known & 5,000 & 34 & 50 & 147.1 & 100.0 & 1.47 & 8 & 26 & 8 \\
 & Novel & 5,000 & 36 & 50 & 138.9 & 100.0 & 1.39 & 4 & 28 & 8 \\
 & Total & 10,000 & 70 & 100 & 142.9 & 100.0 & 1.43 & 8 & 54 & 16 \\
\bottomrule
\end{tabular}
}
\end{table}

\begin{table}[H]
\centering
\caption{\textbf{Category splitting on FG datasets — backbone comparison under Brent + KMeans and transformer K estimation.} Top: Brent + KMeans K estimation. Bottom: transformer K estimation. AMI / Cl.\,Acc.\ are macro-averaged per-coarse KMeans metrics on the backbone features; K Acc.\ uses smooth score $1/(1+|\hat{K}-K^*|)$ averaged over all classes; $\Delta\!\sum\!K = \sum\hat{K} - \sum K^*$ is the signed deviation of the total estimated number of fine classes from the true total. Best value per column is bolded within each section (closest-to-zero for $\Delta\!\sum\!K$).}
\vspace{3pt}
\resizebox{\linewidth}{!}{
\begin{tabular}{c|l|cccc|cccc|cccc}
\toprule
 &  & \multicolumn{4}{c|}{\textbf{CUB-200-2011}} & \multicolumn{4}{c|}{\textbf{Stanford Cars}} & \multicolumn{4}{c}{\textbf{FGVC Aircraft}} \\
 & Method & AMI & Cl.\,Acc. & K Acc. & $\Delta\!\sum\!K$ & AMI & Cl.\,Acc. & K Acc. & $\Delta\!\sum\!K$ & AMI & Cl.\,Acc. & K Acc. & $\Delta\!\sum\!K$ \\
\midrule
\multirow{6}{*}{\rotatebox{90}{Brent-KMeans}} & DINOv1 & 0.38 & 0.52 & 0.36 & +65 & 0.14 & 0.41 & 0.36 & +30 & 0.19 & 0.61 & 0.52 & +37 \\
 & DINOv2 & 0.49 & 0.61 & 0.31 & +83 & 0.36 & 0.50 & 0.28 & +131 & 0.53 & 0.76 & 0.73 & +10 \\
 & UNO+ & 0.54 & 0.57 & 0.49 & -60 & 0.63 & 0.71 & 0.57 & -38 & 0.77 & \textbf{0.87} & 0.86 & -10 \\
 & GCD & \textbf{0.56} & \textbf{0.71} & 0.55 & \textbf{-13} & \textbf{0.64} & \textbf{0.73} & 0.58 & \textbf{+14} & 0.72 & 0.84 & \textbf{0.87} & \textbf{-2} \\
 & SimGCD & 0.52 & 0.60 & 0.49 & -37 & 0.64 & 0.73 & \textbf{0.61} & -26 & 0.75 & 0.86 & 0.85 & -8 \\
 & SelEx & 0.54 & 0.67 & \textbf{0.56} & -21 & 0.51 & 0.67 & 0.52 & +30 & \textbf{0.78} & 0.87 & 0.85 & -7 \\
\midrule
\multirow{6}{*}{\rotatebox{90}{Transformer}} & DINOv1 & 0.24 & 0.45 & 0.38 & -24 & 0.36 & 0.55 & \textbf{0.53} & -73 & \textbf{0.78} & \textbf{0.87} & \textbf{0.87} & -14 \\
 & DINOv2 & 0.46 & 0.53 & \textbf{0.42} & -43 & 0.39 & 0.58 & 0.47 & -19 & \textbf{0.78} & \textbf{0.87} & \textbf{0.87} & -14 \\
 & UNO+ & 0.23 & 0.41 & 0.29 & -39 & 0.39 & 0.58 & 0.35 & +17 & 0.72 & 0.83 & 0.84 & \textbf{-5} \\
 & GCD & 0.38 & 0.51 & 0.36 & -42 & 0.46 & 0.58 & 0.36 & +20 & 0.75 & 0.85 & 0.84 & -11 \\
 & SimGCD & \textbf{0.48} & \textbf{0.55} & 0.37 & \textbf{-19} & \textbf{0.54} & \textbf{0.62} & 0.40 & \textbf{+10} & 0.75 & 0.85 & 0.84 & -11 \\
 & SelEx & 0.40 & 0.55 & 0.40 & -48 & 0.33 & 0.51 & 0.41 & -23 & 0.78 & 0.86 & 0.86 & -11 \\
\bottomrule
\end{tabular}
}
\label{tab:fg_combined_backbone}
\end{table}

\Cref{tab:fg_combined_backbone} reports all-coarse metrics; \cref{tab:fg_combined_backbone_split} restricts to multi-class classes (true $K\!>\!1$) so that the splitting performance can be read in isolation from the no-split background rate.
The most striking difference from \dataset is that GCD-style fine-tuning of the visual backbone (GCD, SimGCD, SelEx) helps much more on these benchmarks than on \dataset.
On CUB-200-2011, GCD with Brent + KMeans reaches Cl.\,Acc.\ $0.71$ versus $0.61$ for frozen DINOv2 --- a ten-point improvement that is essentially absent on \dataset, where the visual backbones cluster within a few points of each other regardless of fine-tuning.
We attribute this to the smaller and more homogeneous label spaces of the existing benchmarks (200 classes in CUB, 196 in Stanford Cars), where contrastive fine-tuning has more room to specialize to the test distribution.
FGVC Aircraft is anomalous in the opposite direction: most coarse families contain only one fine variant, so methods that lean toward predicting $K=1$ score very high on the all-coarse view (the DINOv1 transformer reaches Cl.\,Acc.\ $0.87$ in \cref{tab:fg_combined_backbone}) while the split-only view (\cref{tab:fg_combined_backbone_split}) shows that on the few real splits these methods barely beat chance (AMI$_s$ near zero across most backbones).

\begin{table}[H]
\centering
\caption{\textbf{Category splitting on FG datasets — split-only metrics (restricted to classes with true $K>1$).} Top: Brent + KMeans K estimation. Bottom: transformer K estimation. All metrics are restricted to multi-class classes (subscript $_\text{s}$). $\Delta\!\sum\!K_\text{s}$ sums $\hat{K}-K^*$ only over multi-class classes. Best per column is bolded within each section (closest-to-zero for $\Delta\!\sum\!K_\text{s}$).}
\vspace{3pt}
\resizebox{\linewidth}{!}{
\begin{tabular}{c|l|cccc|cccc|cccc}
\toprule
 &  & \multicolumn{4}{c|}{\textbf{CUB-200-2011}} & \multicolumn{4}{c|}{\textbf{Stanford Cars}} & \multicolumn{4}{c}{\textbf{FGVC Aircraft}} \\
 & Method & AMI$_\text{s}$ & Cl.\,Acc.$_\text{s}$ & K Acc.$_\text{s}$ & $\Delta\!\sum\!K_\text{s}$ & AMI$_\text{s}$ & Cl.\,Acc.$_\text{s}$ & K Acc.$_\text{s}$ & $\Delta\!\sum\!K_\text{s}$ & AMI$_\text{s}$ & Cl.\,Acc.$_\text{s}$ & K Acc.$_\text{s}$ & $\Delta\!\sum\!K_\text{s}$ \\
\midrule
\multirow{6}{*}{\rotatebox{90}{Brent-KMeans}} & DINOv1 & 0.40 & 0.47 & 0.30 & +52 & 0.09 & 0.33 & 0.35 & -9 & 0.00 & 0.39 & 0.49 & -7 \\
 & DINOv2 & \textbf{0.64} & 0.63 & 0.27 & +64 & 0.51 & 0.53 & 0.29 & +79 & -0.00 & 0.37 & 0.42 & \textbf{-4} \\
 & UNO+ & 0.33 & 0.37 & 0.25 & -60 & 0.48 & 0.57 & 0.36 & -40 & 0.09 & 0.47 & 0.42 & -11 \\
 & GCD & 0.58 & \textbf{0.64} & \textbf{0.48} & \textbf{-19} & \textbf{0.63} & 0.67 & 0.46 & \textbf{+8} & 0.10 & \textbf{0.48} & \textbf{0.62} & -7 \\
 & SimGCD & 0.37 & 0.45 & 0.29 & -38 & 0.62 & 0.67 & \textbf{0.51} & -31 & 0.11 & 0.48 & 0.44 & -10 \\
 & SelEx & 0.48 & 0.57 & 0.48 & -27 & 0.61 & \textbf{0.68} & 0.50 & +14 & \textbf{0.11} & 0.47 & 0.44 & -10 \\
\midrule
\multirow{6}{*}{\rotatebox{90}{Transformer}} & DINOv1 & 0.35 & 0.46 & \textbf{0.44} & -42 & 0.00 & 0.30 & 0.27 & -73 & 0.00 & 0.40 & 0.40 & -14 \\
 & DINOv2 & 0.28 & 0.36 & 0.22 & -46 & 0.49 & 0.52 & \textbf{0.45} & -34 & 0.00 & 0.40 & 0.40 & -14 \\
 & UNO+ & 0.33 & 0.41 & 0.31 & -57 & 0.60 & 0.60 & 0.38 & -12 & \textbf{0.01} & \textbf{0.41} & \textbf{0.49} & \textbf{-11} \\
 & GCD & 0.41 & 0.43 & 0.30 & -54 & 0.59 & 0.59 & 0.33 & \textbf{-7} & 0.00 & 0.40 & 0.40 & -14 \\
 & SimGCD & \textbf{0.55} & \textbf{0.51} & 0.31 & \textbf{-31} & \textbf{0.65} & \textbf{0.61} & 0.32 & -8 & 0.00 & 0.40 & 0.40 & -14 \\
 & SelEx & 0.36 & 0.43 & 0.30 & -57 & 0.46 & 0.47 & 0.41 & -43 & -0.00 & 0.39 & 0.38 & \textbf{-11} \\
\bottomrule
\end{tabular}
}
\label{tab:fg_combined_backbone_split}
\end{table}

\noindent\textbf{Reading the metrics together.}
Three recommendations follow from the evidence in this section.
\textit{(1)} Always inspect $\Delta\!\sum\!K$ and Split Precision alongside aggregate Cl.\,Acc.\ and AMI: a method that defaults to ``no split'' can quietly reach $0.75$--$0.80$ Cl.\,Acc.\ on \dataset by exploiting the $\sim$50\% no-split prevalence without ever making a correct split call (\eg UNO+ under Brent + KMeans on the full dataset, Cl.\,Acc.\ $0.80$ but Split Prec.\ $0.00$).
\textit{(2)} Concatenating geographic metadata with visual backbone features leads to consistent gains across both K-estimators, both backbone types, and both dataset sizes, and the SINR-only baseline already captures most of the geographic signal.
\textit{(3)} The transformer K-estimator and Brent-Agglo (under SelEx with geographic features on the full set) currently sit at the top of the leaderboard on \dataset, but neither approaches the oracle ceiling of $0.93$, leaving substantial headroom for new methods.

\section{Generalized Category Discovery Experiments}
\label{sec:gcd}

In addition to the discovery-by-splitting setting evaluated in the main text, PS-SPLIT can be used as a generalized category discovery (GCD) benchmark, in which a model is given labeled examples of a subset of categories (\textit{known}) and unlabeled examples of all categories, and must cluster the unlabeled set so that ``new'' categories are recovered alongside the known ones~\cite{vaze22generalized}.
The dataset release includes three different sets of known/novel splits:
\begin{itemize}[itemsep=2pt,topsep=0pt,leftmargin=10pt,parsep=0pt]
\item \emph{single} splits each pair so that one of the two generated species is in the known set and the other is in the novel set;
\item \emph{paired} keeps the two splitting species together (both known or both novel) while placing the taxonomically close control species on the other side;
\item \emph{discovery-by-splitting} is the same setting used in the main paper, where each set of aligned coarse categories (the splitting pair plus the control species) are placed together, either all known, or all novel.
\end{itemize}

\paragraph{Baselines.}
We re-implement four baselines on \dataset (\cref{tab:acc}, \emph{single} setting) and standard GCD benchmarks~\cite{wah2011caltech,maji2013fine,krause20133d, krizhevsky2009learning}.
\textbf{UNO}~\cite{fini2021unified}, \textbf{GCD}~\cite{vaze22generalized}, \textbf{SimGCD}~\cite{wen2023simple}, and \textbf{SelEx}~\cite{rastegar2024selexselfexpertisefinegrainedgeneralized} are trained on the dataset using the same DINOv2 ViT-B with registers~\cite{oquab2023dinov2,darcet2024vision} so that backbone choice does not confound the comparison.
Numbers differ slightly from the published values for two reasons: (i) all methods use the same DINOv2 backbone; and (ii) we apply training-script updates that have been merged into the official repositories after publication, which generally improve performance.
\textbf{DINO}~\cite{caron2021emerging} and \textbf{DINOv2}~\cite{oquab2023dinov2} without further training are also compared using the balanced semi-supervised $k$-means clustering algorithm of SelEx~\cite{rastegar2024selexselfexpertisefinegrainedgeneralized} to cluster the image features.
Clustering accuracy for this task is computed with a single Hungarian assignment over all unlabeled instances, as described in Appendix~\ref{sec:metrics_appendix}. 

\paragraph{Results.}
Similar trends on the CUB, Cars, Aircraft, and the CIFAR datasets are seen on \dataset, with SelEx as the strongest baseline, followed by SimGCD and GCD.
The best performance on \dataset is lower than for any other dataset, showing \dataset offers value as a benchmark with room for improvement as more powerful category discovery methods are developed.
For SelEx, performance is particularly low for the novel classes, showing that the approach may be struggling to identify the very similar novel classes which have their splitting pair included in the known classes.
As in the real-world ``category discovery'' task of taxonomic splitting, multiple modalities of data may be required to accurately distinguish between visually similar species, such as vocalizations, geography or genetics.
The inclusion of geographic metadata in \dataset allows for an approximation of this approach which could lead to improved performance on this category discovery task.

\begin{table}[H]
\centering
\caption{\textbf{Clustering accuracy on all datasets for the Generalized Category Discovery task.} \textbf{All}: all classes;
         \textbf{Old}: seen classes; \textbf{New}: unseen classes. Highest accuracy for each dataset in bold. For strong category discovery approaches like SelEx, \textbf{PS-SPLIT} provides the most challenge.}
\label{tab:acc}
\resizebox{\textwidth}{!}{%
\begin{tabular}{l ccc ccc ccc ccc ccc ccc}
\toprule
 & \multicolumn{3}{c}{\textsc{PS-SPLIT}}
 & \multicolumn{3}{c}{\textsc{CUB-200}}
 & \multicolumn{3}{c}{\textsc{Scars}}
 & \multicolumn{3}{c}{\textsc{Aircraft}}
 & \multicolumn{3}{c}{\textsc{CIFAR-10}}
 & \multicolumn{3}{c}{\textsc{CIFAR-100}} \\
\cmidrule(lr){2-4}\cmidrule(lr){5-7}\cmidrule(lr){8-10}%
\cmidrule(lr){11-13}\cmidrule(lr){14-16}\cmidrule(lr){17-19}
Method & All & Old & New
       & All & Old & New
       & All & Old & New
       & All & Old & New
       & All & Old & New
       & All & Old & New \\
\midrule
DINO     & 43.2 & 45.7 & 39.0
         & 42.1 & 48.2 & 36.4
         & 12.7 & 11.7 & 14.0
         & 17.6 & 12.4 & 21.9
         & 88.8 & 91.0 & 87.8
         & 65.0 & 68.5 & 58.0 \\
DINOv2   & 68.8 & 70.8 & 65.3
         & 80.9 & 80.0 & 81.7
         & 56.1 & 49.3 & 65.6
         & 30.7 & 22.8 & 37.3
         & 96.3 & 96.3 & 96.2
         & 81.7 & 81.2 & 82.5 \\
UNO      & 47.8 & 78.8 & 32.1
         & 36.3 & 30.6 & 39.2
         & 34.1 & 29.8 & 36.2
         & 33.1 & 31.6 & 33.9
         & 42.5 & 59.4 & 34.1
         & 39.5 & 45.1 & 28.4 \\
GCD      & 69.1 & 74.2 & 60.3
         & 67.0 & 66.3 & 67.7
         & 68.5 & 62.3 & 77.0
         & 57.5 & 54.2 & 60.1
         & 95.8 & 96.6 & 95.5
         & 69.6 & 75.2 & 58.4 \\
SimGCD   & 76.5 & 66.9 & \textbf{81.3}
         & 73.7 & 79.9 & 70.6
         & 70.6 & \textbf{83.9} & 64.3
         & 66.4 & 70.0 & 64.6
         & \textbf{98.5} & 96.0 & \textbf{99.7}
         & 87.7 & 87.6 & 87.9 \\
SELEX & \textbf{76.6} & \textbf{80.5} & 70.1
         & \textbf{90.6} & \textbf{92.8} & \textbf{88.6}
         & \textbf{82.4} & 75.0 & \textbf{92.7}
         & \textbf{81.2} & \textbf{77.3} & \textbf{84.3}
         & 98.1 & \textbf{98.1} & 98.2
         & \textbf{89.2} & \textbf{89.1} & \textbf{89.2} \\
\bottomrule
\end{tabular}}
\end{table}

\section{Evaluation Metrics}
\label{sec:metrics_appendix}

Let $\mathcal{C}$ denote the set of novel (test) coarse classes and, for a coarse class $c \in \mathcal{C}$, let $K^*_c$ be its true number of fine classes, $\hat{K}_c$ the predicted number, $y_i$ the true fine label of instance $i$, and $\hat{y}_i \in \{1,\ldots,\hat{K}_c\}$ its predicted cluster.
Let $\mathcal{C}_s = \{c \in \mathcal{C} : K^*_c > 1\}$ be the splitting coarse classes.
All per-class metrics are computed independently for each coarse class and then macro-averaged over $\mathcal{C}$, or over $\mathcal{C}_s$ for the split-only variants (subscript $s$).

\noindent\textbf{Clustering accuracy.}
For coarse class $c$ with instances $\mathcal{I}_c$,
\begin{equation*}
\text{Cl.\,Acc}_c = \max_{\sigma} \frac{1}{|\mathcal{I}_c|} \sum_{i \in \mathcal{I}_c} \mathbb{1}\left[y_i = \sigma(\hat{y}_i)\right],
\end{equation*}
where $\sigma$ ranges over one-to-one mappings between predicted clusters and true fine classes, computed with the Hungarian algorithm.
For a control class ($K^*_c=1$) predicted as a single cluster, $\text{Cl.\,Acc}_c=1$; for a splitting class predicted as a single cluster, $\text{Cl.\,Acc}_c$ equals the fraction of instances in the larger fine class.

\noindent\textbf{Adjusted mutual information.}
$\text{AMI}_c$ is the adjusted mutual information between the true fine labels and the predicted clusters of coarse class $c$, which corrects mutual information for agreement expected by chance. By convention, $\text{AMI}_c=1$ when both the prediction and the ground truth consist of a single cluster, and $\text{AMI}_c=0$ when exactly one of them does.

\noindent\textbf{K accuracy.}
$\text{K\,Acc} = \frac{1}{|\mathcal{C}|}\sum_{c \in \mathcal{C}} \frac{1}{1+|\hat{K}_c - K^*_c|}$, which is 1 for a correct prediction and decreases smoothly with the size of the error. K\,Acc$_s$ averages over $\mathcal{C}_s$ only.

\noindent\textbf{Total cardinality error.}
$\Delta\!\sum\!K = \sum_{c \in \mathcal{C}} \hat{K}_c - \sum_{c \in \mathcal{C}} K^*_c$. Positive values indicate over-splitting and negative values under-splitting.

\noindent\textbf{Split precision.}
$\text{Split Prec.} = \frac{|\{c : \hat{K}_c > 1 \wedge K^*_c > 1\}|}{|\{c : \hat{K}_c > 1\}|}$, \ie of the coarse classes predicted to split, the fraction that truly split. It is undefined (reported as ``--'') for methods that never predict a split.

\noindent\textbf{Generalized category discovery accuracy.}
For the GCD experiments in \cref{tab:acc} we follow the standard protocol of~\cite{vaze22generalized}: a single Hungarian assignment between predicted clusters and ground-truth classes is computed over all unlabeled instances, and clustering accuracy is then reported over all instances (All), and separately over the instances belonging to known (Old) and novel (New) classes using this same assignment.

\section{PS-SPLIT Construction}
\label{sec:construction}

PS-SPLIT is built from research-grade observations on iNaturalist~\cite{iNaturalist} whose licenses permit redistribution for research, restricted to the class Aves (birds).
iNaturalist provides a large source of image data of species, however, the only criteria to be accepted as research-grade is that community members agree that the image contains evidence of the species being present, which does not necessarily align with images that are useful for computer vision tasks.
Due to this we construct a multi-stage pipeline to maximize the quality of the images that are included in PS-SPLIT.
The pipeline runs in five stages: split selection, control selection, image cap with user diversification, MegaDetector~\cite{beery2019efficient} crop filtering, and a vision-language quality filter. Splits whose two new species are too visually similar to be discriminated above chance are removed by a final DINOv2~\cite{oquab2023dinov2} screen. The resulting dataset contains 79{,}061 images across 771 species drawn from 257 splits.

\paragraph{Splits and controls.}
Candidate splits are taken from the iNaturalist taxonomic changes page~\cite{iNaturalistSplit}, restricted to two-way splits in which both generated species have at least 20 research-grade observations and have not undergone any further taxonomic changes. For every split, a third taxonomically close ``control'' species is selected from the same genus (or family, when the genus has no qualifying member), preferring species with similar observation counts and overlapping geographic range that have not undergone any taxonomic changes on iNaturalist. The choice of control species is intended to prevent discovery by category splitting methods from learning to predict that a category splits based on coarse-category size, broad taxonomic groupings, or geographic region.
We do not require controls to be visually indistinguishable from their splitting species, as the control is a different coarse class and the splitting decision is made within a single coarse class. However, as controls are taxonomically close they are typically somewhat visually similar, which prevents shortcuts such as ``all ducks should be split''.  For example, \cref{fig:albatross_final} and \cref{fig:nutcracker_final} show the visually similar control species alongside the corresponding splitting species.

\paragraph{Image cap and user diversification.}
For each splitting species we keep up to 100 research-grade images downloaded at the original upload size, preferring images from distinct users so that no single contributor's photographic style dominates, which could allow classes to be separated based on features related to image quality or stylistic choices rather than the species present. We select a number of images for the taxonomically close control species equal to the total of the splitting species.

\begin{figure}[H]
\centering

    \begin{minipage}[c][0.105\textwidth][c]{0.16\textwidth}%
        \centering
        \includegraphics[width=\linewidth,height=\linewidth,keepaspectratio]{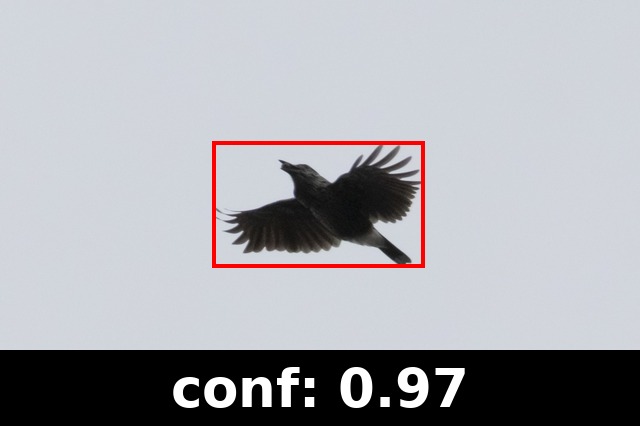}%
    \end{minipage}
    \begin{minipage}[c][0.105\textwidth][c]{0.16\textwidth}%
        \centering
        \includegraphics[width=\linewidth,height=\linewidth,keepaspectratio]{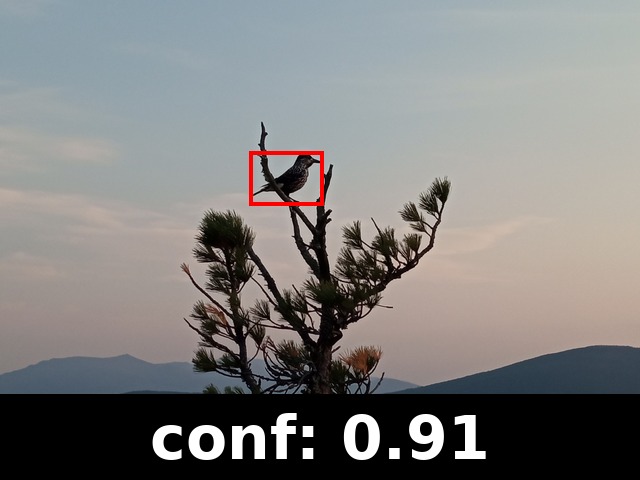}%
    \end{minipage}
    \begin{minipage}[c][0.105\textwidth][c]{0.16\textwidth}%
        \centering
        \includegraphics[width=\linewidth,height=\linewidth,keepaspectratio]{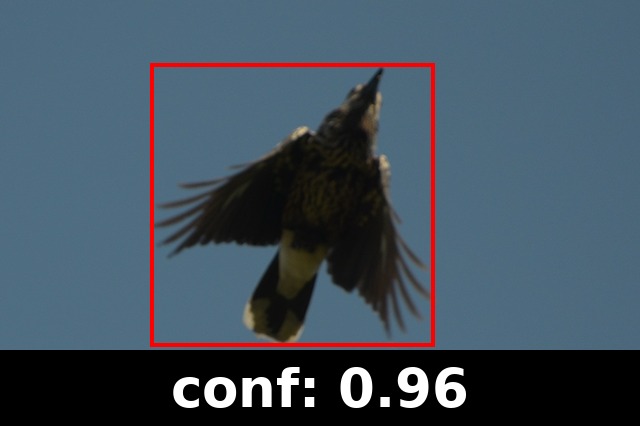}%
    \end{minipage}
    \begin{minipage}[c][0.105\textwidth][c]{0.16\textwidth}%
        \centering
        \includegraphics[width=\linewidth,height=\linewidth,keepaspectratio]{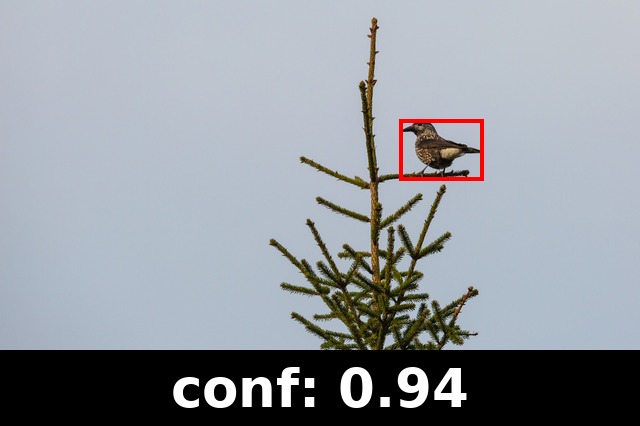}%
    \end{minipage}
    \begin{minipage}[c][0.105\textwidth][c]{0.16\textwidth}%
        \centering
        \includegraphics[width=\linewidth,height=\linewidth,keepaspectratio]{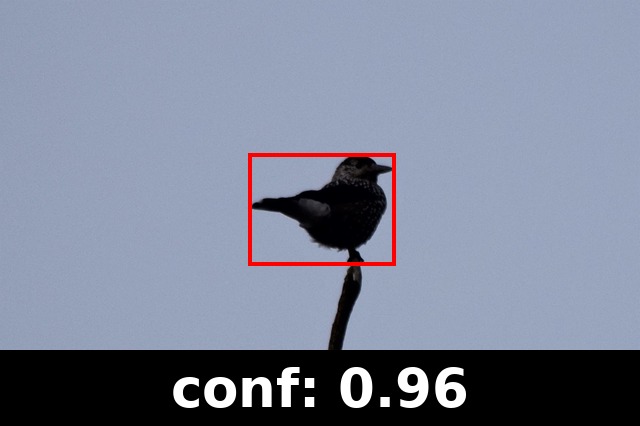}%
    \end{minipage}
    \begin{minipage}[c][0.105\textwidth][c]{0.16\textwidth}%
        \centering
        \includegraphics[width=\linewidth,height=\linewidth,keepaspectratio]{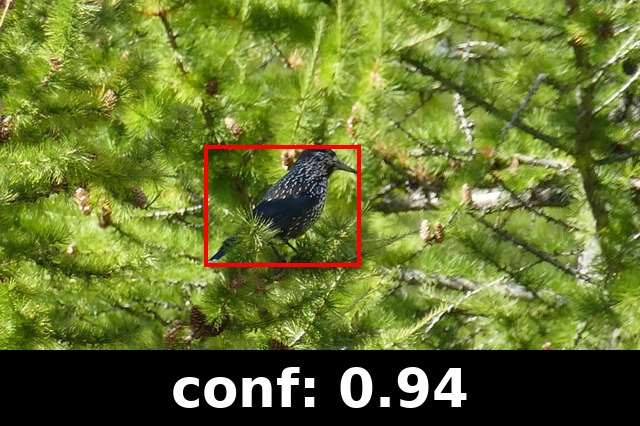}%
    \end{minipage}%

\vspace{15pt}

    \begin{minipage}[c][0.105\textwidth][c]{0.16\textwidth}%
        \centering
        \includegraphics[width=\linewidth,height=\linewidth,keepaspectratio]{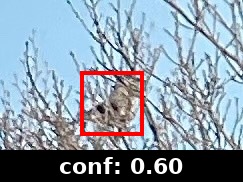}%
    \end{minipage}
    \begin{minipage}[c][0.105\textwidth][c]{0.16\textwidth}%
        \centering
        \includegraphics[width=\linewidth,height=\linewidth,keepaspectratio]{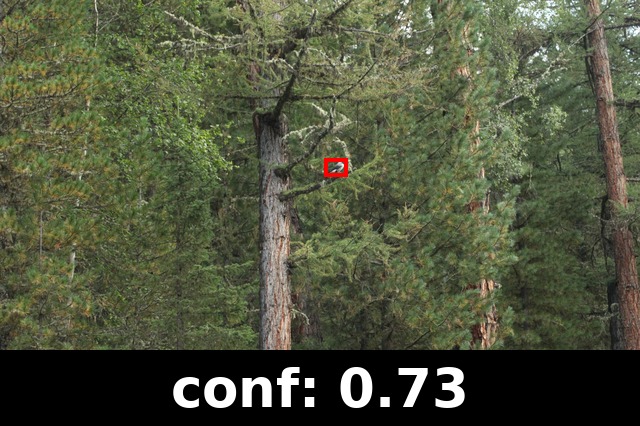}%
    \end{minipage}
    \begin{minipage}[c][0.105\textwidth][c]{0.16\textwidth}%
        \centering
        \includegraphics[width=\linewidth,height=\linewidth,keepaspectratio]{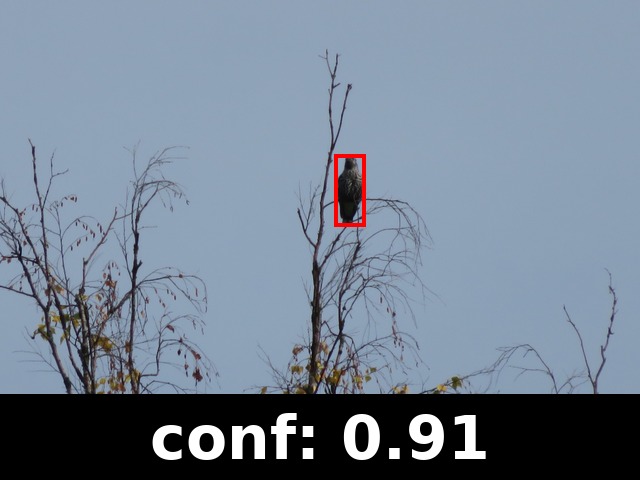}%
    \end{minipage}
    \begin{minipage}[c][0.105\textwidth][c]{0.16\textwidth}%
        \centering
        \includegraphics[width=\linewidth,height=\linewidth,keepaspectratio]{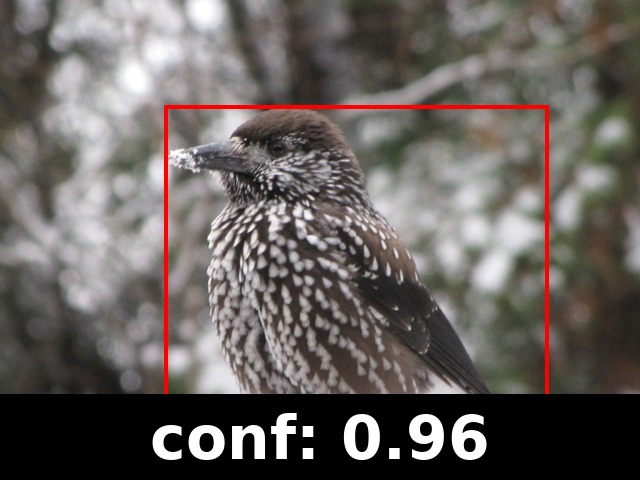}%
    \end{minipage}
    \begin{minipage}[c][0.105\textwidth][c]{0.16\textwidth}%
        \centering
        \includegraphics[width=\linewidth,height=\linewidth,keepaspectratio]{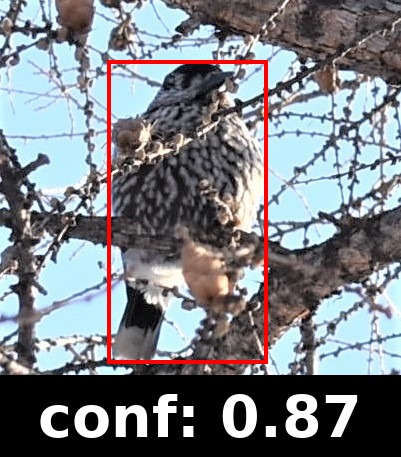}%
    \end{minipage}
    \begin{minipage}[c][0.105\textwidth][c]{0.16\textwidth}%
        \centering
        \includegraphics[width=\linewidth,height=\linewidth,keepaspectratio]{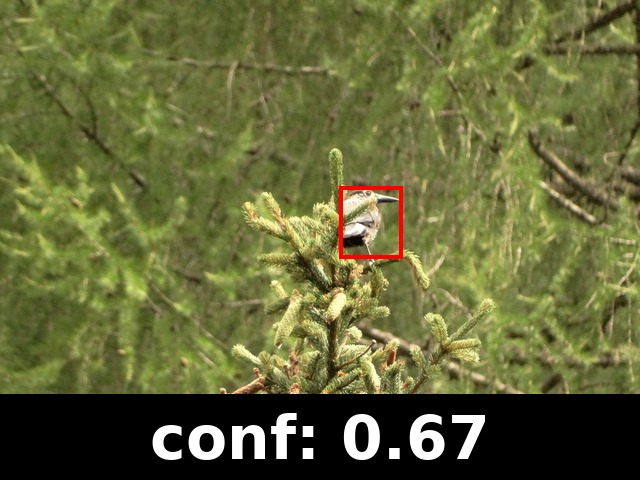}%
    \end{minipage}%

\vspace{15pt}

    \begin{minipage}[c][0.105\textwidth][c]{0.16\textwidth}%
        \centering
        \includegraphics[width=\linewidth,height=\linewidth,keepaspectratio]{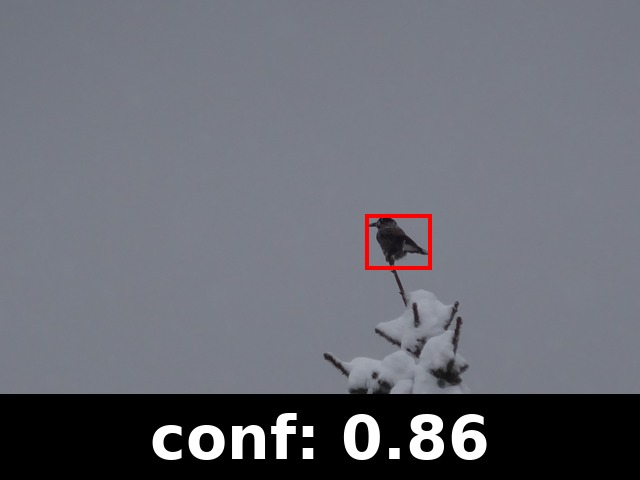}%
    \end{minipage}
    \begin{minipage}[c][0.105\textwidth][c]{0.16\textwidth}%
        \centering
        \includegraphics[width=\linewidth,height=\linewidth,keepaspectratio]{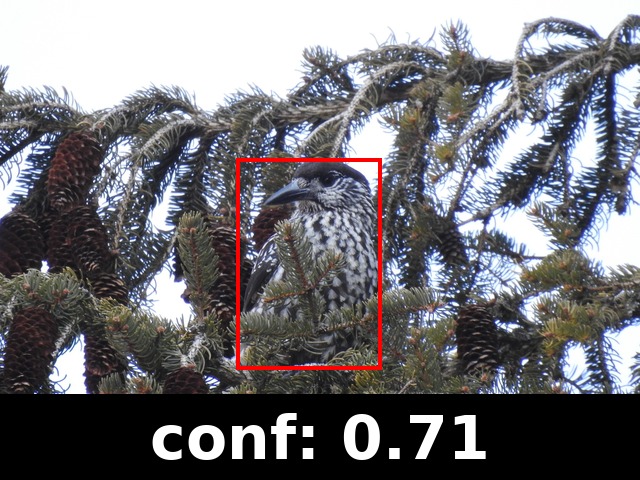}%
    \end{minipage}
    \begin{minipage}[c][0.105\textwidth][c]{0.16\textwidth}%
        \centering
        \includegraphics[width=\linewidth,height=\linewidth,keepaspectratio]{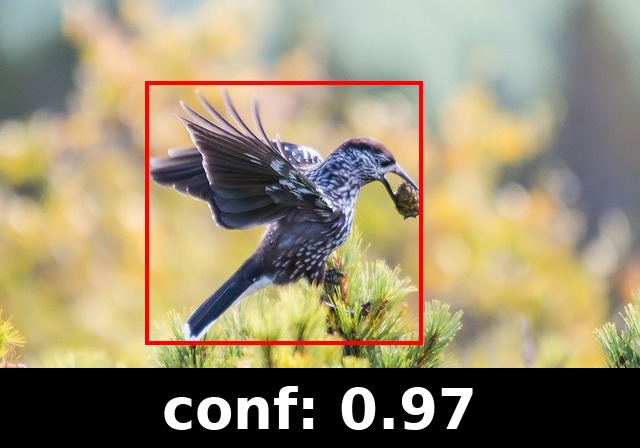}%
    \end{minipage}
    \begin{minipage}[c][0.105\textwidth][c]{0.16\textwidth}%
        \centering
        \includegraphics[width=\linewidth,height=\linewidth,keepaspectratio]{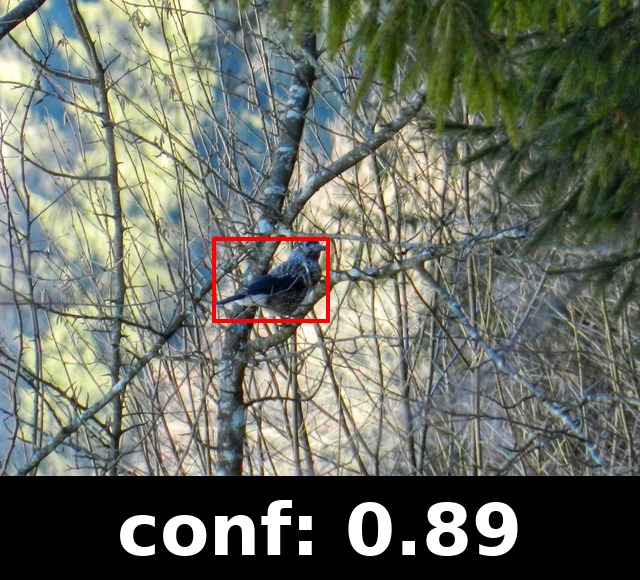}%
    \end{minipage}
    \begin{minipage}[c][0.105\textwidth][c]{0.16\textwidth}%
        \centering
        \includegraphics[width=\linewidth,height=\linewidth,keepaspectratio]{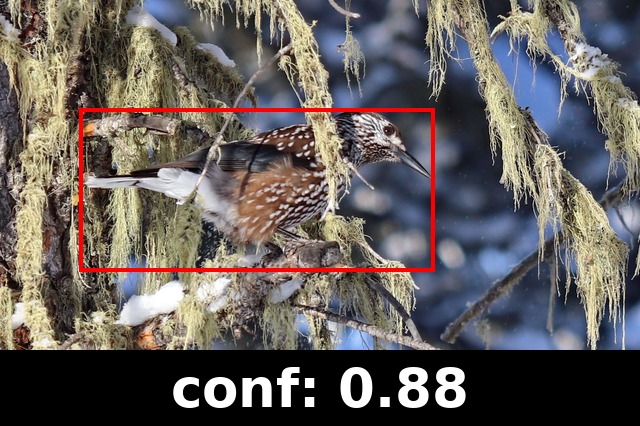}%
    \end{minipage}
    \begin{minipage}[c][0.105\textwidth][c]{0.16\textwidth}%
        \centering
        \includegraphics[width=\linewidth,height=\linewidth,keepaspectratio]{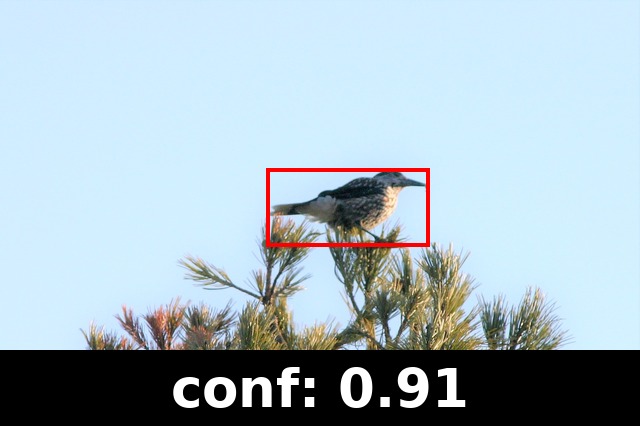}%
    \end{minipage}%

\vspace{15pt}
\caption{\textbf{Raw iNat photos with MegaDetector detections.} Red boxes show the detected animal region for \textit{Nucifraga caryocatactes} (Northern Nutcracker); the confidence score is overlaid on each box.
}
\label{fig:nutcracker_raw}
\end{figure}

\begin{figure}[H]
\centering

    \begin{minipage}{0.47\textwidth}
    \centering

    \begin{minipage}{0.32\linewidth}
        \includegraphics[width=\linewidth]{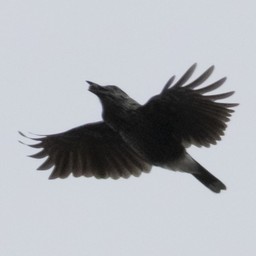}
    \end{minipage}%
    \hspace{0.2em}%
    \begin{minipage}{0.32\linewidth}
        \includegraphics[width=\linewidth]{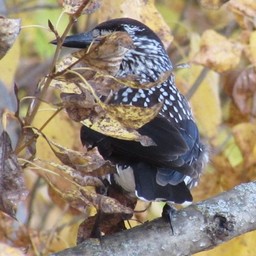}
    \end{minipage}%
    \hspace{0.2em}%
    \begin{minipage}{0.32\linewidth}
        \includegraphics[width=\linewidth]{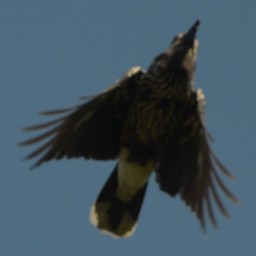}
    \end{minipage}

    \vspace{0.3em}

    \begin{minipage}{0.32\linewidth}
        \includegraphics[width=\linewidth]{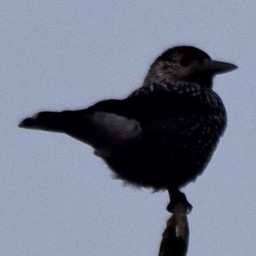}
    \end{minipage}%
    \hspace{0.2em}%
    \begin{minipage}{0.32\linewidth}
        \includegraphics[width=\linewidth]{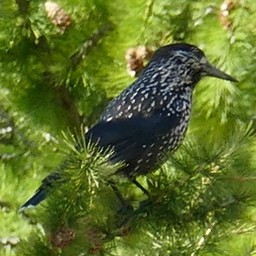}
    \end{minipage}%
    \hspace{0.2em}%
    \begin{minipage}{0.32\linewidth}
        \includegraphics[width=\linewidth]{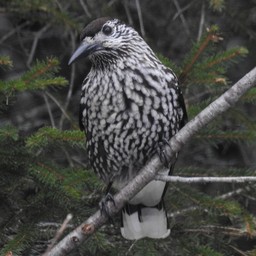}
    \end{minipage}

    \vspace{0.3em}

    \begin{minipage}{0.32\linewidth}
        \includegraphics[width=\linewidth]{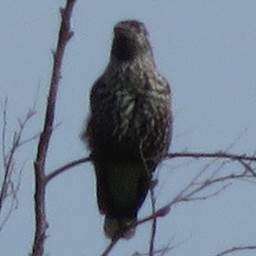}
    \end{minipage}%
    \hspace{0.2em}%
    \begin{minipage}{0.32\linewidth}
        \includegraphics[width=\linewidth]{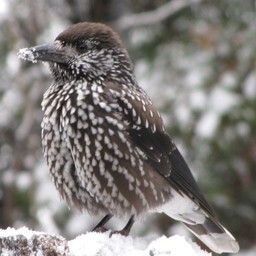}
    \end{minipage}%
    \hspace{0.2em}%
    \begin{minipage}{0.32\linewidth}
        \includegraphics[width=\linewidth]{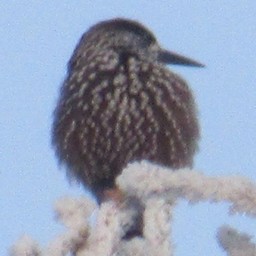}
    \end{minipage}

    \caption*{\textbf{(a) MD accepted}}
    \end{minipage}
    \hspace{1em}
    \begin{minipage}{0.47\textwidth}
    \centering

    \begin{minipage}{0.32\linewidth}
        \includegraphics[width=\linewidth]{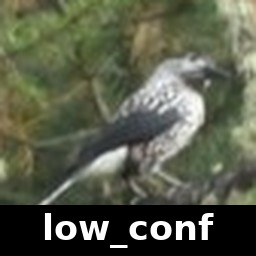}
    \end{minipage}%
    \hspace{0.2em}%
    \begin{minipage}{0.32\linewidth}
        \includegraphics[width=\linewidth]{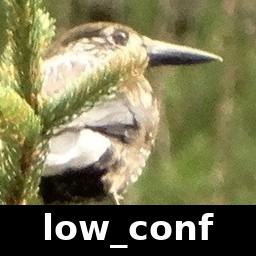}
    \end{minipage}%
    \hspace{0.2em}%
    \begin{minipage}{0.32\linewidth}
        \includegraphics[width=\linewidth]{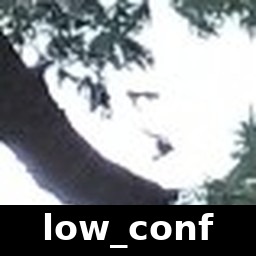}
    \end{minipage}

    \vspace{0.3em}

    \begin{minipage}{0.32\linewidth}
        \includegraphics[width=\linewidth]{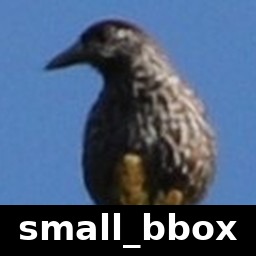}
    \end{minipage}%
    \hspace{0.2em}%
    \begin{minipage}{0.32\linewidth}
        \includegraphics[width=\linewidth]{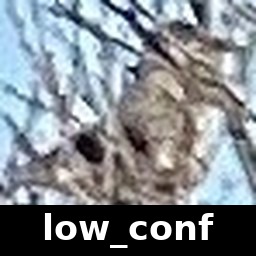}
    \end{minipage}%
    \hspace{0.2em}%
    \begin{minipage}{0.32\linewidth}
        \includegraphics[width=\linewidth]{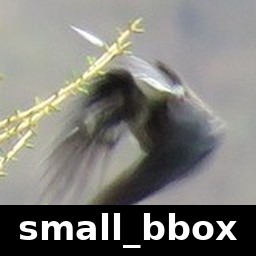}
    \end{minipage}

    \vspace{0.3em}

    \begin{minipage}{0.32\linewidth}
        \includegraphics[width=\linewidth]{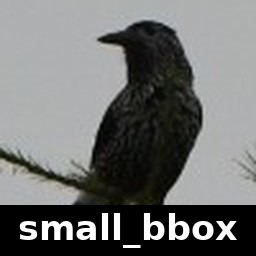}
    \end{minipage}%
    \hspace{0.2em}%
    \begin{minipage}{0.32\linewidth}
        \includegraphics[width=\linewidth]{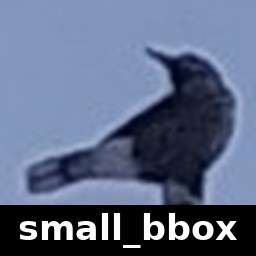}
    \end{minipage}%
    \hspace{0.2em}%
    \begin{minipage}{0.32\linewidth}
        \includegraphics[width=\linewidth]{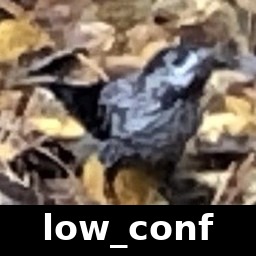}
    \end{minipage}

    \caption*{\textbf{(b) MD rejected}}
    \end{minipage}

\vspace{-5pt}
\caption{\textbf{MegaDetector 256\texttimes256 crops.} Left: crops of \textit{Nucifraga caryocatactes} (Northern Nutcracker) that pass the confidence ($\geq$0.75) and minimum bounding-box size ($\geq$190\,px) thresholds. Right: rejected crops; failure reason is overlaid (\texttt{low\_conf}: detection confidence below threshold; \texttt{small\_bbox}: detected region too small).}
\label{fig:nutcracker_crops}
\end{figure}

\paragraph{MegaDetector filter.}
Every image is passed through a MDv5 MegaDetector~\cite{beery2019efficient}. We retain only images with animal class detections with confidence $\geq 0.75$ whose bounding box has a shortest edge $\geq 190$\,pixels, and crop and resize the bounding box region to a 256$\times$256\,px image. Confidence and minimum-bounding-box thresholds discards small and distant images, as well as observations uploaded to iNaturalist that show evidence of an animal such as feathers, bones, and nests, but do not show the animal itself. The images that are discarded are unlikely to be useful for fine-grained recognition. \cref{fig:nutcracker_raw,fig:nutcracker_crops} illustrate the raw photos with MegaDetector boxes overlaid and the resulting accept / reject crops for the Northern Nutcracker.

\begin{figure}[H]
\centering
\setlength{\fboxsep}{6pt}%
\fbox{%
\begin{minipage}{\dimexpr0.95\linewidth-2\fboxsep-2\fboxrule\relax}\small
\textbf{System prompt:} You are a strict quality-control reviewer for a scientific bird image dataset. You must assess each image objectively and concisely.\\[0.4em]
\textbf{Per-image prompt:} Review this bird photograph for dataset inclusion. Answer each question with true or false, then classify the pose. Questions:
\begin{enumerate}\setlength\itemsep{0pt}
\item Does the image contain a bird (not just a nest, egg, feather, or footprint)?
\item Is the bird clearly an adult (not a chick or obvious juvenile with sparse feathers / different plumage)?
\item Is most of the bird's body visible in the frame (not just head, wing, or beak close-up)?
\item Is the bird reasonably large and in focus (not a tiny distant speck)?
\item Are there visible watermarks, logos, or overlaid descriptive text on the image?
\item Is this an unusual capture method (e.g., photo of a screen, through a telescope eyepiece, extremely heavy processing/illustration)?
\item Are there other birds or animals in the image that would significantly distract from the main subject?
\end{enumerate}
Pose classification (choose one): \texttt{"perched"}, \texttt{"flying"}, \texttt{"on-ground"}, \texttt{"in-water"}, \texttt{"other"}. Return a JSON object with the seven answers and the pose, and no other text.
\end{minipage}
}
\captionof{figure}{\textbf{VLM quality filter prompt used to score every MegaDetector crop.} Q5--Q7 are negative criteria; an image is retained only if Q1--Q4 are true and Q5--Q7 are false.}
\label{lst:vlm-prompt}
\end{figure}

\begin{figure}[H]
\centering

    \begin{minipage}{0.47\textwidth}
    \centering

    \begin{minipage}{0.32\linewidth}
        \includegraphics[width=\linewidth]{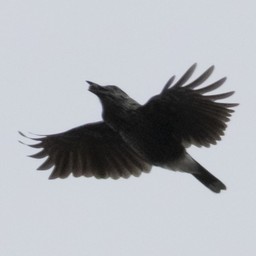}
    \end{minipage}%
    \hspace{0.2em}%
    \begin{minipage}{0.32\linewidth}
        \includegraphics[width=\linewidth]{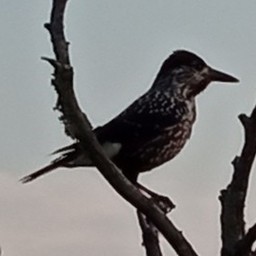}
    \end{minipage}%
    \hspace{0.2em}%
    \begin{minipage}{0.32\linewidth}
        \includegraphics[width=\linewidth]{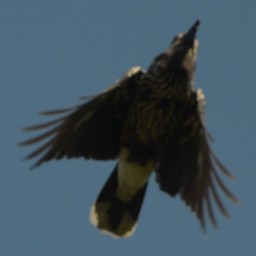}
    \end{minipage}

    \vspace{0.3em}

    \begin{minipage}{0.32\linewidth}
        \includegraphics[width=\linewidth]{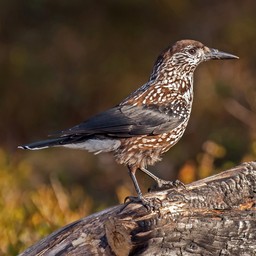}
    \end{minipage}%
    \hspace{0.2em}%
    \begin{minipage}{0.32\linewidth}
        \includegraphics[width=\linewidth]{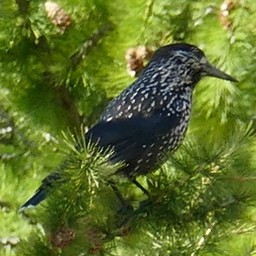}
    \end{minipage}%
    \hspace{0.2em}%
    \begin{minipage}{0.32\linewidth}
        \includegraphics[width=\linewidth]{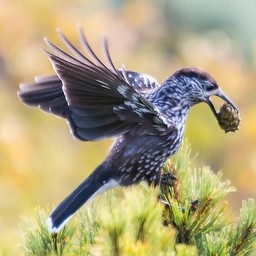}
    \end{minipage}

    \vspace{0.3em}

    \begin{minipage}{0.32\linewidth}
        \includegraphics[width=\linewidth]{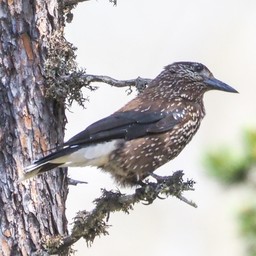}
    \end{minipage}%
    \hspace{0.2em}%
    \begin{minipage}{0.32\linewidth}
        \includegraphics[width=\linewidth]{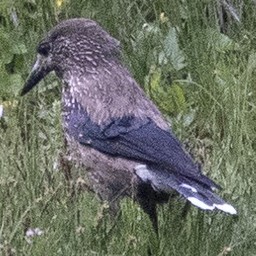}
    \end{minipage}%
    \hspace{0.2em}%
    \begin{minipage}{0.32\linewidth}
        \includegraphics[width=\linewidth]{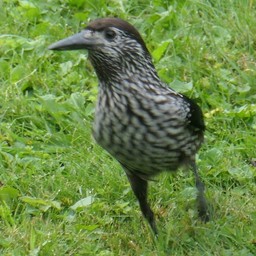}
    \end{minipage}

    \caption*{\textbf{(a) VLM accepted (pre-pose filter)}}
    \end{minipage}
    \hspace{1em}
    \begin{minipage}{0.47\textwidth}
    \centering

    \begin{minipage}{0.32\linewidth}
        \includegraphics[width=\linewidth]{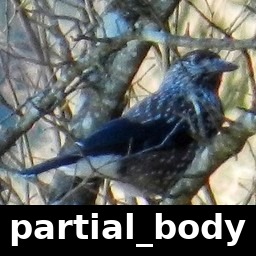}
    \end{minipage}%
    \hspace{0.2em}%
    \begin{minipage}{0.32\linewidth}
        \includegraphics[width=\linewidth]{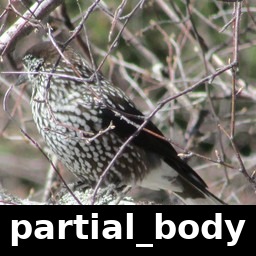}
    \end{minipage}%
    \hspace{0.2em}%
    \begin{minipage}{0.32\linewidth}
        \includegraphics[width=\linewidth]{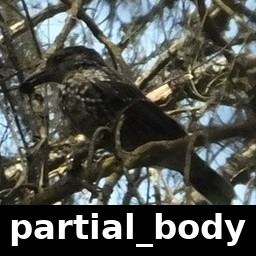}
    \end{minipage}

    \vspace{0.3em}

    \begin{minipage}{0.32\linewidth}
        \includegraphics[width=\linewidth]{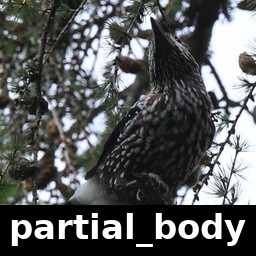}
    \end{minipage}%
    \hspace{0.2em}%
    \begin{minipage}{0.32\linewidth}
        \includegraphics[width=\linewidth]{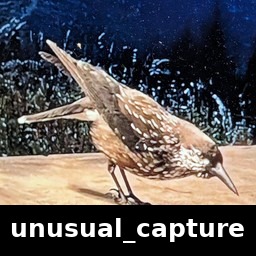}
    \end{minipage}%
    \hspace{0.2em}%
    \begin{minipage}{0.32\linewidth}
        \includegraphics[width=\linewidth]{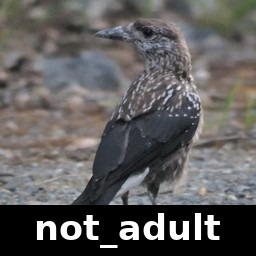}
    \end{minipage}

    \vspace{0.3em}

    \begin{minipage}{0.32\linewidth}
        \includegraphics[width=\linewidth]{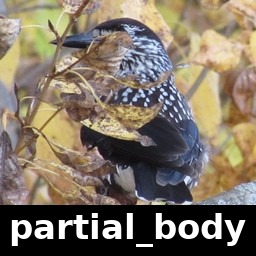}
    \end{minipage}%
    \hspace{0.2em}%
    \begin{minipage}{0.32\linewidth}
        \includegraphics[width=\linewidth]{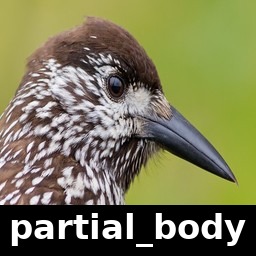}
    \end{minipage}%
    \hspace{0.2em}%
    \begin{minipage}{0.32\linewidth}
        \includegraphics[width=\linewidth]{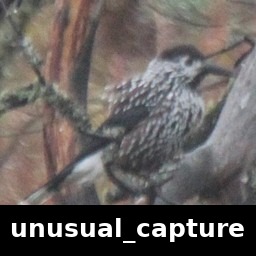}
    \end{minipage}

    \caption*{\textbf{(b) VLM rejected}}
    \end{minipage}

\vspace{-5pt}
\caption{\textbf{VLM quality filtering.} Left: images of \textit{Nucifraga caryocatactes} (Northern Nutcracker) accepted by the VLM across all poses, before pose-consistency filtering is applied. Right: images rejected for quality reasons, with rejection reason overlaid (\texttt{partial\_body}: bird heavily cropped; \texttt{not\_adult}: chick or juvenile; \texttt{unusual\_capture}: screen photo or heavy processing).}
\label{fig:nutcracker_vlm}
\end{figure}

\paragraph{Vision-language quality filter.}
Crops that pass MegaDetector are reviewed by a Qwen2.5-VL-7B-Instruct VLM~\cite{bai2025qwen25vltechnicalreport} reviewer. The system prompt and per-image prompt template are reproduced in \cref{lst:vlm-prompt}.
The reviewer answers seven binary questions about the image: is a bird present? (not a nest, egg, or footprint); is the bird an adult?; is most of the body in frame?; is the subject in focus?; is there text or a watermark present in the image?; is the image unusual (such as a photograph of a screen, photograph through a telescope eyepiece, or an image with heavy processing or illustration included)?; and are there any other prominent animals in frame?
The reviewer also assigns a broad pose label from \{\texttt{perched}, \texttt{flying}, \texttt{on-ground}, \texttt{in-water}, \texttt{other}\}.
An image is only kept if responses to all questions match our criteria.
After per-image filtering, a pose-consistency pass selects the most common pose across the two species in a split and keeps only images of that pose, ensuring that the split decision cannot be made by detecting pose alone; the filter falls back to the full pose set for any species that would otherwise drop below 20 images.
The control species is filtered in the same way, separately from its splitting pair. We do not enforce that the control pose matches the splitting pose, but in practice the pose distributions are very similar, with 82.3\% of splitting species and 85\% of control species being \texttt{perched}, reflecting the types of images users upload to iNaturalist. Most coarse classes strongly favour a single pose even without filtering.
If images have been rejected by the Megadetector or the VLM reviewer, additional images are downloaded and the process repeats until we have reached the maximum image cap or have exhausted all suitable images for that species.
\cref{fig:nutcracker_vlm} shows images for the Northern Nutcracker that have been accepted or rejected by the reviewer.

\begin{figure}[H]
\centering

    \begin{minipage}{0.47\textwidth}
    \centering

    \begin{minipage}{0.32\linewidth}
        \includegraphics[width=\linewidth]{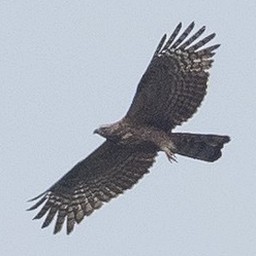}
    \end{minipage}%
    \hspace{0.2em}%
    \begin{minipage}{0.32\linewidth}
        \includegraphics[width=\linewidth]{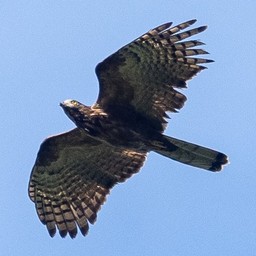}
    \end{minipage}%
    \hspace{0.2em}%
    \begin{minipage}{0.32\linewidth}
        \includegraphics[width=\linewidth]{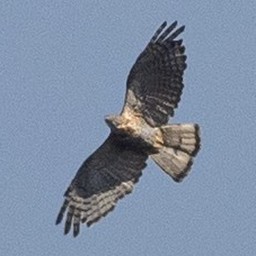}
    \end{minipage}

    \vspace{0.3em}

    \begin{minipage}{0.32\linewidth}
        \includegraphics[width=\linewidth]{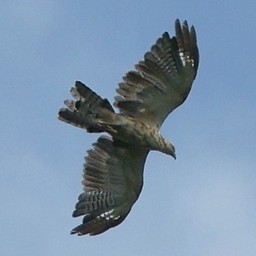}
    \end{minipage}%
    \hspace{0.2em}%
    \begin{minipage}{0.32\linewidth}
        \includegraphics[width=\linewidth]{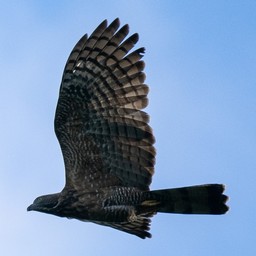}
    \end{minipage}%
    \hspace{0.2em}%
    \begin{minipage}{0.32\linewidth}
        \includegraphics[width=\linewidth]{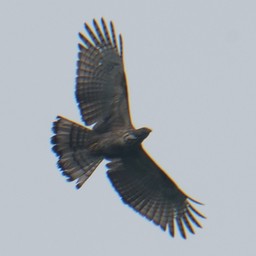}
    \end{minipage}

    \vspace{0.3em}

    \begin{minipage}{0.32\linewidth}
        \includegraphics[width=\linewidth]{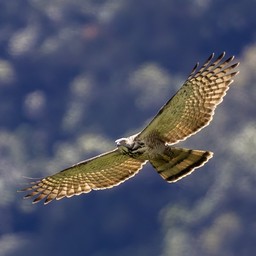}
    \end{minipage}%
    \hspace{0.2em}%
    \begin{minipage}{0.32\linewidth}
        \includegraphics[width=\linewidth]{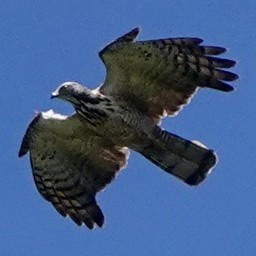}
    \end{minipage}%
    \hspace{0.2em}%
    \begin{minipage}{0.32\linewidth}
        \includegraphics[width=\linewidth]{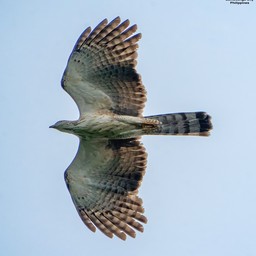}
    \end{minipage}

    \caption*{\textbf{(a) \textit{Pernis steerei} (Philippine)}}
    \end{minipage}
    \hspace{1em}
    \begin{minipage}{0.47\textwidth}
    \centering

    \begin{minipage}{0.32\linewidth}
        \includegraphics[width=\linewidth]{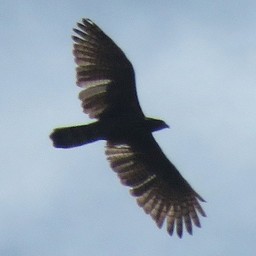}
    \end{minipage}%
    \hspace{0.2em}%
    \begin{minipage}{0.32\linewidth}
        \includegraphics[width=\linewidth]{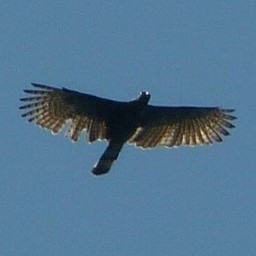}
    \end{minipage}%
    \hspace{0.2em}%
    \begin{minipage}{0.32\linewidth}
        \includegraphics[width=\linewidth]{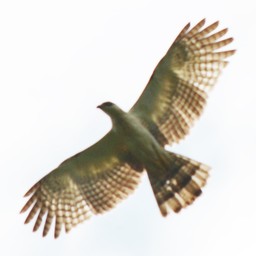}
    \end{minipage}

    \vspace{0.3em}

    \begin{minipage}{0.32\linewidth}
        \includegraphics[width=\linewidth]{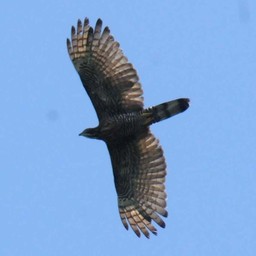}
    \end{minipage}%
    \hspace{0.2em}%
    \begin{minipage}{0.32\linewidth}
        \includegraphics[width=\linewidth]{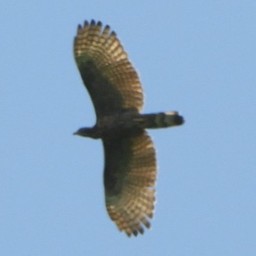}
    \end{minipage}%
    \hspace{0.2em}%
    \begin{minipage}{0.32\linewidth}
        \includegraphics[width=\linewidth]{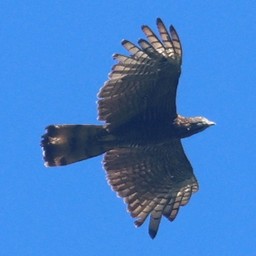}
    \end{minipage}

    \vspace{0.3em}

    \begin{minipage}{0.32\linewidth}
        \includegraphics[width=\linewidth]{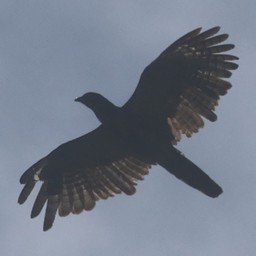}
    \end{minipage}%
    \hspace{0.2em}%
    \begin{minipage}{0.32\linewidth}
        \includegraphics[width=\linewidth]{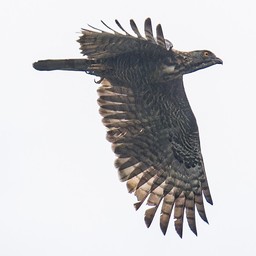}
    \end{minipage}%
    \hspace{0.2em}%
    \begin{minipage}{0.32\linewidth}
        \includegraphics[width=\linewidth]{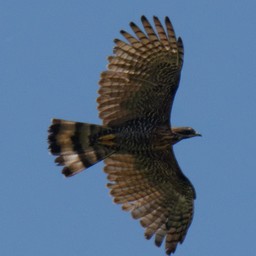}
    \end{minipage}

    \caption*{\textbf{(b) \textit{Pernis celebensis} (Sulawesi)}}
    \end{minipage}

\vspace{-5pt}
\caption{\textbf{Removed split (DINOv2 probe accuracy 57.1\%): Philippine Honey-buzzard and Sulawesi Honey-buzzard.}. Like other honey-buzzards, these species exhibit highly variable plumage morphs, making species-level identification from photographs unreliable. A DINOv2 linear probe achieves only 57.1\% accuracy on this pair.}
\label{fig:honeybuzzard_removed}
\end{figure}

\begin{figure}[H]
\centering

    \begin{minipage}{0.47\textwidth}
    \centering

    \begin{minipage}{0.32\linewidth}
        \includegraphics[width=\linewidth]{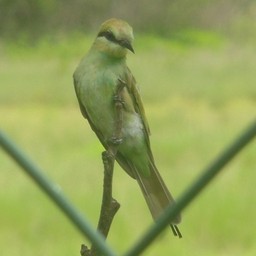}
    \end{minipage}%
    \hspace{0.2em}%
    \begin{minipage}{0.32\linewidth}
        \includegraphics[width=\linewidth]{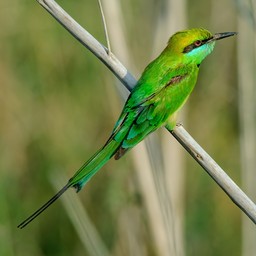}
    \end{minipage}%
    \hspace{0.2em}%
    \begin{minipage}{0.32\linewidth}
        \includegraphics[width=\linewidth]{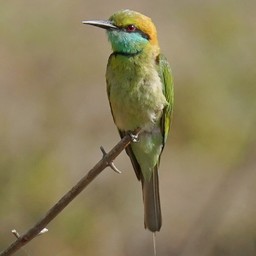}
    \end{minipage}

    \vspace{0.3em}

    \begin{minipage}{0.32\linewidth}
        \includegraphics[width=\linewidth]{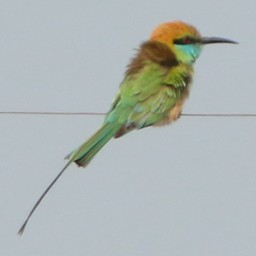}
    \end{minipage}%
    \hspace{0.2em}%
    \begin{minipage}{0.32\linewidth}
        \includegraphics[width=\linewidth]{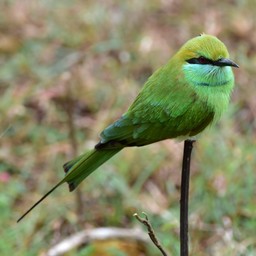}
    \end{minipage}%
    \hspace{0.2em}%
    \begin{minipage}{0.32\linewidth}
        \includegraphics[width=\linewidth]{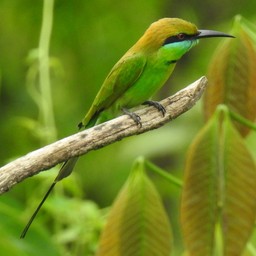}
    \end{minipage}

    \vspace{0.3em}

    \begin{minipage}{0.32\linewidth}
        \includegraphics[width=\linewidth]{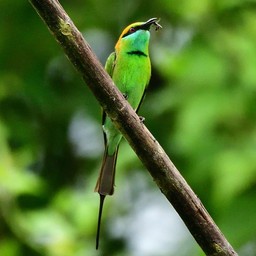}
    \end{minipage}%
    \hspace{0.2em}%
    \begin{minipage}{0.32\linewidth}
        \includegraphics[width=\linewidth]{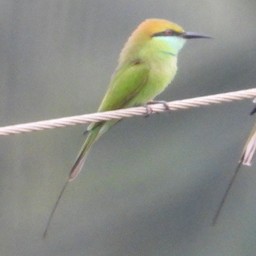}
    \end{minipage}%
    \hspace{0.2em}%
    \begin{minipage}{0.32\linewidth}
        \includegraphics[width=\linewidth]{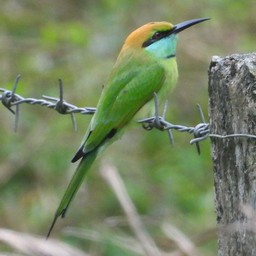}
    \end{minipage}

    \caption*{\textbf{(a) \textit{M.\,o.\,orientalis} (Common)}}
    \end{minipage}
    \hspace{1em}
    \begin{minipage}{0.47\textwidth}
    \centering

    \begin{minipage}{0.32\linewidth}
        \includegraphics[width=\linewidth]{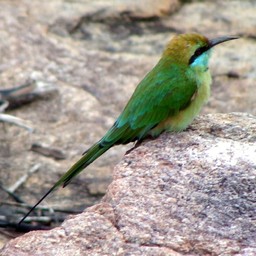}
    \end{minipage}%
    \hspace{0.2em}%
    \begin{minipage}{0.32\linewidth}
        \includegraphics[width=\linewidth]{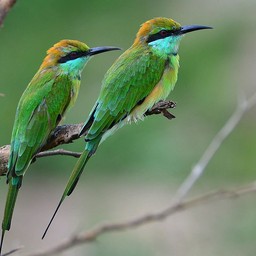}
    \end{minipage}%
    \hspace{0.2em}%
    \begin{minipage}{0.32\linewidth}
        \includegraphics[width=\linewidth]{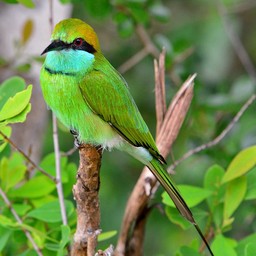}
    \end{minipage}

    \vspace{0.3em}

    \begin{minipage}{0.32\linewidth}
        \includegraphics[width=\linewidth]{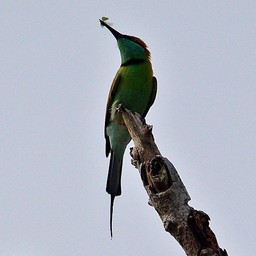}
    \end{minipage}%
    \hspace{0.2em}%
    \begin{minipage}{0.32\linewidth}
        \includegraphics[width=\linewidth]{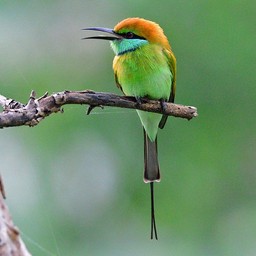}
    \end{minipage}%
    \hspace{0.2em}%
    \begin{minipage}{0.32\linewidth}
        \includegraphics[width=\linewidth]{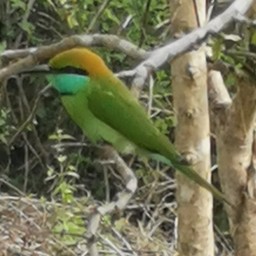}
    \end{minipage}

    \vspace{0.3em}

    \begin{minipage}{0.32\linewidth}
        \includegraphics[width=\linewidth]{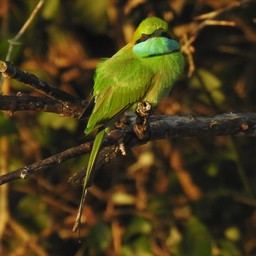}
    \end{minipage}%
    \hspace{0.2em}%
    \begin{minipage}{0.32\linewidth}
        \includegraphics[width=\linewidth]{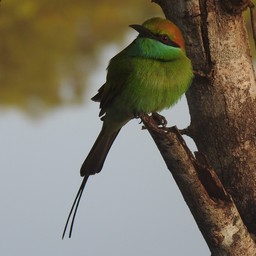}
    \end{minipage}%
    \hspace{0.2em}%
    \begin{minipage}{0.32\linewidth}
        \includegraphics[width=\linewidth]{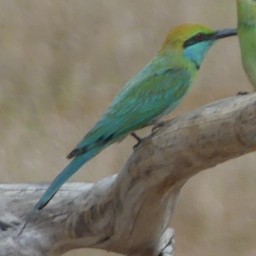}
    \end{minipage}

    \caption*{\textbf{(b) \textit{M.\,o.\,ceylonicus} (Sri Lankan)}}
    \end{minipage}

\vspace{-5pt}
\caption{\textbf{Removed split (DINOv2 probe accuracy 57.3\%): Common Green Bee-eater and Sri Lankan Green Bee-eater.} These are subspecific taxa of \textit{Merops orientalis} elevated to species rank. Both are bright green bee-eaters with rufous-and-blue throat patches; the Sri Lankan form differs only in very subtle throat-colour tone, making them essentially indistinguishable in photographs. A DINOv2 linear probe achieves only 57.3\% accuracy on this pair.}
\label{fig:beeeater_removed}
\end{figure}

\paragraph{Visual-difference screen.}
After per-image filtering, every split is evaluated by a DINOv2~\cite{oquab2023dinov2} ViT-B linear probe trained on 20\% of the species' images and evaluated on the held-out 80\%. Splits whose two new species could not be reliably distinguished (probe accuracy below 60\%) are excluded from the curated subset on the basis that the visual signal alone is insufficient for category discovery.
This is required as some taxonomic splits are based solely on non visual characteristics such as genetic analaysis, and the species produced by the taxonomic split are visually indistinguishable.
However in total only six of the 263 candidate splits were removed (2.3\%); two examples (Philippine~vs.~Sulawesi Honey-buzzard, Common~vs.~Sri Lankan Green Bee-eater) are shown in Figures~\ref{fig:honeybuzzard_removed} and~\ref{fig:beeeater_removed}.
While experts can potentially tell these species apart, it seems to be difficult, particularly for the Philippine Honey-buzzard and Sulawesi Honey-buzzard, where plumage can vary significantly within each species.
As this screen uses a pretrained representation to decide which splits are retained, it could introduce selection bias. We analyze the sensitivity of the benchmark statistics and of the method rankings to the threshold in \cref{sec:filter_sensitivity}.

\paragraph{Final dataset.}
The pipeline produces 79{,}061 images across 771 species (257 splitting pairs + 257 taxonomically similar control species), spanning 22 orders, 94 families, and 262 genera. The ``small'' version retains 64 splitting pairs where the average number of images per class is balanced, and their 64 associated control species, totaling 25{,}285 images across 192 species (14 orders, 46 families, 77 genera), approximately matching the total size of existing fine-grained visual classification datasets~\cite{wah2011caltech,maji2013fine,krause20133d} and keeping a more balanced number of images per class.
The full per-species taxonomy is shipped as supplementary JSON files (\texttt{taxonomy\_full.json} and \texttt{taxonomy\_small.json}; see Appendix~\ref{sec:taxonomy_appendix}); aggregate dataset statistics are reported in \cref{tab:aves_summary_full}. %

\section{Transformer K-Estimator: Architecture and Training}
\label{sec:transformer_appendix}

Here we provide architecture and training details for the lightweight transformer K-estimator introduced as a baseline in \cref{sec:baselines}.
The K-estimator is used purely to predict the number of fine classes $K$ for each test coarse class; downstream clustering is performed with KMeans (or a variant) on the original backbone features, not on the transformer outputs.

\begin{figure}[H]
\centering
\includegraphics[width=0.95\columnwidth]{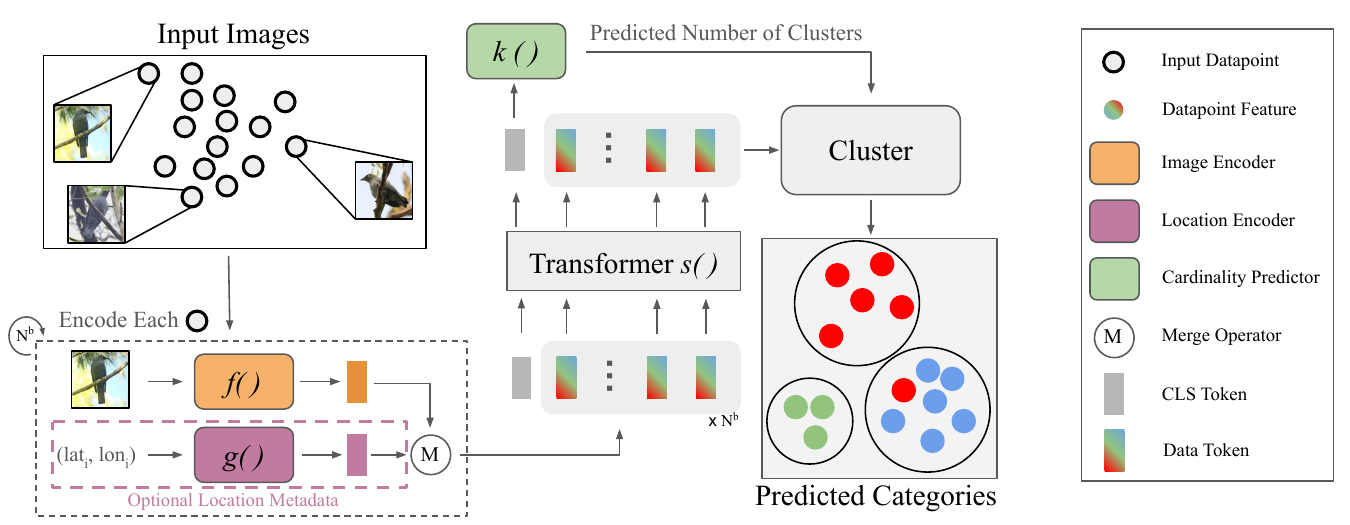}
\caption{\textbf{Transformer K-estimator overview.} Given a set of feature tokens for a single coarse class (one per instance, optionally concatenated with a geographic feature), the transformer $s_\psi$ produces a permutation-invariant CLS representation that the cardinality predictor $k_\omega$ maps to the predicted number of fine classes $K$. Downstream clustering is performed with KMeans on the backbone features given the predicted $K$.}
\label{fig:method}
\end{figure}

\noindent\textbf{Architecture.}
The K-estimator consists of two components: a permutation-invariant transformer $s_\psi$ that processes a set of per-instance feature tokens for a single coarse class, and a cardinality predictor $k_\omega$ that maps the resulting CLS-token representation to a distribution over $\{1,\ldots,K_{\max}\}$.
$s_\psi$ is a 2-layer 8-head transformer with token dimension 128 and feedforward dimension 256, followed by a final MLP with one 128-dimensional hidden layer that projects to a 64-dimensional output.
We exploit the fact that, without positional encodings, transformers are permutation invariant and can take a variable number of tokens as input~\cite{pmlr-v97-lee19d}.
$k_\omega$ is a small MLP with one 128-dimensional hidden layer; its output dimension $K_{\max}$ is set to the largest number of fine classes for any single coarse class in the training set.
For \dataset, $K_{\max}=2$ since all included splits are two-way.
Tokens fed to $s_\psi$ are the visual backbone features $f_\theta(\mathbf{x}_i)$, optionally concatenated with the geographic feature $g_\phi(\mathbf{z}_i)$ and projected by a single linear layer to dimension 128.

\noindent\textbf{Training.}
$s_\psi$ and $k_\omega$ are trained end-to-end on the known coarse classes of \dataset.
We construct training batches in which each sub-batch $\mathcal{B}^c$ consists of instances that all share the same known coarse label $y^c$.
Backbone features (and optional geographic features) are concatenated at the instance level and projected to form the input tokens $\{\mathbf{t}_i\}_{i \in \mathcal{B}^c}$ for $s_\psi$, with a learnable CLS token $\mathbf{t}_{\mathrm{CLS}}$ appended.

Two losses are used.
The first is a supervised contrastive loss on the per-instance output tokens that encourages tokens to group in feature space according to their fine labels.
For each anchor $i$ we define the positive set $\mathcal{N}_i = \{ j \in \mathcal{B}^c \setminus \{i\} : y_j^f = y_i^f \}$ and apply
\begin{equation}
\mathcal{L}^r
= \frac{1}{|\mathcal{B}^c|}
  \sum_{i \in \mathcal{B}^c}
  \frac{1}{|\mathcal{N}_i|}
  \sum_{j \in \mathcal{N}_i}
  -\log
  \frac{\exp(\mathbf{t}_i^{\top} \mathbf{t}_j / \tau)}
       {\sum_{k \in \mathcal{B}^c \setminus \{i\}}
        \exp(\mathbf{t}_i^{\top} \mathbf{t}_k / \tau)},
\label{eq:supcon_fine}
\end{equation}
with temperature $\tau=0.07$.
The second is a categorical cross-entropy loss on the cardinality prediction at the CLS token, given the ground-truth cardinality index $K^* \in \{1,\ldots,K_{\max}\}$:
\begin{equation}
\mathcal{L}^c
= - \log \big[\, sm(k_\omega(\mathbf{t}_{\mathrm{CLS}}))_{K^*}\big].
\end{equation}
The total loss is $\mathcal{L} = \mathcal{L}^c + \lambda \mathcal{L}^r$ with $\lambda = 0.3$.

\noindent\textbf{Optimisation and ensembling.}
We use SGD with momentum 0.9 and learning rate 0.1, following the schedule of SimGCD~\cite{wen2023simple}.
We train for 50 epochs, which takes around ten minutes on a single A6000 GPU for the full \dataset (assuming precomputed backbone features) and less on \dataset-Small.
Taking advantage of the small model size (roughly 500K trainable parameters), we share a single backbone and average logits across an ensemble of five $s_\psi$/$k_\omega$ heads.
At test time the model is given all instances of a single test coarse class (with the test coarse-label space disjoint from training); $k_\omega$ predicts $K$ and downstream KMeans clustering is performed on the original backbone features.

\section{Additional Analyses}
\label{sec:rebuttal_analyses}
Here we report additional analyses on the sensitivity to the visual-difference screen, the effect of geography, and training convergence.

\begin{table}[H]
\centering
\caption{\textbf{Stability of method rankings under stricter visual-difference thresholds, \dataset-Small.} Spearman rank correlation with the ordering at the 0.60 threshold used in the paper, averaged over all sections and metrics of \cref{tab:aves_small_combined_backbone,tab:aves_small_combined_methods}, together with the fraction of coarse classes and images discarded relative to 0.60.}
\label{tab:spearman_thresh}
\vspace{3pt}
\begin{tabular}{lcccc}
\toprule
Threshold & 0.60 (paper) & 0.70 & 0.80 & 0.90 \\
\midrule
Spearman $\rho$ \vs 0.60 ordering & 1.00 (ref.) & 0.96 & 0.88 & 0.78 \\
\% of coarse classes discarded & 0\% & 6.2\% & 30.7\% & 62.6\% \\
\% of images discarded & 0\% & 4.8\% & 28.6\% & 58.0\% \\
\bottomrule
\end{tabular}
\end{table}

\begin{table}[H]
\centering

\caption{\textbf{Sensitivity of PS-SPLIT-Full to the DINOv2 visual-difference probe threshold.} ``Threshold'' is the minimum held-out linear-probe accuracy required for a candidate splitting pair to be retained. The paper uses threshold $=0.60$ (row in bold). Rows for $0.65$--$0.90$ apply a stricter cutoff on top of the candidate pool using a DINOv2 linear probe per pair (5-seed average, evaluated on the official test split). ``All retained'' is the unfiltered candidate pool of 263 pairs, before the $0.60$ threshold is applied.}
\label{tab:aves_threshold_sens_full}
\resizebox{\linewidth}{!}{
\begin{tabular}{lrrrrrrrrr}
\toprule
Threshold & \# Images & \# Coarse & Splits & Controls & \# Fine & Imgs/coarse & Imgs/fine & Fine/coarse & Mean $|\Delta|$ \\
\midrule
All retained & 80,842 & 526 & 263 & 263 & 789 & 153.7 & 102.5 & 1.50 & 2.2 \\
\textbf{0.60 (paper)} & 79,061 & 514 & 257 & 257 & 771 & 153.8 & 102.5 & 1.50 & 2.2 \\
0.65 & 77,789 & 504 & 252 & 252 & 756 & 154.3 & 102.9 & 1.50 & 2.3 \\
0.70 & 75,283 & 482 & 241 & 241 & 723 & 156.2 & 104.1 & 1.50 & 2.4 \\
0.75 & 68,349 & 432 & 216 & 216 & 648 & 158.2 & 105.5 & 1.50 & 2.4 \\
0.80 & 56,451 & 356 & 178 & 178 & 534 & 158.6 & 105.7 & 1.50 & 2.5 \\
0.85 & 45,153 & 276 & 138 & 138 & 414 & 163.6 & 109.1 & 1.50 & 2.5 \\
0.90 & 33,215 & 192 & 96 & 96 & 288 & 173.0 & 115.3 & 1.50 & 2.9 \\
\bottomrule
\end{tabular}
}
\end{table}

\subsection{Sensitivity to the Visual-Difference Screen}
\label{sec:filter_sensitivity}
The visual-difference screen (Appendix~\ref{sec:construction}) removes candidate splits whose two species a DINOv2 linear probe cannot separate with at least 60\% accuracy.
As this uses a pretrained representation to decide which splits are retained, it could bias the benchmark toward splits that this representation can already separate.
At our threshold of 0.60 only 6 of the 263 candidate splits (2.3\%) are removed, so the impact of any such bias is limited.
\Cref{tab:aves_threshold_sens_full} shows how the dataset statistics change for stricter thresholds. Apart from the expected decrease in dataset size, the number of images per class slowly increases, as species with more images have a larger training set for the probe and so tend to survive stricter filtering.

To assess the stability of the method rankings, we recompute the \dataset-Small results of \cref{tab:aves_small_combined_backbone,tab:aves_small_combined_methods} restricted to the coarse classes retained at thresholds of 0.70, 0.80, and 0.90, and report the Spearman rank correlation with the ordering at 0.60, averaged over all sections and metrics, in \cref{tab:spearman_thresh}.
The ranking is very stable for small changes to the threshold, and remains similar even at a very strict threshold of 0.90, which discards 62.6\% of the coarse classes.
The qualitative conclusions are unchanged at every threshold: SINR is the best backbone under Brent + KMeans, Tuning-XMeans degenerates to ``never split'', and the transformer K-estimator retains the highest split-only metrics.
One change with very strict filtering (0.90) is that methods using fine-tuned backbones, such as SelEx and GCD, no longer benefit from additional location features, as the remaining species are visually quite distinct.

\subsection{Geographic Breakdown and the Effect of Location}
\label{sec:geo_breakdown}

\begin{table}[H]
\centering
\caption{\textbf{Average effect of adding geographic (SINR location) features on Clustering Accuracy (top) and AMI (bottom), \dataset{}-Full.} Each novel coarse category is assigned to a continent by plurality vote over its observation locations, so the continent columns are disjoint and sum to Overall. For each backbone, values are averaged over the 8 non-oracle K-estimation/clustering methods of \cref{tab:aves_full_combined_methods} without (w/o) and with (w/) geographic features; Average is the mean over the DINOv2 and SelEx backbones. $\Delta$ is the difference (w/ $-$ w/o, italic), so positive values mean that location features helped in that region. Within each w/o / w/ pair, the better value per column is bolded.}
\label{tab:aves_full_geo_compressed_acc}
\label{tab:aves_full_geo_compressed_ami}
\vspace{3pt}
\resizebox{\linewidth}{!}{
\begin{tabular}{ll|cc|cc|cc|cc|cc|cc|cc}
\toprule
 & & \multicolumn{2}{c}{Overall} & \multicolumn{2}{c}{Asia} & \multicolumn{2}{c}{N.\ America} & \multicolumn{2}{c}{S.\ America} & \multicolumn{2}{c}{Africa} & \multicolumn{2}{c}{Europe} & \multicolumn{2}{c}{Oceania} \\
 & & \multicolumn{2}{c}{($n$=258)} & \multicolumn{2}{c}{($n$=72)} & \multicolumn{2}{c}{($n$=60)} & \multicolumn{2}{c}{($n$=40)} & \multicolumn{2}{c}{($n$=38)} & \multicolumn{2}{c}{($n$=30)} & \multicolumn{2}{c}{($n$=18)} \\
\midrule
Backbone & Features & Cl.\,Acc. & Cl.\,Acc.$_\text{s}$ & Cl.\,Acc. & Cl.\,Acc.$_\text{s}$ & Cl.\,Acc. & Cl.\,Acc.$_\text{s}$ & Cl.\,Acc. & Cl.\,Acc.$_\text{s}$ & Cl.\,Acc. & Cl.\,Acc.$_\text{s}$ & Cl.\,Acc. & Cl.\,Acc.$_\text{s}$ & Cl.\,Acc. & Cl.\,Acc.$_\text{s}$ \\
\midrule
\multirow{3}{*}{DINOv2} & w/o Geo & 0.75 & 0.63 & 0.74 & 0.63 & 0.75 & 0.55 & 0.75 & 0.64 & 0.74 & 0.66 & 0.78 & 0.62 & \textbf{0.80} & \textbf{0.69} \\
 & w/ Geo & \textbf{0.78} & \textbf{0.69} & \textbf{0.76} & \textbf{0.70} & \textbf{0.78} & \textbf{0.66} & \textbf{0.80} & \textbf{0.73} & \textbf{0.77} & \textbf{0.71} & \textbf{0.78} & \textbf{0.62} & 0.79 & 0.67 \\
 & $\Delta$ & \textit{+0.02} & \textit{+0.06} & \textit{+0.03} & \textit{+0.07} & \textit{+0.03} & \textit{+0.11} & \textit{+0.04} & \textit{+0.09} & \textit{+0.03} & \textit{+0.05} & \textit{+0.00} & \textit{+0.00} & \textit{-0.01} & \textit{-0.02} \\
\cmidrule{1-16}
\multirow{3}{*}{SelEx} & w/o Geo & 0.76 & 0.65 & 0.76 & 0.67 & 0.76 & 0.57 & 0.77 & 0.65 & 0.75 & 0.67 & \textbf{0.78} & \textbf{0.64} & \textbf{0.83} & \textbf{0.69} \\
 & w/ Geo & \textbf{0.78} & \textbf{0.69} & \textbf{0.78} & \textbf{0.71} & \textbf{0.76} & \textbf{0.62} & \textbf{0.81} & \textbf{0.75} & \textbf{0.77} & \textbf{0.70} & 0.78 & 0.63 & 0.82 & 0.69 \\
 & $\Delta$ & \textit{+0.02} & \textit{+0.05} & \textit{+0.03} & \textit{+0.05} & \textit{+0.01} & \textit{+0.05} & \textit{+0.04} & \textit{+0.10} & \textit{+0.02} & \textit{+0.03} & \textit{-0.00} & \textit{-0.00} & \textit{-0.01} & \textit{-0.00} \\
\cmidrule{1-16}
\multirow{3}{*}{Average} & w/o Geo & 0.76 & 0.64 & 0.75 & 0.65 & 0.75 & 0.56 & 0.76 & 0.65 & 0.74 & 0.67 & 0.78 & 0.63 & \textbf{0.82} & \textbf{0.69} \\
 & w/ Geo & \textbf{0.78} & \textbf{0.69} & \textbf{0.77} & \textbf{0.70} & \textbf{0.77} & \textbf{0.64} & \textbf{0.80} & \textbf{0.74} & \textbf{0.77} & \textbf{0.70} & \textbf{0.78} & \textbf{0.63} & 0.81 & 0.68 \\
 & $\Delta$ & \textit{+0.02} & \textit{+0.05} & \textit{+0.03} & \textit{+0.06} & \textit{+0.02} & \textit{+0.08} & \textit{+0.04} & \textit{+0.09} & \textit{+0.02} & \textit{+0.04} & \textit{+0.00} & \textit{+0.00} & \textit{-0.01} & \textit{-0.01} \\
\midrule
\midrule
Backbone & Features & AMI & AMI$_\text{s}$ & AMI & AMI$_\text{s}$ & AMI & AMI$_\text{s}$ & AMI & AMI$_\text{s}$ & AMI & AMI$_\text{s}$ & AMI & AMI$_\text{s}$ & AMI & AMI$_\text{s}$ \\
\midrule
\multirow{3}{*}{DINOv2} & w/o Geo & 0.35 & 0.06 & 0.32 & 0.06 & 0.38 & 0.02 & 0.32 & 0.04 & 0.28 & 0.08 & 0.46 & 0.12 & 0.44 & 0.10 \\
 & w/ Geo & \textbf{0.44} & \textbf{0.23} & \textbf{0.39} & \textbf{0.22} & \textbf{0.51} & \textbf{0.28} & \textbf{0.43} & \textbf{0.25} & \textbf{0.36} & \textbf{0.23} & \textbf{0.51} & \textbf{0.16} & \textbf{0.47} & \textbf{0.15} \\
 & $\Delta$ & \textit{+0.09} & \textit{+0.17} & \textit{+0.07} & \textit{+0.16} & \textit{+0.14} & \textit{+0.26} & \textit{+0.11} & \textit{+0.21} & \textit{+0.09} & \textit{+0.15} & \textit{+0.05} & \textit{+0.04} & \textit{+0.03} & \textit{+0.06} \\
\cmidrule{1-16}
\multirow{3}{*}{SelEx} & w/o Geo & 0.36 & 0.09 & 0.33 & 0.12 & 0.39 & 0.03 & 0.33 & 0.07 & 0.29 & 0.10 & 0.44 & 0.13 & 0.48 & 0.12 \\
 & w/ Geo & \textbf{0.46} & \textbf{0.23} & \textbf{0.44} & \textbf{0.25} & \textbf{0.49} & \textbf{0.20} & \textbf{0.46} & \textbf{0.29} & \textbf{0.37} & \textbf{0.21} & \textbf{0.51} & \textbf{0.19} & \textbf{0.55} & \textbf{0.22} \\
 & $\Delta$ & \textit{+0.10} & \textit{+0.14} & \textit{+0.11} & \textit{+0.13} & \textit{+0.11} & \textit{+0.18} & \textit{+0.13} & \textit{+0.22} & \textit{+0.08} & \textit{+0.11} & \textit{+0.07} & \textit{+0.05} & \textit{+0.08} & \textit{+0.11} \\
\cmidrule{1-16}
\multirow{3}{*}{Average} & w/o Geo & 0.36 & 0.08 & 0.32 & 0.09 & 0.38 & 0.02 & 0.33 & 0.05 & 0.28 & 0.09 & 0.45 & 0.13 & 0.46 & 0.11 \\
 & w/ Geo & \textbf{0.45} & \textbf{0.23} & \textbf{0.41} & \textbf{0.23} & \textbf{0.50} & \textbf{0.24} & \textbf{0.45} & \textbf{0.27} & \textbf{0.37} & \textbf{0.22} & \textbf{0.51} & \textbf{0.17} & \textbf{0.51} & \textbf{0.19} \\
 & $\Delta$ & \textit{+0.09} & \textit{+0.16} & \textit{+0.09} & \textit{+0.15} & \textit{+0.12} & \textit{+0.22} & \textit{+0.12} & \textit{+0.22} & \textit{+0.08} & \textit{+0.13} & \textit{+0.06} & \textit{+0.05} & \textit{+0.05} & \textit{+0.08} \\
\bottomrule
\end{tabular}
}
\end{table}

\begin{table}[H]
\centering
\caption{\textbf{Geographic breakdown of Clustering Accuracy, \dataset{}-Full.} Each novel coarse category is assigned to a continent by plurality vote over its citizen-science observation locations (point-in-polygon against Natural Earth country boundaries). Columns are otherwise disjoint and sum to Overall ($n$=258). This is a backbone comparison: rows are backbones, sectioned by K-estimation method (Brent+KMeans vs.\ transformer). Best value per column is bolded and second-best is underlined within each section.}
\resizebox{\linewidth}{!}{
\begin{tabular}{c|l|cc|cc|cc|cc|cc|cc|cc}
\toprule
 & & \multicolumn{2}{c}{Overall ($n$=258)} & \multicolumn{2}{c}{Asia ($n$=72)} & \multicolumn{2}{c}{North America ($n$=60)} & \multicolumn{2}{c}{South America ($n$=40)} & \multicolumn{2}{c}{Africa ($n$=38)} & \multicolumn{2}{c}{Europe ($n$=30)} & \multicolumn{2}{c}{Oceania ($n$=18)} \\
 & Row & Cl.\,Acc. & Cl.\,Acc.$_\text{s}$ & Cl.\,Acc. & Cl.\,Acc.$_\text{s}$ & Cl.\,Acc. & Cl.\,Acc.$_\text{s}$ & Cl.\,Acc. & Cl.\,Acc.$_\text{s}$ & Cl.\,Acc. & Cl.\,Acc.$_\text{s}$ & Cl.\,Acc. & Cl.\,Acc.$_\text{s}$ & Cl.\,Acc. & Cl.\,Acc.$_\text{s}$ \\
\midrule
\multirow{8}{*}{\rotatebox{90}{Brent-KMeans}} & DINOv1 & 0.75 & 0.60 & 0.70 & 0.61 & 0.77 & 0.56 & 0.78 & 0.66 & 0.73 & 0.61 & 0.80 & 0.56 & 0.72 & 0.58 \\
 & DINOv2 & 0.77 & 0.64 & 0.74 & 0.64 & 0.78 & 0.56 & 0.76 & 0.63 & 0.76 & 0.69 & \underline{0.83} & 0.66 & 0.85 & \underline{0.68} \\
 & UNO+ & 0.80 & 0.61 & 0.77 & 0.63 & \textbf{0.82} & 0.54 & \underline{0.80} & 0.64 & 0.77 & 0.63 & 0.81 & 0.56 & \underline{0.88} & 0.64 \\
 & GCD & 0.77 & \underline{0.67} & 0.76 & \underline{0.68} & 0.75 & 0.58 & 0.78 & \underline{0.67} & \textbf{0.82} & \textbf{0.75} & 0.76 & \textbf{0.71} & 0.83 & 0.65 \\
 & SimGCD & 0.80 & 0.62 & 0.77 & 0.63 & \textbf{0.82} & 0.54 & 0.80 & 0.64 & 0.76 & 0.63 & \textbf{0.84} & 0.60 & \underline{0.88} & 0.64 \\
 & SelEx & 0.77 & 0.64 & 0.75 & 0.65 & 0.76 & 0.56 & 0.75 & 0.64 & 0.77 & 0.68 & 0.79 & 0.63 & \textbf{0.89} & \textbf{0.72} \\
 & SINR & \textbf{0.81} & \textbf{0.71} & \underline{0.78} & 0.68 & \underline{0.81} & \textbf{0.72} & \textbf{0.86} & \textbf{0.79} & \underline{0.80} & 0.70 & 0.79 & \underline{0.69} & 0.83 & 0.67 \\
 & DINOv2 SINR & \underline{0.81} & 0.67 & \textbf{0.81} & \textbf{0.70} & 0.81 & \underline{0.65} & 0.79 & 0.67 & 0.79 & \underline{0.71} & 0.81 & 0.59 & 0.83 & 0.62 \\
\midrule
\multirow{8}{*}{\rotatebox{90}{Transformer}} & DINOv1 & 0.65 & 0.59 & 0.62 & 0.58 & 0.67 & 0.57 & 0.65 & 0.60 & 0.63 & 0.60 & 0.76 & 0.60 & 0.65 & 0.58 \\
 & DINOv2 & 0.71 & 0.65 & 0.70 & 0.65 & 0.70 & 0.59 & 0.71 & 0.65 & 0.73 & 0.69 & 0.75 & 0.64 & 0.71 & 0.65 \\
 & UNO+ & \underline{0.80} & 0.61 & \textbf{0.77} & 0.62 & 0.81 & 0.54 & \textbf{0.79} & 0.64 & 0.77 & 0.63 & 0.82 & 0.56 & \textbf{0.88} & 0.64 \\
 & GCD & 0.72 & 0.70 & 0.71 & 0.69 & 0.68 & 0.61 & 0.70 & 0.70 & \underline{0.79} & \underline{0.79} & 0.76 & 0.64 & 0.72 & 0.74 \\
 & SimGCD & 0.67 & 0.66 & 0.67 & 0.66 & 0.64 & 0.60 & 0.67 & 0.67 & 0.68 & 0.67 & 0.72 & \underline{0.68} & 0.72 & \underline{0.76} \\
 & SelEx & 0.69 & 0.69 & 0.70 & 0.71 & 0.66 & 0.60 & 0.70 & 0.70 & 0.71 & 0.74 & 0.70 & \textbf{0.69} & 0.68 & 0.69 \\
 & SINR & 0.79 & \textbf{0.79} & 0.76 & \textbf{0.78} & \underline{0.83} & \underline{0.75} & \underline{0.77} & \textbf{0.87} & 0.79 & \textbf{0.83} & \underline{0.85} & 0.67 & \underline{0.79} & \textbf{0.83} \\
 & DINOv2 SINR & \textbf{0.80} & \underline{0.77} & \underline{0.77} & \underline{0.76} & \textbf{0.84} & \textbf{0.76} & 0.77 & \underline{0.85} & \textbf{0.81} & 0.79 & \textbf{0.86} & 0.66 & 0.75 & 0.73 \\
\bottomrule
\end{tabular}
}
\label{tab:aves_full_geo_breakdown_table1_acc}
\end{table}

As discussed in \cref{sec:limitations}, the construction pipeline may favor splits that are geographically separable.
To investigate how helpful these features are, we summarize the average effect of adding geographic features on clustering accuracy and AMI in \cref{tab:aves_full_geo_compressed_acc}, with results broken down by geographic region.
For this, we assign each novel coarse class in the full \dataset to a continent by plurality vote over its observation locations, and break down performance by continent.
The number of coarse classes per continent is broadly balanced, with the exception of Asia (72) and Oceania (18).
We observe some regional performance differences, \eg lower split-only performance in North America. Averaged over all eight non-oracle K-estimation/clustering methods, the gain from adding location features (Cl.\,Acc., then Cl.\,Acc.$_s$) is: Asia $+0.03$ / $+0.06$, North America $+0.02$ / $+0.08$, South America $+0.04$ / $+0.09$, Africa $+0.02$ / $+0.04$, Europe $+0.00$ / $+0.00$, and Oceania $-0.01$ / $-0.01$.
Looking at the clustering accuracy alone then, geographic features help most in some regions and not at all in others, however when considering the AMI we see improvement in every region.
In \Cref{tab:aves_full_geo_breakdown_table1_acc,tab:aves_full_geo_breakdown_table2_acc} we further investigate how performance varies across different regions, reporting clustering accuracy per continent for different backbones and methods.

\begin{table}[H]
\centering
\caption{\textbf{Geographic breakdown of Clustering Accuracy, \dataset{}-Full.} Each novel coarse category is assigned to a continent by plurality vote over its citizen-science observation locations (point-in-polygon against Natural Earth country boundaries). Columns are otherwise disjoint and sum to Overall ($n$=258). This is a method comparison: rows are K-estimation/clustering methods, sectioned by backbone $\times$ geo-feature config. Best value per column is bolded and second-best is underlined within each section, oracle excluded.}
\resizebox{\linewidth}{!}{
\begin{tabular}{c|l|cc|cc|cc|cc|cc|cc|cc}
\toprule
 & & \multicolumn{2}{c}{Overall ($n$=258)} & \multicolumn{2}{c}{Asia ($n$=72)} & \multicolumn{2}{c}{North America ($n$=60)} & \multicolumn{2}{c}{South America ($n$=40)} & \multicolumn{2}{c}{Africa ($n$=38)} & \multicolumn{2}{c}{Europe ($n$=30)} & \multicolumn{2}{c}{Oceania ($n$=18)} \\
 & Row & Cl.\,Acc. & Cl.\,Acc.$_\text{s}$ & Cl.\,Acc. & Cl.\,Acc.$_\text{s}$ & Cl.\,Acc. & Cl.\,Acc.$_\text{s}$ & Cl.\,Acc. & Cl.\,Acc.$_\text{s}$ & Cl.\,Acc. & Cl.\,Acc.$_\text{s}$ & Cl.\,Acc. & Cl.\,Acc.$_\text{s}$ & Cl.\,Acc. & Cl.\,Acc.$_\text{s}$ \\
\midrule
\multirow{9}{*}{\rotatebox{90}{DINOv2, w/o Geo}} & Oracle & 0.83 & 0.66 & 0.81 & 0.67 & 0.84 & 0.59 & 0.82 & 0.67 & 0.79 & 0.67 & 0.87 & 0.67 & 0.90 & 0.70 \\
 & Tuning-Agglo. & 0.67 & 0.62 & 0.66 & 0.62 & 0.65 & 0.53 & 0.69 & 0.63 & 0.69 & 0.65 & 0.65 & 0.63 & 0.68 & 0.71 \\
 & Tuning-Spectral & \underline{0.78} & 0.63 & \underline{0.77} & \underline{0.64} & 0.77 & \underline{0.57} & 0.77 & 0.63 & 0.75 & 0.63 & 0.79 & 0.63 & \textbf{0.88} & \underline{0.72} \\
 & Tuning-XMeans & \textbf{0.80} & 0.61 & \textbf{0.78} & 0.63 & \textbf{0.82} & 0.54 & \textbf{0.81} & 0.64 & \textbf{0.77} & 0.63 & \underline{0.82} & 0.56 & \underline{0.88} & 0.64 \\
 & Brent-KMeans & 0.77 & \underline{0.64} & 0.74 & 0.64 & 0.78 & 0.56 & 0.76 & 0.63 & \underline{0.76} & \textbf{0.69} & \textbf{0.83} & \textbf{0.66} & 0.85 & 0.68 \\
 & Brent-Agglo. & 0.76 & 0.61 & 0.74 & 0.62 & \underline{0.79} & 0.56 & 0.75 & 0.61 & 0.73 & 0.63 & 0.78 & 0.61 & 0.84 & 0.66 \\
 & Brent-Spectral & 0.75 & 0.63 & 0.76 & 0.62 & 0.73 & 0.55 & 0.76 & \underline{0.67} & 0.76 & 0.68 & 0.78 & \underline{0.65} & 0.73 & 0.71 \\
 & Brent-GMM & 0.76 & 0.62 & 0.74 & 0.61 & 0.75 & 0.54 & \underline{0.78} & \textbf{0.67} & 0.73 & 0.65 & 0.80 & 0.57 & 0.85 & \textbf{0.74} \\
 & Transformer & 0.71 & \textbf{0.65} & 0.70 & \textbf{0.65} & 0.70 & \textbf{0.59} & 0.71 & 0.65 & 0.73 & \underline{0.69} & 0.75 & 0.64 & 0.71 & 0.65 \\
\midrule
\multirow{9}{*}{\rotatebox{90}{DINOv2, w/ Geo}} & Oracle & 0.93 & 0.85 & 0.89 & 0.81 & 0.97 & 0.92 & 0.92 & 0.85 & 0.91 & 0.86 & 0.96 & 0.89 & 0.91 & 0.74 \\
 & Tuning-Agglo. & 0.60 & 0.70 & 0.67 & \underline{0.74} & 0.49 & 0.61 & 0.73 & \underline{0.79} & 0.65 & 0.70 & 0.46 & 0.58 & 0.57 & \underline{0.71} \\
 & Tuning-Spectral & 0.80 & 0.70 & 0.78 & 0.69 & 0.80 & 0.67 & \textbf{0.84} & 0.77 & \underline{0.81} & 0.76 & 0.80 & 0.59 & 0.79 & 0.62 \\
 & Tuning-XMeans & \underline{0.80} & 0.61 & 0.78 & 0.63 & 0.82 & 0.54 & 0.81 & 0.64 & 0.77 & 0.63 & 0.82 & 0.56 & \textbf{0.88} & 0.64 \\
 & Brent-KMeans & \textbf{0.81} & 0.67 & \textbf{0.81} & 0.70 & 0.81 & 0.65 & 0.79 & 0.67 & 0.79 & 0.71 & 0.81 & 0.59 & 0.83 & 0.62 \\
 & Brent-Agglo. & 0.80 & 0.66 & 0.78 & 0.68 & 0.81 & 0.64 & 0.80 & 0.67 & 0.77 & 0.66 & \underline{0.85} & 0.65 & 0.84 & 0.64 \\
 & Brent-Spectral & 0.78 & \underline{0.71} & 0.72 & 0.68 & 0.82 & \underline{0.72} & \underline{0.83} & 0.74 & 0.78 & \underline{0.76} & 0.77 & \underline{0.67} & \underline{0.86} & 0.68 \\
 & Brent-GMM & 0.80 & 0.69 & \underline{0.80} & 0.71 & \underline{0.82} & 0.70 & 0.80 & 0.68 & 0.75 & 0.65 & 0.85 & \textbf{0.68} & 0.79 & 0.69 \\
 & Transformer & 0.80 & \textbf{0.77} & 0.77 & \textbf{0.76} & \textbf{0.84} & \textbf{0.76} & 0.77 & \textbf{0.85} & \textbf{0.81} & \textbf{0.79} & \textbf{0.86} & 0.66 & 0.75 & \textbf{0.73} \\
\midrule
\multirow{9}{*}{\rotatebox{90}{SelEx, w/o Geo}} & Oracle & 0.85 & 0.69 & 0.84 & 0.72 & 0.84 & 0.60 & 0.84 & 0.70 & 0.82 & 0.72 & 0.88 & 0.69 & 0.90 & 0.69 \\
 & Tuning-Agglo. & 0.75 & 0.66 & 0.76 & 0.68 & 0.73 & \underline{0.58} & 0.75 & 0.66 & 0.73 & 0.70 & 0.74 & 0.64 & 0.79 & \textbf{0.78} \\
 & Tuning-Spectral & \underline{0.79} & 0.65 & \textbf{0.80} & 0.69 & 0.78 & 0.54 & 0.78 & 0.65 & 0.75 & 0.67 & 0.82 & \underline{0.67} & 0.87 & 0.72 \\
 & Tuning-XMeans & \textbf{0.80} & 0.61 & \underline{0.78} & 0.63 & \textbf{0.82} & 0.54 & \textbf{0.81} & 0.64 & \underline{0.77} & 0.63 & \underline{0.82} & 0.56 & \underline{0.88} & 0.64 \\
 & Brent-KMeans & 0.77 & 0.64 & 0.75 & 0.65 & 0.76 & 0.56 & 0.75 & 0.64 & 0.77 & 0.68 & 0.79 & 0.63 & \textbf{0.89} & \underline{0.72} \\
 & Brent-Agglo. & 0.79 & 0.63 & 0.76 & 0.63 & 0.78 & 0.56 & \underline{0.79} & 0.64 & \textbf{0.77} & 0.65 & \textbf{0.85} & 0.65 & 0.84 & 0.68 \\
 & Brent-Spectral & 0.76 & \underline{0.67} & 0.75 & \underline{0.70} & 0.73 & 0.57 & 0.78 & \underline{0.69} & 0.76 & \underline{0.72} & 0.74 & 0.64 & 0.79 & 0.64 \\
 & Brent-GMM & 0.77 & 0.62 & 0.75 & 0.64 & \underline{0.79} & 0.57 & 0.76 & 0.62 & 0.72 & 0.61 & 0.79 & 0.60 & 0.87 & 0.69 \\
 & Transformer & 0.69 & \textbf{0.69} & 0.70 & \textbf{0.71} & 0.66 & \textbf{0.60} & 0.70 & \textbf{0.70} & 0.71 & \textbf{0.74} & 0.70 & \textbf{0.69} & 0.68 & 0.69 \\
\midrule
\multirow{9}{*}{\rotatebox{90}{SelEx, w/ Geo}} & Oracle & 0.93 & 0.85 & 0.89 & 0.81 & 0.96 & 0.91 & 0.93 & 0.86 & 0.91 & 0.86 & 0.96 & 0.89 & 0.91 & 0.74 \\
 & Tuning-Agglo. & 0.60 & 0.71 & 0.68 & \textbf{0.76} & 0.49 & 0.59 & 0.73 & \underline{0.82} & 0.65 & 0.70 & 0.46 & 0.57 & 0.57 & 0.71 \\
 & Tuning-Spectral & 0.80 & 0.69 & 0.80 & 0.73 & 0.77 & 0.58 & 0.83 & 0.77 & 0.77 & 0.70 & 0.79 & 0.59 & 0.85 & 0.65 \\
 & Tuning-XMeans & 0.80 & 0.61 & 0.78 & 0.63 & \underline{0.82} & 0.54 & 0.81 & 0.64 & 0.77 & 0.63 & 0.82 & 0.56 & 0.88 & 0.64 \\
 & Brent-KMeans & 0.81 & 0.69 & \underline{0.81} & 0.70 & 0.80 & 0.62 & 0.83 & 0.75 & 0.78 & 0.70 & \underline{0.85} & \textbf{0.71} & 0.85 & 0.64 \\
 & Brent-Agglo. & \textbf{0.82} & 0.68 & 0.80 & 0.68 & \textbf{0.82} & 0.64 & \textbf{0.86} & 0.75 & \underline{0.80} & 0.69 & \textbf{0.87} & 0.67 & 0.86 & 0.64 \\
 & Brent-Spectral & 0.78 & \underline{0.72} & 0.80 & 0.73 & 0.77 & \textbf{0.68} & 0.78 & 0.73 & 0.79 & \underline{0.75} & 0.77 & 0.68 & 0.79 & \underline{0.74} \\
 & Brent-GMM & \underline{0.82} & 0.70 & \textbf{0.81} & 0.72 & 0.80 & 0.65 & \underline{0.84} & 0.70 & 0.78 & 0.70 & 0.83 & \underline{0.68} & \underline{0.88} & \textbf{0.77} \\
 & Transformer & 0.81 & \textbf{0.73} & 0.79 & \underline{0.74} & 0.82 & \underline{0.66} & 0.78 & \textbf{0.86} & \textbf{0.81} & \textbf{0.75} & 0.84 & 0.60 & \textbf{0.89} & 0.73 \\
\bottomrule
\end{tabular}
}
\label{tab:aves_full_geo_breakdown_table2_acc}
\end{table}

\subsection{Training Convergence}
\label{sec:convergence}
\Cref{tab:aves_small_train_convergence} compares performance on the known (training) coarse classes of \dataset-Small with the novel (test) coarse classes. For the transformer with a SelEx backbone, performance on the training classes is high, particularly for the splitting classes (Cl.\,Acc.$_s$ $0.93$, AMI$_s$ $0.83$), indicating that training converges and that the lower performance on the novel coarse classes reflects a generalization gap rather than a failure of training.

\begin{table}[H]
\centering
\caption{\textbf{Does training converge? Known (train) vs.\ novel (test) coarses on \textsc{PS-SPLIT-small}.} Same metrics and same K-estimation methods as \Cref{tab:aves_small_combined_backbone}, computed on the 64 \emph{known} coarse classes (i.e.\ the classes the backbone / K-estimator was trained on) instead of the 64 held-out novel coarses. Brent-KMeans rows re-run the same probe-based grid search leaving each known coarse's own probes out; Transformer rows re-run inference of the already-trained checkpoint over all 64 known coarses. No model is retrained.}
\resizebox{\linewidth}{!}{
\begin{tabular}{l|l|l|cccc|cc|cc}
\toprule
Method & Backbone & Split & Cl.\,Acc. & Cl.\,Acc.$_\text{s}$ & AMI & AMI$_\text{s}$ & K Acc. & K Acc.$_\text{s}$ & $\Delta\!\sum\!K$ & Split Prec. \\
\midrule
\multirow{4}{*}{Brent-KMeans} & \multirow{2}{*}{DINOv2} & Novel & 0.77 & 0.57 & 0.49 & 0.07 & 0.82 & 0.69 & -17 & 0.80 \\
 &  & Train & 0.72 & 0.55 & 0.38 & 0.04 & 0.75 & 0.64 & -14 & 0.50 \\
 & \multirow{2}{*}{SelEx} & Novel & 0.75 & 0.54 & 0.42 & 0.03 & 0.75 & 0.59 & -20 & 0.50 \\
 &  & Train & 0.79 & 0.61 & 0.56 & 0.21 & 0.78 & 0.61 & -22 & 0.70 \\
\midrule
\multirow{4}{*}{Transformer} & \multirow{2}{*}{DINOv2} & Novel & 0.70 & 0.62 & 0.28 & 0.16 & 0.76 & 0.81 & +7 & 0.51 \\
 &  & Train & 0.68 & 0.60 & 0.20 & 0.12 & 0.73 & 0.83 & +12 & 0.48 \\
 & \multirow{2}{*}{SelEx} & Novel & 0.68 & 0.66 & 0.15 & 0.17 & 0.75 & 0.94 & +24 & 0.50 \\
 &  & Train & 0.86 & 0.93 & 0.56 & 0.83 & 0.80 & 0.97 & +21 & 0.57 \\
\bottomrule
\end{tabular}
}
\label{tab:aves_small_train_convergence}
\end{table}

\clearpage
\section{Geographic Distribution}
\label{sec:geo_appendix}

PS-SPLIT inherits its spatial coverage from iNaturalist~\cite{iNaturalist}: every accepted observation in the dataset corresponds to a research-grade record uploaded by a citizen-science contributor and georeferenced at the point of observation. As a consequence, the geographic distribution of PS-SPLIT may more closely mirror that of iNaturalist participation rather than the true global distribution of avian biodiversity. Earlier benchmarks built from the same source~\cite{van2021benchmarking,macaodhaiccv19} have noted similar effects: contributor density is highest in North America, Western Europe, and parts of South and Southeast Asia, and is sparser across much of Africa, the Middle East, and central Asia. Methods that exploit the location metadata in PS-SPLIT (e.g., as in \cite{coleICML2023,chu2019geo}) should account for this sampling bias rather than treat observation density as an unbiased indicator of habitat preference. To partially mitigate the effect at the species level, the construction pipeline (Appendix~\ref{sec:construction}) caps the number of images per species which reduces the global imbalance somewhat, and prefers different uploaders, which works to flatten the localized geographic imbalance from single users uploading many observations from an area local to them.

\begin{figure}[H]
\centering
\includegraphics[width=\linewidth]{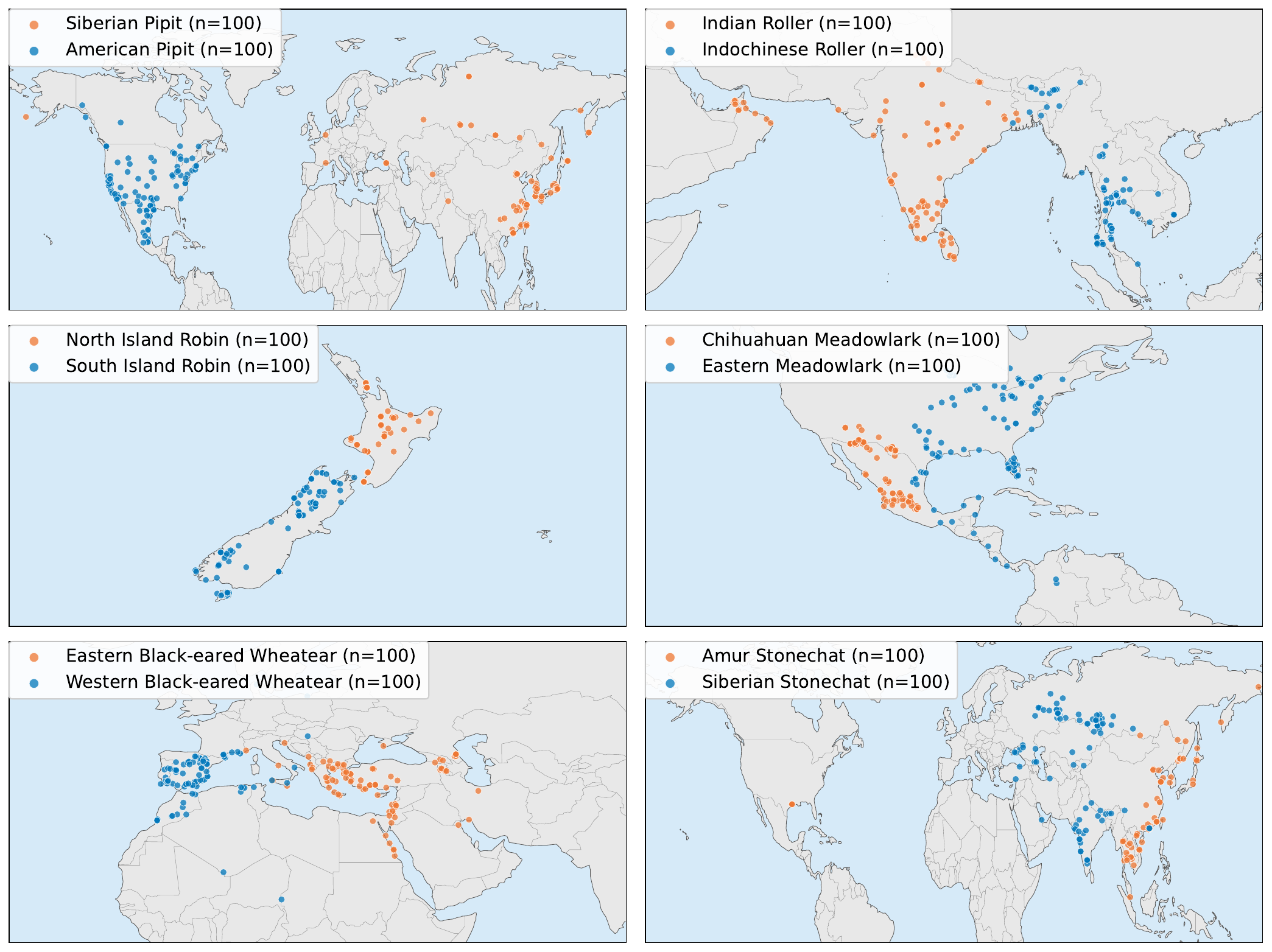}
\caption{\textbf{Geographic distributions of taxonomic split species pairs.} Pairs with clear range separation in left column: top, American Pipit vs.\ Siberian Pipit; middle, South Island Robin vs.\ North Island Robin; bottom, Western Black-eared Wheatear vs.\ Eastern Black-eared Wheatear. Pairs with overlapping or adjacent ranges in right column: top, Indochinese Roller vs.\ Indian Roller; middle, Eastern Meadowlark vs.\ Chihuahuan Meadowlark; bottom, Siberian Stonechat vs.\ Amur Stonechat.}
\label{fig:geo_pairs_combined}
\end{figure}

\begin{figure}[H]
\centering
\includegraphics[width=\linewidth]{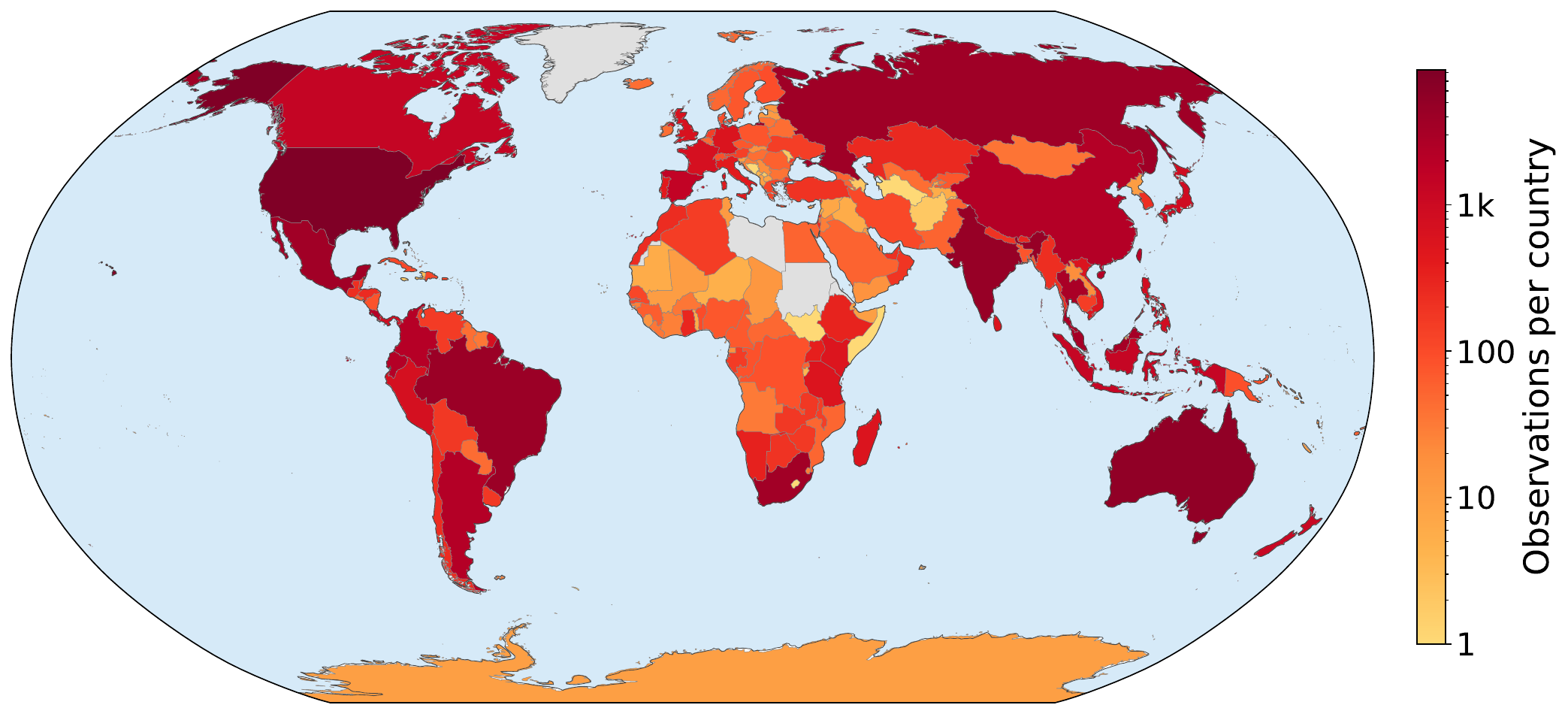}
\caption{\textbf{Per-country observation counts.} Countries with no accepted observations are shown in grey; remaining countries are coloured on a clipped log-scale colormap so that even single-observation countries remain page-visible. Several African and Central Asian countries contribute very few images, reflecting iNaturalist contributor density rather than absence of bird life.}
\label{fig:geo_region_map}
\end{figure}

\begin{figure}[H]
\centering
\begin{minipage}[t]{0.49\linewidth}
\centering
\includegraphics[width=\linewidth]{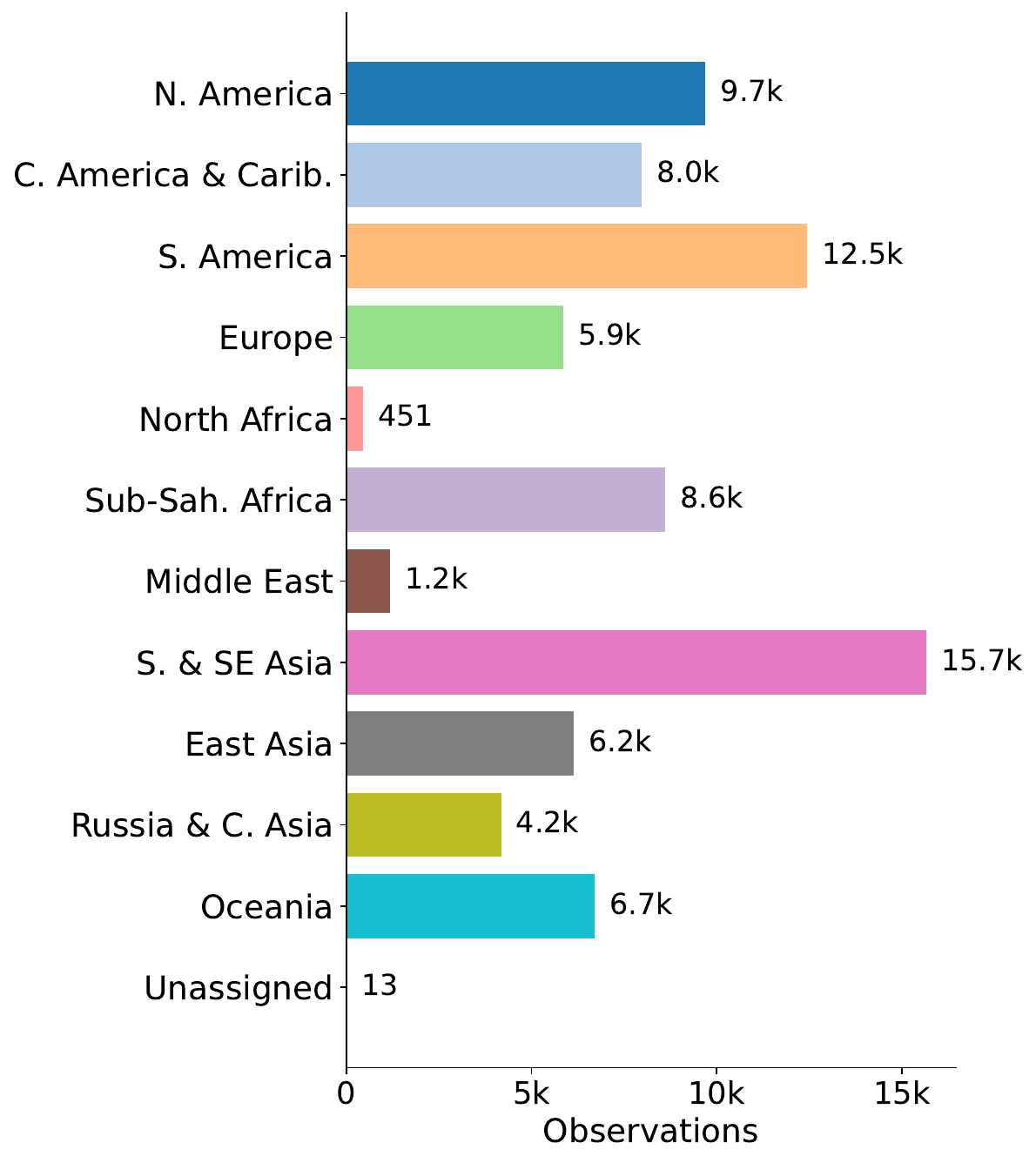}\\
{\small (a) by sub-continental region}
\end{minipage}\hfill
\begin{minipage}[t]{0.49\linewidth}
\centering
\includegraphics[width=\linewidth]{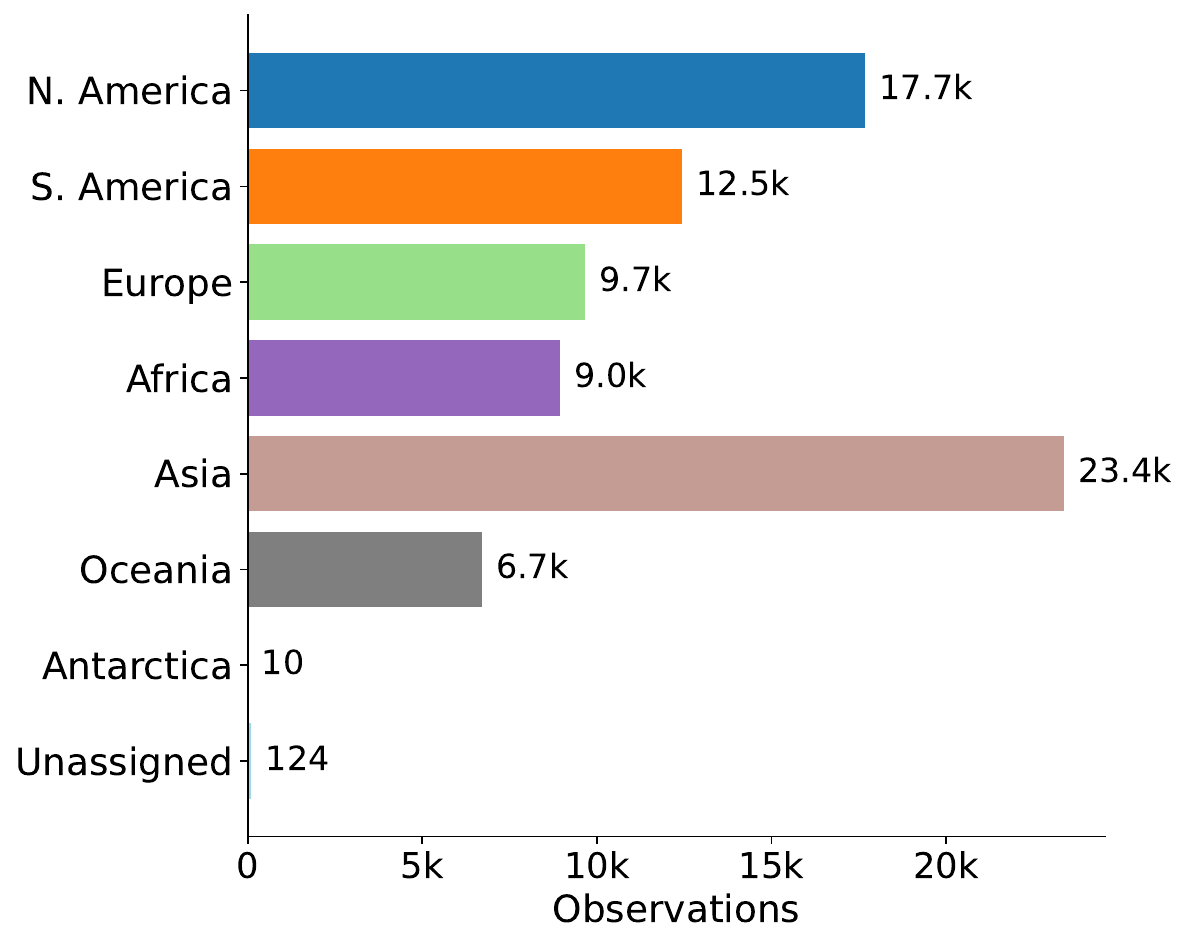}\\
{\small (b) by continent}
\end{minipage}
\caption{\textbf{Observation counts by region and continent.} (a) South \& SE~Asia is the largest sub-continental region (19.8\%), followed by South America (15.8\%) and North America (12.3\%). (b) At the continent level, Asia contributes 29.6\%, North America 22.4\%, and South America 15.8\%.}
\label{fig:geo_bars}
\end{figure}

\cref{tab:geo-regions-continents}, \cref{fig:geo_region_map}, and \cref{fig:geo_bars} summarize the resulting distribution. South \& SE~Asia is the largest single region (19.8\%), followed by South America (15.8\%) and North America (12.3\%); at the continent level, Asia contributes 29.6\% of accepted observations.
\cref{fig:geo_pairs_combined} shows examples of the geographic distributions of taxonomically split species, showing that in some cases there is clear separation between species, for example with each species localized to separate islands, however in other cases there is a continuous geographic distribution of both species, with no clear separation between them.

\begin{table}[H]
  \centering
  \caption{\textbf{Geographic distribution of accepted observations across PS-SPLIT.} Statistics are broken down by 11 sub-continental regions (left) and by continent (right).}
  \vspace{3pt}
  \resizebox{0.8\linewidth}{!}{
  \label{tab:geo-regions-continents}
  \begin{tabular}{lrr@{\hspace{2.5em}}lrr}
    \toprule
    \textbf{Region} & \textbf{Count} & \textbf{\%} & \textbf{Continent} & \textbf{Count} & \textbf{\%} \\
    \midrule
    North America                & 9{,}713  & 12.3 & North America   & 17{,}704 & 22.4 \\
    Central America \& Caribbean & 7{,}991  & 10.1 & South America   & 12{,}463 & 15.8 \\
    South America                & 12{,}463 & 15.8 & Europe          &  9{,}681 & 12.2 \\
    Europe                       & 5{,}877  & 7.4  & Africa          &  8{,}960 & 11.3 \\
    North Africa                 & 451      & 0.6  & Asia            & 23{,}397 & 29.6 \\
    Sub-Saharan Africa           & 8{,}623  & 10.9 & Oceania         &  6{,}722 & 8.5  \\
    Middle East                  & 1{,}183  & 1.5  & Antarctica      & 10       & 0.0  \\
    South \& SE Asia             & 15{,}676 & 19.8 & Unassigned      & 124      & 0.2  \\
    East Asia                    & 6{,}150  & 7.8  &                 &          &      \\
    Russia \& Central Asia       & 4{,}189  & 5.3  &                 &          &      \\
    Oceania                      & 6{,}722  & 8.5  &                 &          &      \\
    Unassigned                   & 13       & 0.0  &                 &          &      \\
    \midrule
    \textbf{Total}               & \textbf{79{,}061} & \textbf{100.0} & \textbf{Total} & \textbf{79{,}061} & \textbf{100.0} \\
    \bottomrule
  \end{tabular}
  }
\end{table}

\begin{figure}[H]
\centering
\includegraphics[width=\linewidth]{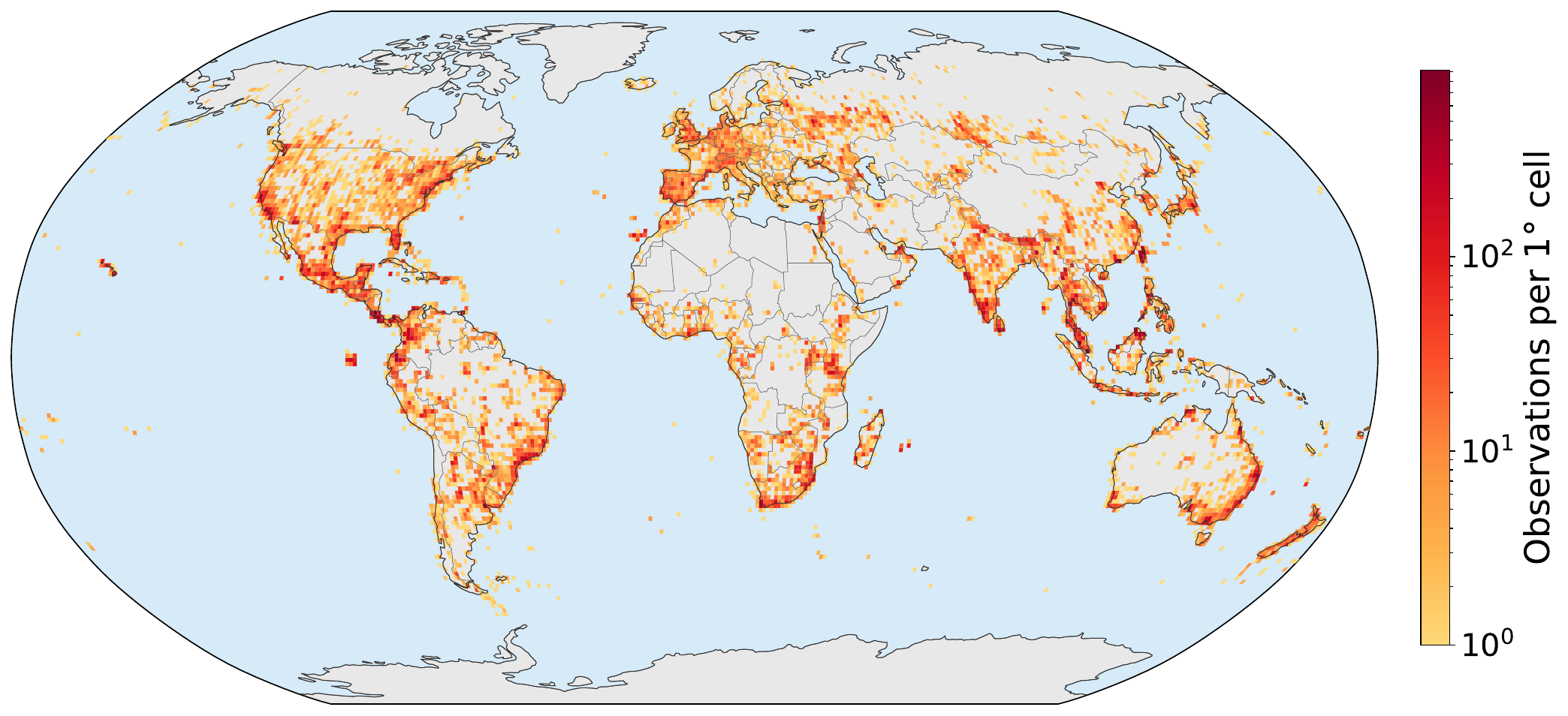}
\caption{\textbf{Global density of accepted observations.} Each $1^{\circ}$ grid cell is coloured by the number of accepted observations it contains (log scale, with the lower 25\% of the colormap clipped so that even single-observation cells are visible at zoomed-out reading sizes). Coverage spans all major continents; the densest cells occur in South and SE~Asia, South America, and North America.}
\label{fig:geo_heatmap_full}
\end{figure}

\section{Taxonomic Distribution}
\label{sec:taxonomy_appendix}

\begin{figure}[H]
\centering
\includegraphics[width=0.9\linewidth]{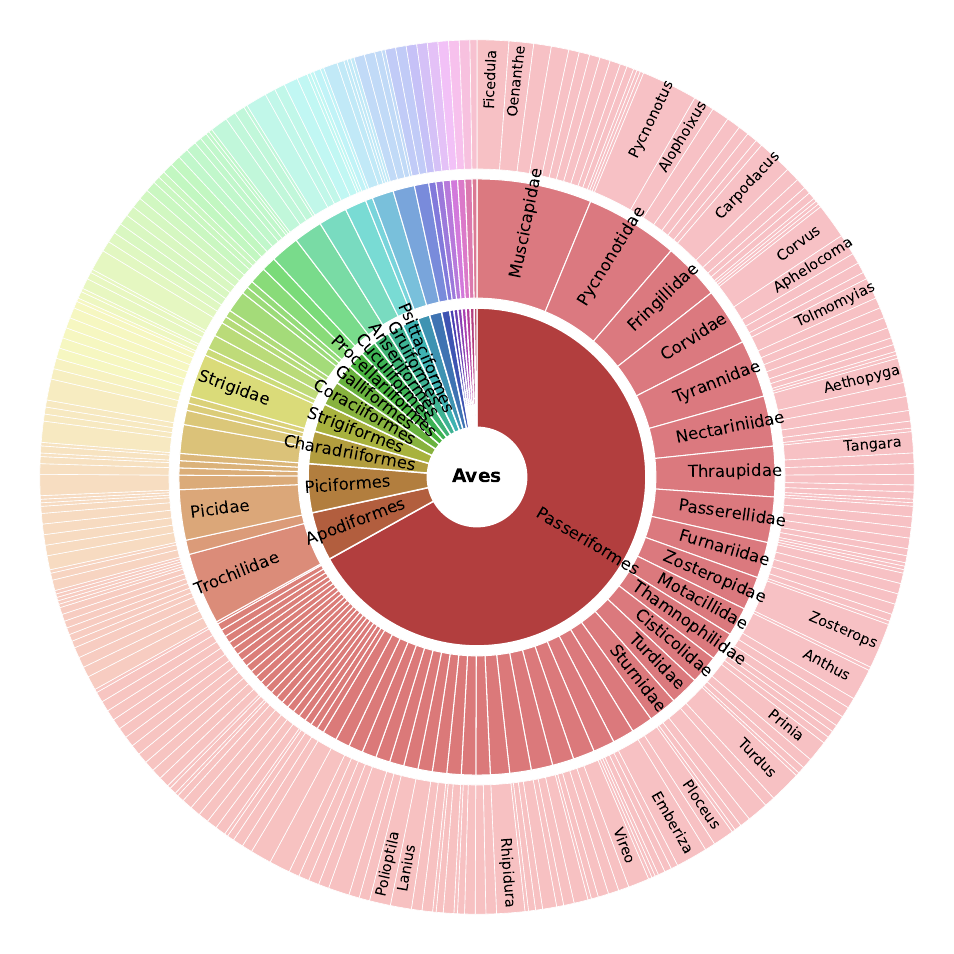}
\caption{\textbf{Taxonomic distribution of PS-SPLIT.} Concentric rings show Order (innermost), Family (middle), and Genus (outermost). Each segment is proportional to the number of species it contains. 
To keep the figure legible, only segments that are large enough to fit a readable label are labeled; the full list of orders, families, and genera is provided in \texttt{taxonomy\_full.json} (Appendix~\ref{sec:taxonomy_appendix}). 
The full dataset spans 22 orders, 94 families, and 262 genera. Passeriformes is the largest avian order, comprising more than half of all extant bird species~\cite{Clements2024eBirdClements}, and unsurprisingly accounts for the largest single share of PS-SPLIT.}
\label{fig:taxonomy_full}
\end{figure}

\begin{figure}[H]
\centering
\includegraphics[width=0.9\linewidth]{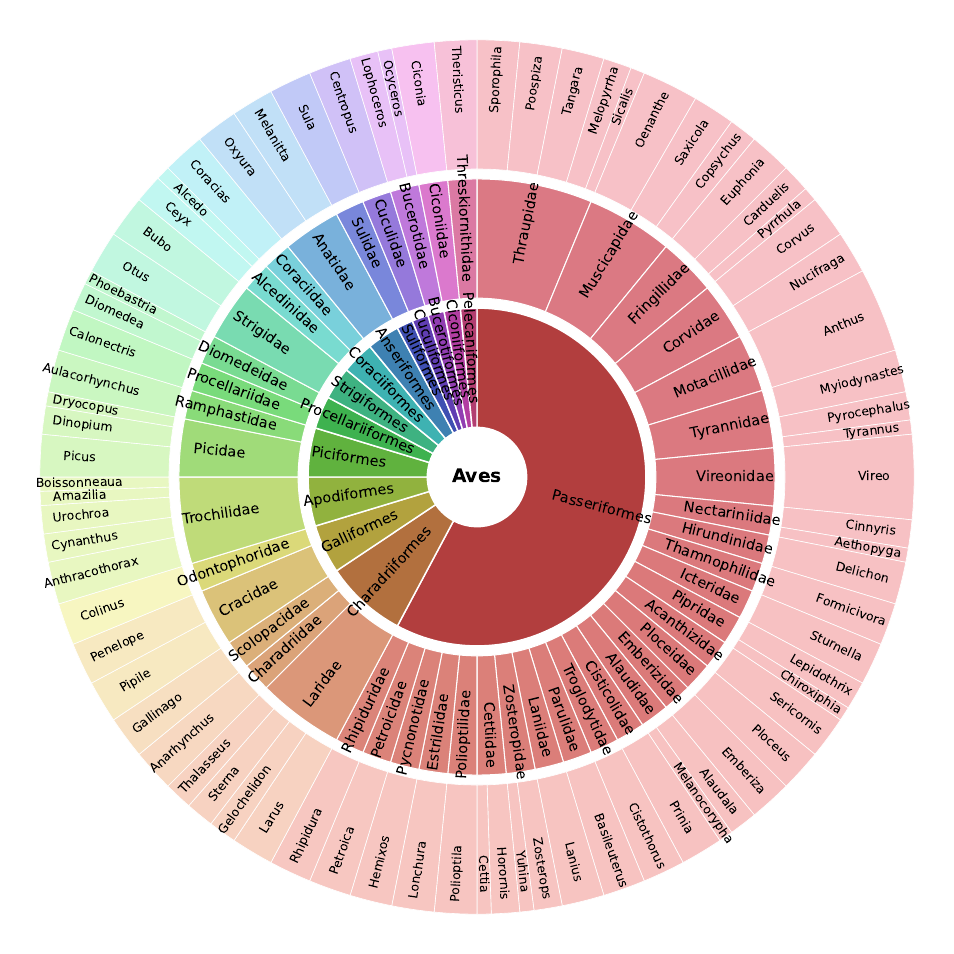}
\caption{\textbf{Taxonomic distribution of PS-SPLIT (small) (192 species).} Concentric rings show Order (innermost), Family (middle), and Genus (outermost). Each segment is proportional to the number of species it contains.
The small dataset spans 14 orders, 46 families, and 77 genera. As in the full dataset, Passeriformes accounts for the largest share of species, consistent with it being the largest avian order overall~\cite{Clements2024eBirdClements}. 
}
\label{fig:taxonomy_small}
\end{figure}

\dataset spans 22 orders, 94 families, and 262 genera; the small subset spans 14 orders, 46 families, and 77 genera.
\Cref{fig:taxonomy_full} visualises the order/family/genus distribution for the full dataset as concentric rings, with the larger segments labelled, and \cref{fig:taxonomy_small} shows the corresponding figure for the small subset.
The full per-species taxonomy --- iNaturalist taxon IDs, binomial nomenclature, splitting/control type, and per-species image counts --- is shipped as supplementary JSON files (\texttt{taxonomy\_full.json} and \texttt{taxonomy\_small.json}) rather than as in-paper tables, since 771 species do not fit naturally inside a PDF.
The JSON is hierarchical: each \texttt{Aves~$\to$~Order~$\to$~Family~$\to$~Genus~$\to$~Species} branch carries a \texttt{summary} block with \texttt{n\_orders}, \texttt{n\_families}, \texttt{n\_genera}, \texttt{n\_species}, and \texttt{n\_images} aggregated at that level, and each species leaf carries its iNaturalist \texttt{taxon\_id}, its \texttt{type} (\texttt{Split} or \texttt{Control}), and its image count.

\noindent\textbf{Coverage and balance.}
The taxonomy is dominated by \textit{Passeriformes} (perching birds), the largest avian order in nature with roughly half of all extant species~\cite{Clements2024eBirdClements}; \dataset reflects this --- Passeriformes accounts for the largest share of species in both \cref{fig:taxonomy_full} and \cref{fig:taxonomy_small}.
Other orders represented by multiple split--control pairs include \textit{Charadriiformes} (shorebirds and gulls), \textit{Apodiformes} (hummingbirds), \textit{Strigiformes} (owls), and \textit{Procellariiformes} (tubenoses); a long tail of additional orders contributes one or a few pairs each.
This concentration reflects underlying biology rather than any sampling choice: groups that are both species-rich and currently subject to active taxonomic revision contribute more documented splits.
The pairing of every splitting coarse class with a taxonomically close control (Appendix~\ref{sec:construction}) keeps the order/family distribution roughly symmetric between splitting and non-splitting halves.

\begin{figure}[H]
\centering

    \begin{minipage}{0.31\textwidth}
    \centering

    \begin{minipage}{0.32\linewidth}
        \includegraphics[width=\linewidth]{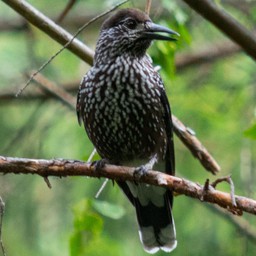}
    \end{minipage}%
    \hspace{0.2em}%
    \begin{minipage}{0.32\linewidth}
        \includegraphics[width=\linewidth]{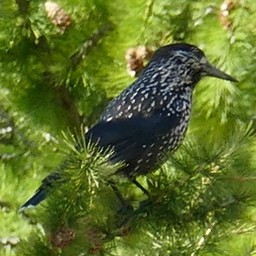}
    \end{minipage}%
    \hspace{0.2em}%
    \begin{minipage}{0.32\linewidth}
        \includegraphics[width=\linewidth]{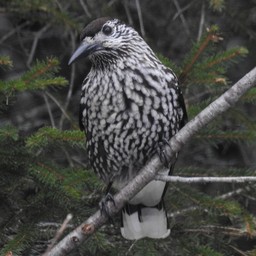}
    \end{minipage}

    \vspace{0.3em}

    \begin{minipage}{0.32\linewidth}
        \includegraphics[width=\linewidth]{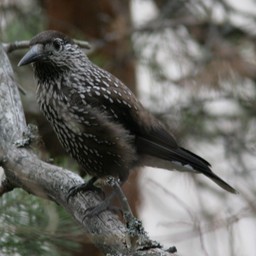}
    \end{minipage}%
    \hspace{0.2em}%
    \begin{minipage}{0.32\linewidth}
        \includegraphics[width=\linewidth]{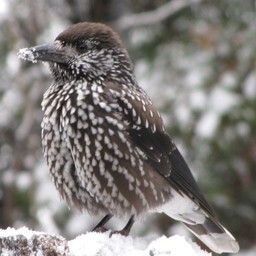}
    \end{minipage}%
    \hspace{0.2em}%
    \begin{minipage}{0.32\linewidth}
        \includegraphics[width=\linewidth]{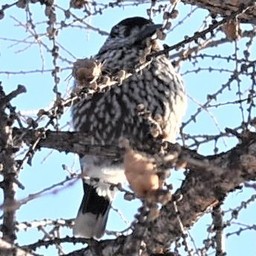}
    \end{minipage}

    \vspace{0.3em}

    \begin{minipage}{0.32\linewidth}
        \includegraphics[width=\linewidth]{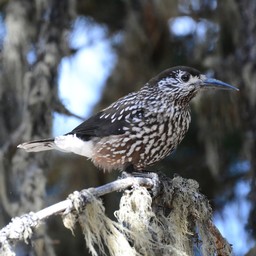}
    \end{minipage}%
    \hspace{0.2em}%
    \begin{minipage}{0.32\linewidth}
        \includegraphics[width=\linewidth]{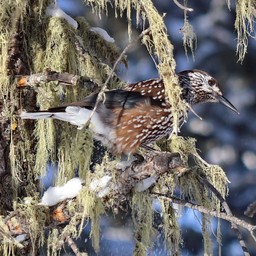}
    \end{minipage}%
    \hspace{0.2em}%
    \begin{minipage}{0.32\linewidth}
        \includegraphics[width=\linewidth]{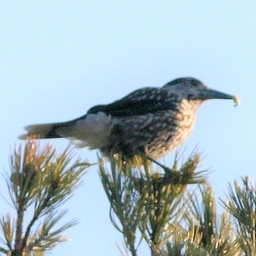}
    \end{minipage}

    \caption*{\textbf{(a) \textit{N. caryocatactes} (split)}}
    \end{minipage}%
    \hspace{0.5em}%
    \begin{minipage}{0.31\textwidth}
    \centering

    \begin{minipage}{0.32\linewidth}
        \includegraphics[width=\linewidth]{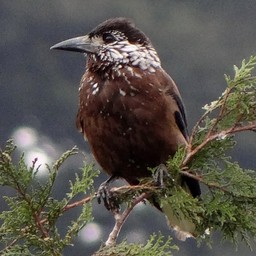}
    \end{minipage}%
    \hspace{0.2em}%
    \begin{minipage}{0.32\linewidth}
        \includegraphics[width=\linewidth]{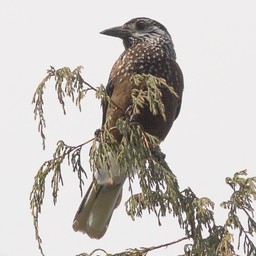}
    \end{minipage}%
    \hspace{0.2em}%
    \begin{minipage}{0.32\linewidth}
        \includegraphics[width=\linewidth]{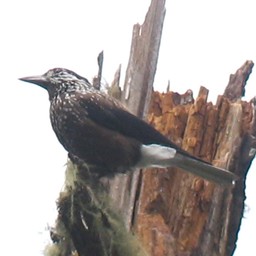}
    \end{minipage}

    \vspace{0.3em}

    \begin{minipage}{0.32\linewidth}
        \includegraphics[width=\linewidth]{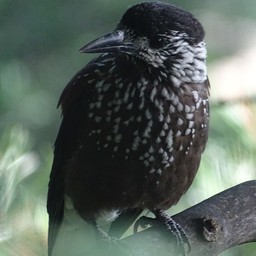}
    \end{minipage}%
    \hspace{0.2em}%
    \begin{minipage}{0.32\linewidth}
        \includegraphics[width=\linewidth]{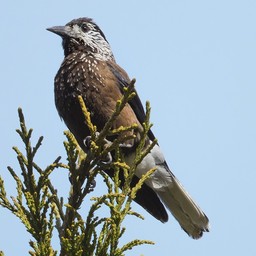}
    \end{minipage}%
    \hspace{0.2em}%
    \begin{minipage}{0.32\linewidth}
        \includegraphics[width=\linewidth]{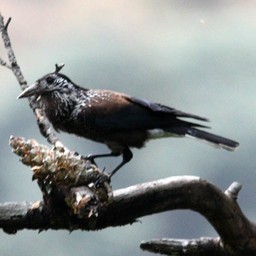}
    \end{minipage}

    \vspace{0.3em}

    \begin{minipage}{0.32\linewidth}
        \includegraphics[width=\linewidth]{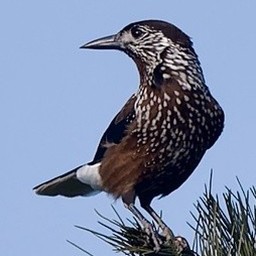}
    \end{minipage}%
    \hspace{0.2em}%
    \begin{minipage}{0.32\linewidth}
        \includegraphics[width=\linewidth]{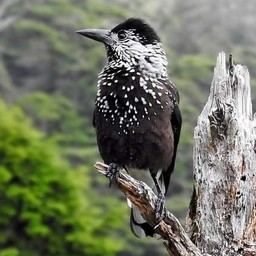}
    \end{minipage}%
    \hspace{0.2em}%
    \begin{minipage}{0.32\linewidth}
        \includegraphics[width=\linewidth]{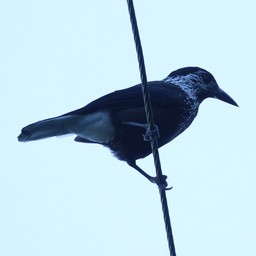}
    \end{minipage}

    \caption*{\textbf{(b) \textit{N. hemispila} (split)}}
    \end{minipage}%
    \hspace{0.5em}%
    \begin{minipage}{0.31\textwidth}
    \centering

    \begin{minipage}{0.32\linewidth}
        \includegraphics[width=\linewidth]{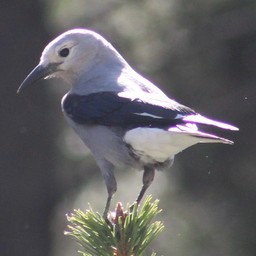}
    \end{minipage}%
    \hspace{0.2em}%
    \begin{minipage}{0.32\linewidth}
        \includegraphics[width=\linewidth]{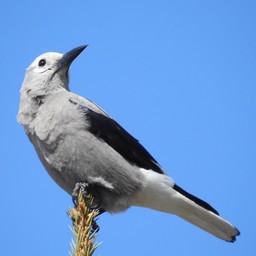}
    \end{minipage}%
    \hspace{0.2em}%
    \begin{minipage}{0.32\linewidth}
        \includegraphics[width=\linewidth]{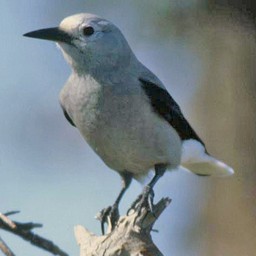}
    \end{minipage}

    \vspace{0.3em}

    \begin{minipage}{0.32\linewidth}
        \includegraphics[width=\linewidth]{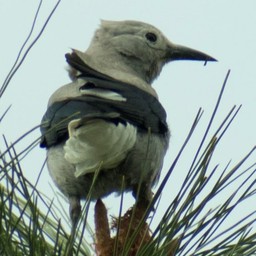}
    \end{minipage}%
    \hspace{0.2em}%
    \begin{minipage}{0.32\linewidth}
        \includegraphics[width=\linewidth]{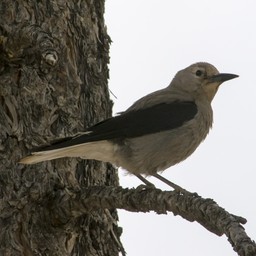}
    \end{minipage}%
    \hspace{0.2em}%
    \begin{minipage}{0.32\linewidth}
        \includegraphics[width=\linewidth]{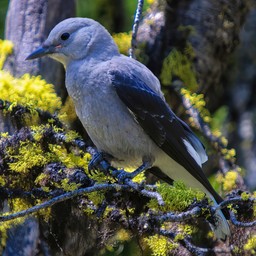}
    \end{minipage}

    \vspace{0.3em}

    \begin{minipage}{0.32\linewidth}
        \includegraphics[width=\linewidth]{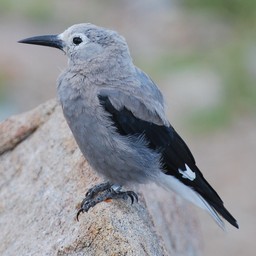}
    \end{minipage}%
    \hspace{0.2em}%
    \begin{minipage}{0.32\linewidth}
        \includegraphics[width=\linewidth]{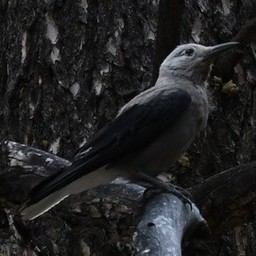}
    \end{minipage}%
    \hspace{0.2em}%
    \begin{minipage}{0.32\linewidth}
        \includegraphics[width=\linewidth]{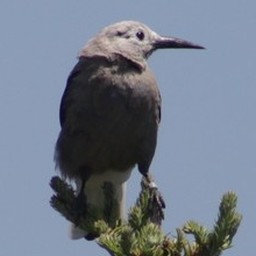}
    \end{minipage}

    \caption*{\textbf{(c) \textit{N. columbiana} (control)}}
    \end{minipage}

\vspace{-5pt}
\caption{\textbf{PS-SPLIT images for the Eurasian Nutcracker taxonomic split, with its taxonomically close control species.} The Southern Nutcracker \textit{Nucifraga hemispila} is substantially browner on the back and belly and much less spotted overall than the Northern Nutcracker \textit{Nucifraga caryocatactes} ~\cite{Clements2024eBirdClements}. Clark's Nutcracker \textit{Nucifraga columbiana} is a North American congener that has not undergone any recent taxonomic splits and serves as the size-balanced control for this pair.}
\label{fig:nutcracker_final}
\end{figure}

\section{Additional PS-SPLIT Images}
\label{sec:additional_images}

\begin{figure}[H]
\centering

    \begin{minipage}{0.31\textwidth}
    \centering

    \begin{minipage}{0.32\linewidth}
        \includegraphics[width=\linewidth]{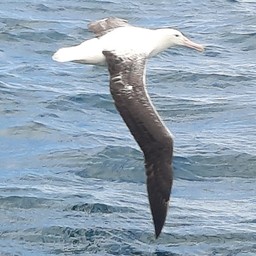}
    \end{minipage}%
    \hspace{0.2em}%
    \begin{minipage}{0.32\linewidth}
        \includegraphics[width=\linewidth]{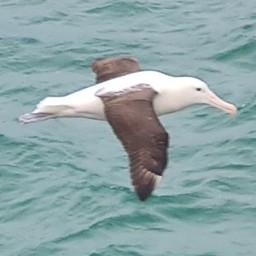}
    \end{minipage}%
    \hspace{0.2em}%
    \begin{minipage}{0.32\linewidth}
        \includegraphics[width=\linewidth]{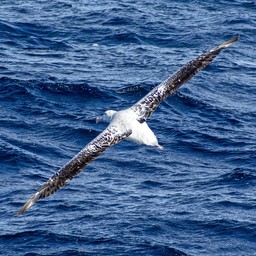}
    \end{minipage}

    \vspace{0.3em}

    \begin{minipage}{0.32\linewidth}
        \includegraphics[width=\linewidth]{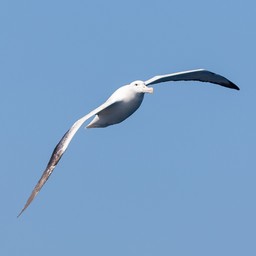}
    \end{minipage}%
    \hspace{0.2em}%
    \begin{minipage}{0.32\linewidth}
        \includegraphics[width=\linewidth]{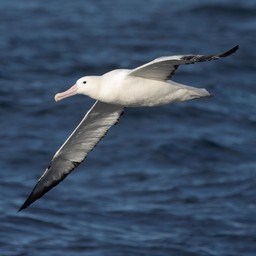}
    \end{minipage}%
    \hspace{0.2em}%
    \begin{minipage}{0.32\linewidth}
        \includegraphics[width=\linewidth]{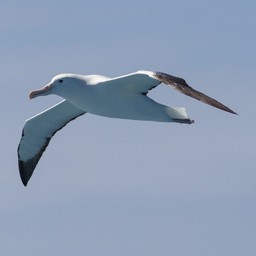}
    \end{minipage}

    \vspace{0.3em}

    \begin{minipage}{0.32\linewidth}
        \includegraphics[width=\linewidth]{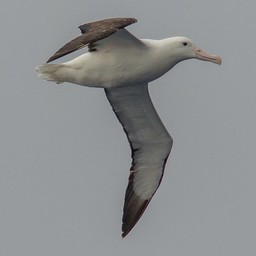}
    \end{minipage}%
    \hspace{0.2em}%
    \begin{minipage}{0.32\linewidth}
        \includegraphics[width=\linewidth]{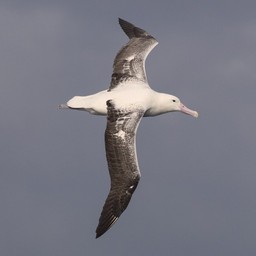}
    \end{minipage}%
    \hspace{0.2em}%
    \begin{minipage}{0.32\linewidth}
        \includegraphics[width=\linewidth]{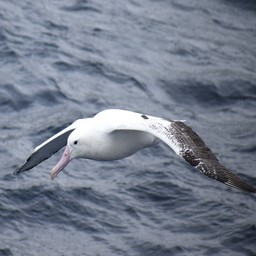}
    \end{minipage}

    \caption*{\textbf{(a) \textit{D. epomophora} (split)}}
    \end{minipage}%
    \hspace{0.5em}%
    \begin{minipage}{0.31\textwidth}
    \centering

    \begin{minipage}{0.32\linewidth}
        \includegraphics[width=\linewidth]{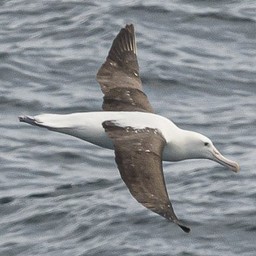}
    \end{minipage}%
    \hspace{0.2em}%
    \begin{minipage}{0.32\linewidth}
        \includegraphics[width=\linewidth]{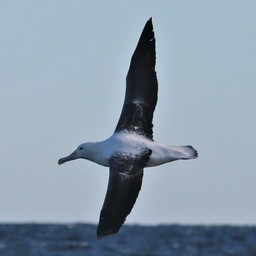}
    \end{minipage}%
    \hspace{0.2em}%
    \begin{minipage}{0.32\linewidth}
        \includegraphics[width=\linewidth]{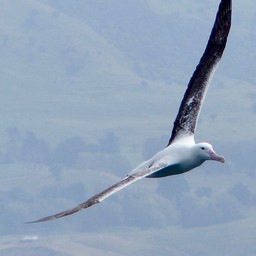}
    \end{minipage}

    \vspace{0.3em}

    \begin{minipage}{0.32\linewidth}
        \includegraphics[width=\linewidth]{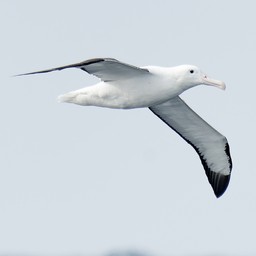}
    \end{minipage}%
    \hspace{0.2em}%
    \begin{minipage}{0.32\linewidth}
        \includegraphics[width=\linewidth]{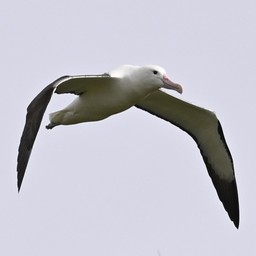}
    \end{minipage}%
    \hspace{0.2em}%
    \begin{minipage}{0.32\linewidth}
        \includegraphics[width=\linewidth]{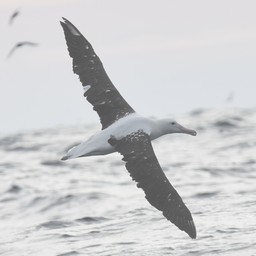}
    \end{minipage}

    \vspace{0.3em}

    \begin{minipage}{0.32\linewidth}
        \includegraphics[width=\linewidth]{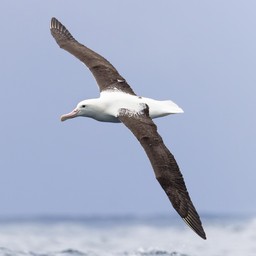}
    \end{minipage}%
    \hspace{0.2em}%
    \begin{minipage}{0.32\linewidth}
        \includegraphics[width=\linewidth]{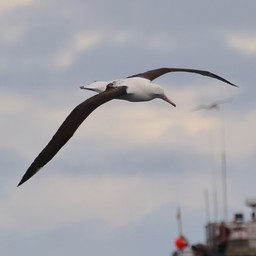}
    \end{minipage}%
    \hspace{0.2em}%
    \begin{minipage}{0.32\linewidth}
        \includegraphics[width=\linewidth]{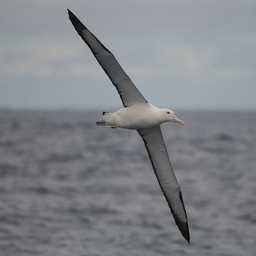}
    \end{minipage}

    \caption*{\textbf{(b) \textit{D. sanfordi} (split)}}
    \end{minipage}%
    \hspace{0.5em}%
    \begin{minipage}{0.31\textwidth}
    \centering

    \begin{minipage}{0.32\linewidth}
        \includegraphics[width=\linewidth]{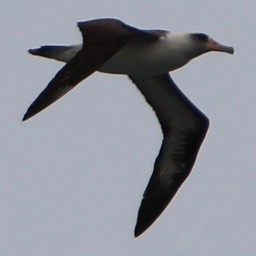}
    \end{minipage}%
    \hspace{0.2em}%
    \begin{minipage}{0.32\linewidth}
        \includegraphics[width=\linewidth]{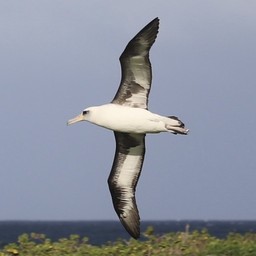}
    \end{minipage}%
    \hspace{0.2em}%
    \begin{minipage}{0.32\linewidth}
        \includegraphics[width=\linewidth]{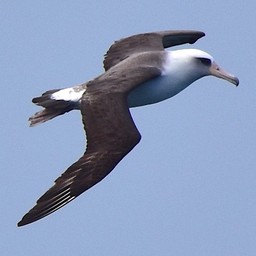}
    \end{minipage}

    \vspace{0.3em}

    \begin{minipage}{0.32\linewidth}
        \includegraphics[width=\linewidth]{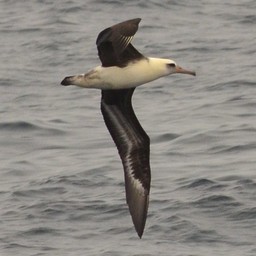}
    \end{minipage}%
    \hspace{0.2em}%
    \begin{minipage}{0.32\linewidth}
        \includegraphics[width=\linewidth]{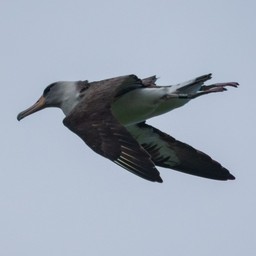}
    \end{minipage}%
    \hspace{0.2em}%
    \begin{minipage}{0.32\linewidth}
        \includegraphics[width=\linewidth]{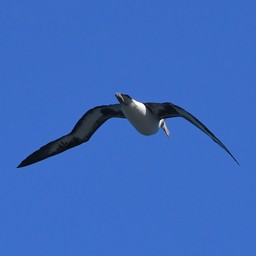}
    \end{minipage}

    \vspace{0.3em}

    \begin{minipage}{0.32\linewidth}
        \includegraphics[width=\linewidth]{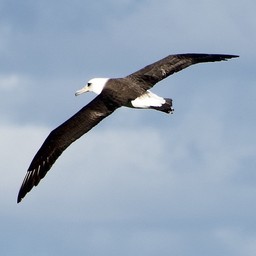}
    \end{minipage}%
    \hspace{0.2em}%
    \begin{minipage}{0.32\linewidth}
        \includegraphics[width=\linewidth]{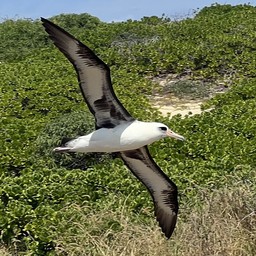}
    \end{minipage}%
    \hspace{0.2em}%
    \begin{minipage}{0.32\linewidth}
        \includegraphics[width=\linewidth]{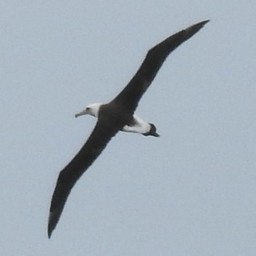}
    \end{minipage}

    \caption*{\textbf{(c) \textit{P. immutabilis} (control)}}
    \end{minipage}

\vspace{-5pt}
\caption{\textbf{PS-SPLIT images from the Royal Albatross taxonomic split, with its taxonomically close control species.} (a) Southern Royal Albatross \textit{Diomedea epomophora}; (b) Northern Royal Albatross \textit{Diomedea sanfordi}, recently elevated to species rank~\cite{Clements2024eBirdClements}. (c) Laysan Albatross \textit{Phoebastria immutabilis} is a closely related (same family Diomedeidae) control species that has not undergone any recent taxonomic splits.}
\label{fig:albatross_final}
\end{figure}

\begin{figure}[H]
\centering

    \begin{minipage}{0.47\textwidth}
    \centering

    \begin{minipage}{0.32\linewidth}
        \includegraphics[width=\linewidth]{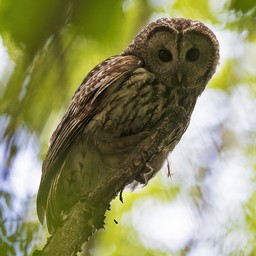}
    \end{minipage}%
    \hspace{0.2em}%
    \begin{minipage}{0.32\linewidth}
        \includegraphics[width=\linewidth]{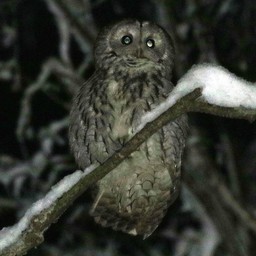}
    \end{minipage}%
    \hspace{0.2em}%
    \begin{minipage}{0.32\linewidth}
        \includegraphics[width=\linewidth]{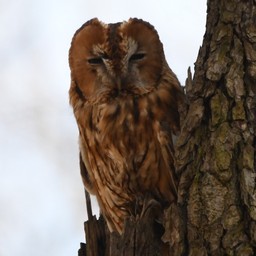}
    \end{minipage}

    \vspace{0.3em}

    \begin{minipage}{0.32\linewidth}
        \includegraphics[width=\linewidth]{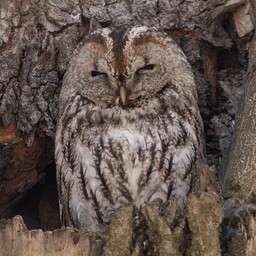}
    \end{minipage}%
    \hspace{0.2em}%
    \begin{minipage}{0.32\linewidth}
        \includegraphics[width=\linewidth]{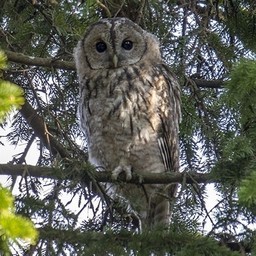}
    \end{minipage}%
    \hspace{0.2em}%
    \begin{minipage}{0.32\linewidth}
        \includegraphics[width=\linewidth]{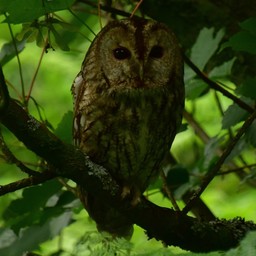}
    \end{minipage}

    \vspace{0.3em}

    \begin{minipage}{0.32\linewidth}
        \includegraphics[width=\linewidth]{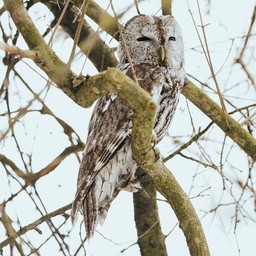}
    \end{minipage}%
    \hspace{0.2em}%
    \begin{minipage}{0.32\linewidth}
        \includegraphics[width=\linewidth]{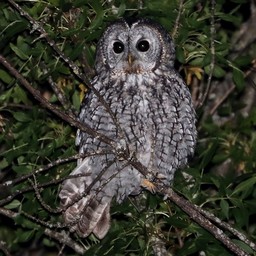}
    \end{minipage}%
    \hspace{0.2em}%
    \begin{minipage}{0.32\linewidth}
        \includegraphics[width=\linewidth]{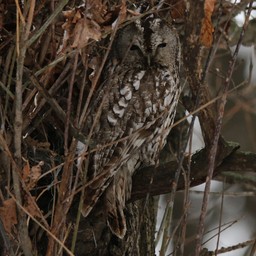}
    \end{minipage}

    \caption*{\textbf{(a) \textit{Strix aluco} (Tawny Owl)}}
    \end{minipage}
    \hspace{1em}
    \begin{minipage}{0.47\textwidth}
    \centering

    \begin{minipage}{0.32\linewidth}
        \includegraphics[width=\linewidth]{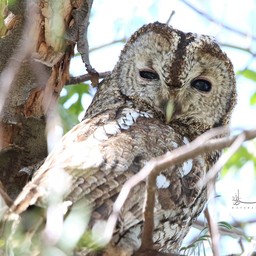}
    \end{minipage}%
    \hspace{0.2em}%
    \begin{minipage}{0.32\linewidth}
        \includegraphics[width=\linewidth]{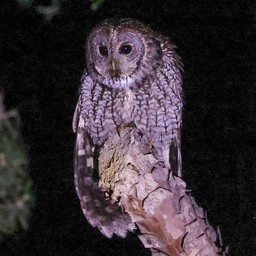}
    \end{minipage}%
    \hspace{0.2em}%
    \begin{minipage}{0.32\linewidth}
        \includegraphics[width=\linewidth]{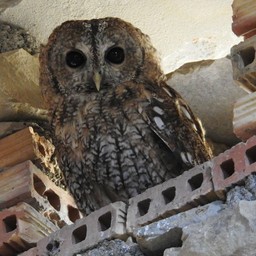}
    \end{minipage}

    \vspace{0.3em}

    \begin{minipage}{0.32\linewidth}
        \includegraphics[width=\linewidth]{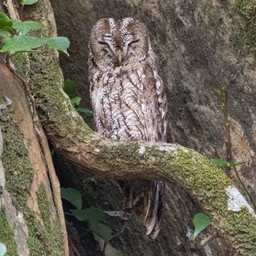}
    \end{minipage}%
    \hspace{0.2em}%
    \begin{minipage}{0.32\linewidth}
        \includegraphics[width=\linewidth]{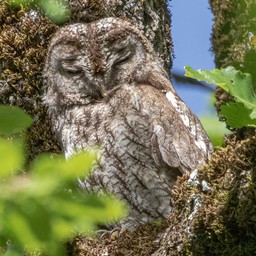}
    \end{minipage}%
    \hspace{0.2em}%
    \begin{minipage}{0.32\linewidth}
        \includegraphics[width=\linewidth]{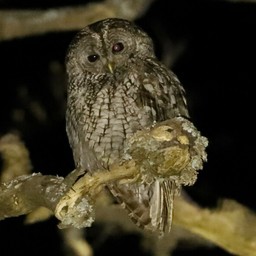}
    \end{minipage}

    \vspace{0.3em}

    \begin{minipage}{0.32\linewidth}
        \includegraphics[width=\linewidth]{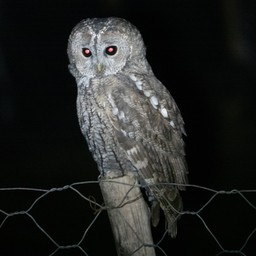}
    \end{minipage}%
    \hspace{0.2em}%
    \begin{minipage}{0.32\linewidth}
        \includegraphics[width=\linewidth]{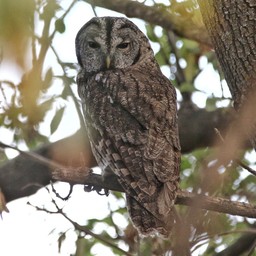}
    \end{minipage}%
    \hspace{0.2em}%
    \begin{minipage}{0.32\linewidth}
        \includegraphics[width=\linewidth]{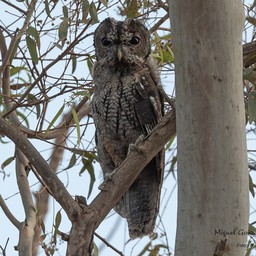}
    \end{minipage}

    \caption*{\textbf{(b) \textit{Strix mauritanica} (Maghreb Owl)}}
    \end{minipage}

\vspace{-5pt}
\caption{\textbf{A difficult split included in the full dataset: Tawny Owl and Maghreb Owl are visually near-indistinguishable.} The Maghreb Owl \textit{Strix mauritanica} was split from the Tawny Owl \textit{Strix aluco} in the 2021 Clements checklist. Despite being treated as distinct species, the two are essentially identical in plumage and cannot be reliably distinguished from photographs alone. This split is retained in the full dataset as an example of a genuinely hard fine-grained recognition problem.}
\label{fig:tawny_maghreb_removed}
\end{figure}

In \cref{fig:nutcracker_final,fig:albatross_final,fig:tawny_maghreb_removed} we provide some additional qualitative examples from \dataset.

\newpage
\section*{NeurIPS Paper Checklist}

\begin{enumerate}

\item {\bf Claims}
    \item[] Question: Do the main claims made in the abstract and introduction accurately reflect the paper's contributions and scope?
    \item[] Answer: \answerYes{}
    \item[] Justification: The abstract and \cref{sec:intro} state the dataset (\dataset; 79K bird images from 257 real taxonomic splits with 257 matched controls), the new task (discovery by category splitting), the four adapted GCD baselines, and the transformer K-estimator; all of these are evaluated in \cref{sec:exps}.

\item {\bf Limitations}
    \item[] Question: Does the paper discuss the limitations of the work performed by the authors?
    \item[] Answer: \answerYes{}
    \item[] Justification: \cref{sec:limitations} is a dedicated Limitations and Discussion section covering multiple limitations: visual-only assumption, birds-only restriction, $K_{\max}=2$ (two-way splits only), citizen-science biases, control matching being taxonomic rather than ecological, and splitting-only scope.

\item {\bf Theory assumptions and proofs}
    \item[] Question: For each theoretical result, does the paper provide the full set of assumptions and a complete (and correct) proof?
    \item[] Answer: \answerNA{}
    \item[] Justification: The paper presents an empirical benchmark with adapted baselines and contains no theoretical results or proofs.

    \item {\bf Experimental result reproducibility}
    \item[] Question: Does the paper fully disclose all the information needed to reproduce the main experimental results of the paper to the extent that it affects the main claims and/or conclusions of the paper (regardless of whether the code and data are provided or not)?
    \item[] Answer: \answerYes{}
    \item[] Justification: The dataset construction pipeline is fully described in \cref{sec:dataset} and Appendix~\ref{sec:construction}. Backbones (DINOv1, DINOv2 with registers, GCD, SimGCD, UNO+, SelEx, SINR) and the K-estimation / clustering methods are described in \cref{sec:baselines} with citations to the original implementations; the transformer K-estimator architecture, losses, optimiser, and training schedule are given in Appendix~\ref{sec:transformer_appendix}.

\item {\bf Open access to data and code}
    \item[] Question: Does the paper provide open access to the data and code, with sufficient instructions to faithfully reproduce the main experimental results, as described in supplemental material?
    \item[] Answer: \answerYes{}
    \item[] Justification: \cref{sec:dataset} the dataset and code (including the construction pipeline and the K-estimator implementations) will be released

\item {\bf Experimental setting/details}
    \item[] Question: Does the paper specify all the training and test details (e.g., data splits, hyperparameters, how they were chosen, type of optimizer) necessary to understand the results?
    \item[] Answer: \answerYes{}
    \item[] Justification: \cref{sec:baselines} describes backbones, geographic encoder, K-estimation methods, and clustering choices; the 50/50 known/novel coarse-class protocol is given in \cref{sec:task}; full transformer architecture and training hyperparameters (SGD with momentum 0.9, learning rate 0.1, 50 epochs, 5-head ensemble, $\lambda=0.3$, $\tau=0.07$, $K_{\max}=2$) are provided in Appendix~\ref{sec:transformer_appendix}.

\item {\bf Experiment statistical significance}
    \item[] Question: Does the paper report error bars suitably and correctly defined or other appropriate information about the statistical significance of the experiments?
    \item[] Answer: \answerNo{}
    \item[] Justification: Results are reported from a single training seed for each fine-tuned backbone; we did not run multi-seed experiments for the full backbone $\times$ K-estimation grid due to compute cost. Many category discovery papers report results from previous category discovery papers without reimplementing as training each approach can take multiple days per dataset depending on compute available. As we are introducing a new task and dataset, each backbone needed to be trained for each dataset and task, limiting the runs that were feasible. We will add error bars in a future revision.

\item {\bf Experiments compute resources}
    \item[] Question: For each experiment, does the paper provide sufficient information on the computer resources (type of compute workers, memory, time of execution) needed to reproduce the experiments?
    \item[] Answer: \answerYes{}
    \item[] Justification: Appendix~\ref{sec:transformer_appendix} reports that the transformer K-estimator trains in roughly ten minutes on a single A6000 GPU on full \dataset (assuming precomputed backbone features) and less on \dataset-Small. Backbone fine-tuning compute follows the original publications cited in \cref{sec:baselines}; all reported runs used a single A6000 GPU.

\item {\bf Code of ethics}
    \item[] Question: Does the research conducted in the paper conform, in every respect, with the NeurIPS Code of Ethics \url{https://neurips.cc/public/EthicsGuidelines}?
    \item[] Answer: \answerYes{}
    \item[] Justification: The work conforms with the NeurIPS Code of Ethics. The dataset is built only from research-grade iNaturalist observations whose licences permit redistribution for research; no human subjects are involved; \cref{sec:limitations} discusses appropriate use and the role of expert and genetic confirmation when methods evaluated on \dataset are applied in practice.

\item {\bf Broader impacts}
    \item[] Question: Does the paper discuss both potential positive societal impacts and negative societal impacts of the work performed?
    \item[] Answer: \answerYes{}
    \item[] Justification: \cref{sec:intro} motivates the work as supporting biodiversity monitoring and species discovery (positive impact). \cref{sec:limitations} discusses potential misuse: methods evaluated on \dataset should propose candidate splits for downstream expert review and genetic confirmation, not be used as a substitute for taxonomic decisions, and the dataset inherits known geographic and taxonomic biases that could affect deployment.

\item {\bf Safeguards}
    \item[] Question: Does the paper describe safeguards that have been put in place for responsible release of data or models that have a high risk for misuse (e.g., pre-trained language models, image generators, or scraped datasets)?
    \item[] Answer: \answerNA{}
    \item[] Justification: \dataset comprises research-grade citizen-science bird images released under licences that permit redistribution for research; it does not contain personal data, generative-model outputs, or content with high risk of misuse, so no special safeguards are required for release.

\item {\bf Licenses for existing assets}
    \item[] Question: Are the creators or original owners of assets (e.g., code, data, models), used in the paper, properly credited and are the license and terms of use explicitly mentioned and properly respected?
    \item[] Answer: \answerYes{}
    \item[] Justification: All cited assets are credited at point of use: iNaturalist research-grade observations~\cite{iNaturalist} (only those whose licences permit redistribution for research are included); MegaDetector v5~\cite{beery2019efficient}; Qwen2.5-VL-7B-Instruct~\cite{bai2025qwen25vltechnicalreport}; DINOv1~\cite{caron2021emerging}; DINOv2 with registers~\cite{oquab2023dinov2,darcet2024vision}; SINR~\cite{coleICML2023}; GCD~\cite{vaze22generalized}; SimGCD~\cite{wen2023simple}; UNO+~\cite{fini2021unified}; SelEx~\cite{rastegar2024selexselfexpertisefinegrainedgeneralized}; CUB-200-2011~\cite{wah2011caltech}; Stanford Cars~\cite{krause20133d}; FGVC Aircraft~\cite{maji2013fine}.

\item {\bf New assets}
    \item[] Question: Are new assets introduced in the paper well documented and is the documentation provided alongside the assets?
    \item[] Answer: \answerYes{}
    \item[] Justification: \dataset is the new asset. \cref{sec:dataset} provides headline statistics, the construction pipeline summary, and a comparison with existing fine-grained benchmarks; Appendix~\ref{sec:construction} contains the full construction pipeline; Appendix~\ref{sec:taxonomy_appendix} describes the supplementary per-species JSON release (taxon IDs, binomial nomenclature, splitting/control type, image counts); a datasheet, hosting plan, and licensing details accompany the submission as supplementary material.

\item {\bf Crowdsourcing and research with human subjects}
    \item[] Question: For crowdsourcing experiments and research with human subjects, does the paper include the full text of instructions given to participants and screenshots, if applicable, as well as details about compensation (if any)?
    \item[] Answer: \answerNA{}
    \item[] Justification: The paper does not involve crowdsourcing or research with human subjects. The underlying iNaturalist observations were contributed by volunteer naturalists for biodiversity science independently of and prior to this work; we use only the resulting research-grade images and associated metadata under their existing redistribution-permitting licences.

\item {\bf Institutional review board (IRB) approvals or equivalent for research with human subjects}
    \item[] Question: Does the paper describe potential risks incurred by study participants, whether such risks were disclosed to the subjects, and whether Institutional Review Board (IRB) approvals (or an equivalent approval/review based on the requirements of your country or institution) were obtained?
    \item[] Answer: \answerNA{}
    \item[] Justification: The work does not involve human-subjects research and therefore does not require IRB approval.

\item {\bf Declaration of LLM usage}
    \item[] Question: Does the paper describe the usage of LLMs if it is an important, original, or non-standard component of the core methods in this research? Note that if the LLM is used only for writing, editing, or formatting purposes and does \emph{not} impact the core methodology, scientific rigor, or originality of the research, declaration is not required.
    \item[] Answer: \answerYes{}
    \item[] Justification: Qwen2.5-VL-7B-Instruct~\cite{bai2025qwen25vltechnicalreport} is used as a vision--language quality reviewer in stage four of the dataset-construction pipeline (\cref{sec:dataset}, Appendix~\ref{sec:construction}); it scores seven binary quality criteria per image and assigns a coarse pose label, with intra-split pose consistency enforced. The full prompt template is reproduced in the appendix. No LLMs are used in the methods being benchmarked.

\end{enumerate}

\end{document}